\documentclass[nospthms,twocolumn]{svjour3}

\usepackage[svgnames,dvipsnames]{xcolor}
\usepackage[T1]{fontenc}
\usepackage{lmodern} 
\usepackage{graphicx}
\usepackage{amsmath,amsfonts,amssymb,amsthm}
\usepackage{pifont} 
\usepackage{mathtools,thmtools}
\usepackage{subcaption}
\usepackage{wrapfig}
\usepackage[normalem]{ulem}
\usepackage{ifthen}
\usepackage{booktabs}
\usepackage{capt-of}
\usepackage{footnote}
\usepackage{caption}
\usepackage{upgreek}
\usepackage{gensymb}
\usepackage{float}
\usepackage[inline]{enumitem}
\usepackage{import}
\usepackage{stmaryrd}
\usepackage[super]{nth}
\usepackage{bbm}
\usepackage{textcomp}
\usepackage{adjustbox}
\usepackage{stackengine}
\usepackage{complexity}
\usepackage{tabularx}

\usepackage[ruled,linesnumbered,nosemicolon]{algorithm2e}
\SetFuncArgSty{}
\definecolor{codegreen}{rgb}{0,0.6,0}

\newlength{\commentWidth}
\SetCommentSty{mycomntsty}

\usepackage[abbreviations,nonumberlist]{glossaries-extra}
\setabbreviationstyle{long-short}
\GlsXtrEnableInitialTagging{abbreviation}{\itag}

\newabbreviation{abb:iid}{\textit{i.i.d.}}{\itag{i}ndependent and \itag{i}dentically \itag{d}istributed}
\newabbreviation{abb:kl}{KL}{\itag{K}ullback-\itag{L}eibler}

\newabbreviation{abb:mim}{MIM}{\itag{m}an-in-the-\itag{m}iddle}

\newabbreviation{abb:ai}{AI}{\itag{A}rtificial \itag{I}ntelligence}

\newabbreviation{abb:ml}{ML}{\itag{M}achine \itag{L}earning}
\newabbreviation{abb:ood}{OOD}{\itag{O}ut \itag{O}f \itag{D}istribution}
\newabbreviation{abb:rl}{RL}{\itag{R}einforcement \itag{L}earning}
\newabbreviation{abb:knn}{\(k\)NN}{\itag{\(k\)}-\itag{n}earest \itag{n}eighbors}
\newabbreviation{abb:dgp}{DGP}{\itag{d}ata \itag{g}eneration \itag{p}rocess}

\newabbreviation[longplural={\itag{s}upport \itag{v}ector \itag{m}achines}]%
{abb:svm}{SVM}{\itag{S}upport \itag{V}ector \itag{M}achine}

\newabbreviation{abb:dl}{DL}{\itag{D}eep \itag{L}earning}
\newabbreviation{abb:drl}{DRL}{\itag{D}eep \itag{R}einforcement \itag{L}earning}

\newabbreviation[longplural={\itag{n}eural \itag{n}etworks}]%
{abb:nn}{NN}{\itag{N}eural \itag{N}etwork}

\newabbreviation[longplural={\itag{d}deep \itag{n}eural \itag{n}etworks}]%
{abb:dnn}{DNN}{\itag{D}eep \itag{N}eural \itag{N}etwork}

\newabbreviation[longplural={\itag{g}ated \itag{r}ecurrent \itag{u}nits}]%
{abb:gru}{GRU}{\itag{G}ated \itag{R}ecurrent \itag{U}nit}

\newabbreviation[longplural={\itag{g}raph \itag{n}eural \itag{n}etworks}]%
{abb:gnn}{GNN}{\itag{G}raph \itag{N}eural \itag{N}etwork}

\newabbreviation[longplural={\itag{c}onvolutional \itag{n}eural \itag{n}etworks}]%
{abb:cnn}{CNN}{\itag{C}onvolutional \itag{N}eural \itag{N}etwork}

\newabbreviation[longplural={\itag{g}enerative \itag{a}dversarial \itag{n}etworks}]%
{abb:gan}{GAN}{\itag{G}enerative \itag{A}dversarial \itag{N}etwork}

\newabbreviation{abb:mlp}{MLP}{\itag{M}ulti\itag{L}ayer \itag{P}erceptron}

\newabbreviation{abb:fl}{FL}{\itag{F}ederated \itag{L}earning}
\newabbreviation{abb:fedavg}{FedAvg}{\itag{fed}erated \itag{av}era\itag{g}ing}
\newabbreviation{abb:gar}{GAR}{\itag{G}radient \itag{A}ggregation \itag{R}ule}

\newabbreviation{abb:erm}{ERM}{\itag{E}mpirical \itag{R}isk \itag{M}inimization}
\newabbreviation{abb:vc}{VC}{\itag{V}apnik-\itag{C}hervonenkis}
\newabbreviation{abb:ferm}{FERM}%
{\itag{F}ederated \itag{E}mpirical \itag{R}isk \itag{M}inimization}
\newabbreviation{abb:dro}{DRO}{\itag{D}istributionally \itag{R}obust \itag{O}ptimization}
\newabbreviation{abb:irm}{IRM}{\itag{I}nvariant \itag{R}isk \itag{M}inimization}

\newabbreviation{abb:mnist}{MNIST}%
{\itag{M}odified \itag{N}ational \itag{I}nstitute of \itag{S}tandards and \itag{T}echnology}
\newabbreviation{abb:cifar}{CIFAR}{\itag{C}anadian \itag{I}nstitute for \itag{A}dvanced \itag{F}oundational \itag{R}esearch}

\newabbreviation[longplural={\itag{m}embership \itag{i}nference \itag{a}ttacks}]%
{abb:mia}{MIA}{\itag{m}embership \itag{i}nference \itag{a}ttack}

\newabbreviation[longplural={\itag{c}lass \itag{r}epresentative \itag{i}nference \itag{a}ttacks}]%
{abb:cria}{CRIA}{\itag{c}lass \itag{r}epresentative \itag{i}nference \itag{a}ttack}

\newabbreviation{abb:pia}{PIA}{\itag{P}roperty \itag{I}nference \itag{A}ttack}

\newabbreviation{abb:dra}{DRA}{\itag{D}ata \itag{R}ecoverability \itag{A}ttack}

\newabbreviation{abb:dlg}{DLG}{\itag{Deep} \itag{Leakage} from \itag{G}radients}

\newabbreviation{abb:PoL}{PoL}{\itag{p}roof \itag{o}f \itag{L}earning}
\newabbreviation{abb:ios}{IOS}{\itag{I}terative \itag{O}utlier \itag{S}cissor}

\newabbreviation{abb:dp}{DP}{\itag{d}ifferential \itag{p}rivacy}
\newabbreviation{abb:he}{HE}{\itag{h}omomorphic \itag{e}ncryption}
\newabbreviation{abb:smc}{SMC}{\itag{S}ecure \itag{M}ultiparty \itag{C}omputation}

\newabbreviation{abb:ec}{EC}{\itag{E}dge \itag{C}omputing}

\newabbreviation{abb:mec}{MEC}{\itag{M}obile \itag{E}dge \itag{C}omputing}
\newabbreviation{abb:fc}{FC}{\itag{F}og \itag{C}omputing}

\newabbreviation{abb:iot}{IoT}{\itag{I}nternet \itag{o}f \itag{T}hings}
\newabbreviation{abb:iov}{IoV}{\itag{I}nternet \itag{o}f \itag{V}ehicles}
\newabbreviation[%
  longplural={\itag{a}utonomous \itag{v}ehicles},%
]%
{abb:av}{AV}{\itag{A}utonomous \itag{V}ehicle}
\newabbreviation{abb:vanet}{VANET}{\itag{V}ehicular \itag{A}d-hoc \itag{NET}work}
\newabbreviation{abb:aiot}{AIoT}{\itag{A}rtificial \itag{I}ntelligence \itag{o}f \itag{T}hings}

\newabbreviation{abb:dq}{DQ}{\itag{D}ata \itag{Q}uality}
\newabbreviation{abb:ovb}{OVB}{\itag{O}mitted \itag{V}ariable \itag{B}ias}
\usepackage{asymptote}

\usepackage{forest}

\usepackage{tikz}
\usepackage{times}
\usetikzlibrary{trees,positioning,shapes,shadows,arrows.meta}
\usetikzlibrary{mindmap,backgrounds} 

\tikzset{%
  parent/.style  = {%
      draw, rounded corners=2pt, text width=3cm,
      align=center, font=\sffamily, rectangle,
      inner sep=.1cm, fill=blue!20
    },
  root/.style   = {parent, rounded corners=2pt, thin, align=center, fill=green!30},
  child/.style = {parent, thin, rounded corners=2pt, align=center, fill=pink!60,},
  gchild/.style = {parent, thin, rounded corners=2pt, align=left, fill=green!30, align=center},
  g2child/.style = {parent, thin, align=left, fill=yellow!30, align=center},
  g3child/.style = {parent, thin, align=left, fill=red!30, align=center},
  edge from parent/.style={draw=black, edge from parent fork left}
}
\newcommand{\cmark}{\textcolor{ForestGreen}{\ding{51}}}%
\newcommand{\xmark}{\textcolor{red}{\ding{55}}}%

\newcommand{\reals}{\mathbb{R}}       
\newcommand{\naturals}{\mathbb{N}}    

\newcommand{\positive}[1]{#1_{+}}

\let\epsilon\undefined
\newcommand{\epsilon}{\varepsilon}

\let\phi\undefined
\newcommand{\phi}{\varphi}

\let\hat\widehat

\let\tilde\widetilde

\let\originalmiddle=\middle                          
\def\middle#1{\mathrel{}\originalmiddle#1\mathrel{}} 

\newcommand{\abs}[1]{\left\lvert#1\right\rvert}

\DeclareMathOperator{\powerset}{\mathcal{P}}

\let\Pr\undefined
\DeclareMathOperator{\Pr}{\mathbb{P}}
\DeclareMathOperator{\expect}{\mathbb{E}}

\newcommand{\follows}{\mathrel{\sim}}
\newcommand{\independent}{\mathrel{\perp\kern-5pt\perp}}

\DeclareMathOperator*{\argmin}{argmin}

\DeclareMathOperator{\bigO}{\mathcal{O}}

\newcommand{\norm}[1]{\left\lVert#1\right\rVert}
\newcommand{\lpnorm}[2][2]{\norm{#2}_{#1}}

\newclass{\classP}{P}
\newclass{\classNP}{NP}
\newclass{\classNPO}{NPO}

\newlang{\knapsack}{KNAPSACK}

\DeclareMathOperator{\relu}{ReLU}

\theoremstyle{definition}

\newtheorem{principle}{Principle}
\theoremstyle{remark}
\newtheorem*{remark}{Remark}

\subimport{preamble}{hyperref}
\usepackage[sort&compress,numbers]{natbib}                                
\begin{document}

\title{Data Quality in Edge Machine Learning: A State-of-the-Art Survey}
\author{%
  Mohammed D. Belgoumri \and
  Mohamed Reda Bouadjenek \and
  Sunil Aryal \and
  Hakim Hacid
}
\date{Received: / Accepted:}

\maketitle

\begin{abstract}
  Data-driven \glsxtrfull{abb:ai} systems trained using \glsxtrfull{abb:ml} are shaping an ever-increasing (in size and importance) portion of our lives, including, but not limited to, recommendation systems, autonomous driving technologies, healthcare diagnostics, financial services, and personalized marketing.
  On the one hand, the outsized influence of these systems imposes a high standard of quality, particularly in the data used to train them.
  On the other hand, establishing and maintaining standards of \glsxtrfull{abb:dq} becomes more challenging due to the proliferation of Edge Computing and Internet of Things devices, along with their increasing adoption for training and deploying \glsxtrshort{abb:ml} models.
  The nature of the edge environment--characterized by limited resources, decentralized data storage, and processing--exacerbates data-related issues, making them more frequent, severe, and difficult to detect and mitigate.
  From these observations, it follows that \glsxtrshort{abb:dq} research for edge \glsxtrshort{abb:ml} is a critical and urgent exploration track for the safety and robust usefulness of present and future \glsxtrshort{abb:ai} systems.
  Despite this fact, \glsxtrshort{abb:dq} research for edge \glsxtrshort{abb:ml} is still in its infancy.
  The literature on this subject remains fragmented and scattered across different research communities, with no comprehensive survey to date.
  Hence, this paper aims to fill this gap by providing a global view of the existing literature from multiple disciplines that can be grouped under the umbrella of \glsxtrshort{abb:dq} for edge \glsxtrshort{abb:ml}.
  Specifically, we present a tentative definition of data quality in Edge computing, which we use to establish a set of \glsxtrshort{abb:dq} dimensions.
  We explore each dimension in detail, including existing solutions for mitigation.
\end{abstract}

\section{Introduction}\label{sec:introduction}

The recent surge in both the scale and prevalence of \gls{abb:ai} systems~\cite{murshedMachineLearningNetwork2022} has led to a significant increase in data and computational demands.
Consequently, the traditional model of solely relying on cloud computing for training and inference is reaching its limits~\cite{singhEdgeAISurvey2023}.
Combined with emerging concerns about data privacy and security, and recent advances in \gls{abb:ec}, this has led to a paradigm shift in the field of \gls{abb:ml}.
A new approach has emerged, which advocates for performing as much of the \gls{abb:ml} workload as possible close to the data source (on the so-called edge of the network), offloading only the most computationally intensive tasks to the cloud~\cite{wangEdgeAIConvergence2020}.

In doing so, edge \gls{abb:ml} promises several benefits over centralized \gls{abb:ml}.
One major such benefit is the reduction of the volume and frequency of data transfers,
thereby reducing latency, network bandwidth consumption, and the risk of data breaches%
~\cite{singhEdgeAISurvey2023,wangEdgeAIConvergence2020}.
Moreover, the distributed nature of the edge naturally precludes a single point of failure,
making the system more resilient and fault-tolerant~\cite{taikDataQualityBasedScheduling2021}.

However, edge \gls{abb:ml}, at least in its current form,
is not universally superior to centralized \gls{abb:ml}.
While it does address many of the latter's shortcomings,
it also introduces challenges of its own.
Chief among these challenges is the question of \gls{abb:dq}.
Broadly speaking, \gls{abb:dq} refers to the fitness of data for a given purpose%
~\cite{mahantiDataQualityDimensions2019,hassensteinDataQualityConcepts2022}.
Already a major concern in centralized \gls{abb:ml}~\cite{camachoQualityQualityOut2023},
\gls{abb:dq} is even more relevant in the context of edge \gls{abb:ml},
where resource constraints and the distributed nature of the system
conspire to make data faults more frequent, more severe, and harder to detect and correct.


Owing to the critical role \gls{abb:dq} plays
in the successful training and deployment of \gls{abb:ml} models on the edge,
many researchers have begun to investigate the topic,
producing a large and growing corpus of literature on the subject.
However, due to the multidisciplinary nature of \gls{abb:dq},
and the lack of an established, unified, and widely accepted framework for studying it,
the literature is fragmented,
with little communication or collaboration
between researchers from relevant but different fields.

\begin{table*}[hbt]
  \centering
  \adjustbox{max width=\textwidth}{%
    \begin{tabular}{lccccccccc}
      \toprule
                                                                  &               &                                            &                                        & \multicolumn{6}{c}{\textbf{Data quality}}                                                                                                \\
      \cmidrule{5-10}
      \textbf{Survey}                                             & \textbf{Year} & \textbf{\stackanchor{Federated}{learning}} & \textbf{\stackanchor{Edge}{computing}} & \emph{Independence}                       & \emph{Attribute skew} & \emph{Label noise} & \emph{Fairness} & \emph{Privacy} & \emph{Trust} \\
      \midrule
      \citet{sidiDataQualitySurvey2012}                           & 2012          & \xmark                                     & \xmark                                 & \xmark                                    & \cmark                & \cmark             & \xmark          & \xmark         & \cmark       \\
      \citet{wangEdgeAIConvergence2020}                           & 2020          & \cmark                                     & \cmark                                 & \xmark                                    & \xmark                & \cmark             & \cmark          & \cmark         & \cmark       \\
      \citet{hanSurveyLabelnoiseRepresentation2021}               & 2021          & \xmark                                     & \xmark                                 & \xmark                                    & \xmark                & \cmark             & \xmark          & \xmark         & \xmark       \\
      \citet{jereTaxonomyAttacksFederated2021}                    & 2021          & \cmark                                     & \xmark                                 & \xmark                                    & \xmark                & \xmark             & \xmark          & \cmark         & \cmark       \\
      \citet{xiaSurveyFederatedLearning2021}                      & 2021          & \cmark                                     & \cmark                                 & \xmark                                    & \cmark                & \xmark             & \xmark          & \cmark         & \cmark       \\
      \citet{yinComprehensiveSurveyPrivacypreserving2021}         & 2021          & \cmark                                     & \cmark                                 & \xmark                                    & \xmark                & \xmark             & \xmark          & \cmark         & \xmark       \\
      \citet{ferraguigSurveyBiasMitigation2021}                   & 2021          & \cmark                                     & \cmark                                 & \xmark                                    & \xmark                & \xmark             & \cmark          & \xmark         & \xmark       \\
      \citet{truongPrivacyPreservationFederated2021}              & 2021          & \cmark                                     & \cmark                                 & \xmark                                    & \xmark                & \xmark             & \xmark          & \cmark         & \xmark       \\
      \citet{zhuFederatedLearningNonIID2021}                      & 2021          & \cmark                                     & \cmark                                 & \xmark                                    & \cmark                & \cmark             & \xmark          & \xmark         & \xmark       \\
      \citet{abrehaFederatedLearningEdge2022}                     & 2022          & \cmark                                     & \cmark                                 & \xmark                                    & \cmark                & \xmark             & \xmark          & \cmark         & \cmark       \\
      \citet{liuThreatsAttacksDefenses2022}                       & 2022          & \cmark                                     & \xmark                                 & \xmark                                    & \xmark                & \xmark             & \xmark          & \cmark         & \cmark       \\
      \citet{mehrabiSurveyBiasFairness2022}                       & 2022          & \xmark                                     & \xmark                                 & \xmark                                    & \xmark                & \xmark             & \cmark          & \xmark         & \xmark       \\
      \citet{murshedMachineLearningNetwork2022}                   & 2022          & \cmark                                     & \cmark                                 & \xmark                                    & \xmark                & \xmark             & \xmark          & \cmark         & \cmark       \\
      \citet{xiaPoisoningAttacksFederated2023}                    & 2023          & \cmark                                     & \xmark                                 & \xmark                                    & \xmark                & \xmark             & \xmark          & \xmark         & \cmark       \\
      \citet{shiChallengesApproachesMitigating2022}               & 2022          & \cmark                                     & \xmark                                 & \xmark                                    & \xmark                & \xmark             & \xmark          & \xmark         & \cmark       \\
      \citet{rodriguez-barrosoSurveyFederatedLearning2023}        & 2023          & \cmark                                     & \xmark                                 & \xmark                                    & \cmark                & \xmark             & \xmark          & \cmark         & \cmark       \\
      \citet{abyaneUnderstandingQualityChallenges2023}            & 2023          & \cmark                                     & \xmark                                 & \xmark                                    & \xmark                & \xmark             & \cmark          & \cmark         & \cmark       \\
      \citet{whangDataCollectionQuality2023}                      & 2023          & \xmark                                     & \xmark                                 & \xmark                                    & \xmark                & \cmark             & \cmark          & \xmark         & \xmark       \\
      \citet{gallegosBiasFairnessLarge2023}                       & 2023          & \xmark                                     & \xmark                                 & \xmark                                    & \xmark                & \xmark             & \cmark          & \xmark         & \xmark       \\
      \citet{leeSurveySocialBias2023}                             & 2023          & \xmark                                     & \xmark                                 & \xmark                                    & \xmark                & \xmark             & \cmark          & \xmark         & \xmark       \\
      \citet{gongSurveyDatasetQuality2023}                        & 2023          & \xmark                                     & \xmark                                 & \xmark                                    & \cmark                & \cmark             & \xmark          & \cmark         & \xmark       \\
      \citet{wangCollaborativeMachineLearning2023}                & 2023          & \cmark                                     & \xmark                                 & \xmark                                    & \xmark                & \xmark             & \xmark          & \cmark         & \cmark       \\
      \citet{rafiFairnessPrivacyPreserving2024}                   & 2024          & \cmark                                     & \xmark                                 & \xmark                                    & \xmark                & \xmark             & \cmark          & \cmark         & \xmark       \\
      \citet{sanchezsanchezFederatedTrustSolutionTrustworthy2024} & 2024          & \cmark                                     & \cmark                                 & \xmark                                    & \xmark                & \xmark             & \cmark          & \cmark         & \cmark       \\
      Our survey                                                  & 2024          & \cmark                                     & \cmark                                 & \cmark                                    & \cmark                & \cmark             & \cmark          & \cmark         & \cmark       \\
      \bottomrule
    \end{tabular}%
  }
  \caption{Comparison of this survey to other surveys.}\label{tab:comparison_to_other_surveys}
\end{table*}

Simultaneously a symptom and a partial cause of this fragmentation,
is the lack of a survey of the state of the art
on \gls{abb:dq} in edge \gls{abb:ml}.
While a few comprehensive surveys have been published covering \gls{abb:dq}
in the general contexts of both
\gls{abb:ml}~\cite{gongDiversityMachineLearning2019},
and \gls{abb:ec}~\cite{karkouchDataQualityInternet2016},
to the best of our knowledge,
none exists that addresses it for the edge \gls{abb:ml} case.
Instead, there is a plethora of surveys,
each focusing on a different aspect of \gls{abb:ml} data quality,
many of which do address the specific case of edge \gls{abb:ml}.
\Cref{tab:comparison_to_other_surveys} provides a list of such surveys,
along with the aspects of \gls{abb:dq} they cover.


In this survey, we aim to provide a broad and comprehensive overview
of the state of the art on \gls{abb:dq} in edge \gls{abb:ml},
gathering works from the different \gls{abb:ml} research areas,
such as noise-resilience, fairness, privacy, and Byzantine fault tolerance
under the unifying umbrella of \gls{abb:dq}.
To this end, we establish a guiding principle for defining \gls{abb:dq}
based on the formulation of \gls{abb:ml} as an optimization problem.
We use this principle, along with theoretical sufficient conditions
for the convergence of \gls{abb:ml} algorithms,
to devise a list of \gls{abb:dq} dimensions.
For each dimension, we provide a detailed discussion of existing works,
including common definitions, the faults that can occur and the taxonomy thereof,
the proposed solutions,
and, where applicable, the ways in which the dimension has been studied
in the context of edge \gls{abb:ml}.


The remainder of this paper is organized as follows.
\Cref{sec:background} provides preliminary background information on edge \gls{abb:ml},
including a general discussion of \glsxtrlong{abb:ml},
a brief overview of the edge environment,
and the algorithms used to perform \gls{abb:ml} on the edge,
with particular focus on \glsxtrlong{abb:fl}.
\Cref{sec:data_quality} is dedicated to \gls{abb:dq} in \gls{abb:ml},
its definition, dimensions, and the families of solutions proposed to ensure it.
\Crefrange{sec:independence}{sec:trust} address one \gls{abb:dq} dimension each,
providing definitions, possible faults, proposed solutions (if any),
all in the context of edge \gls{abb:ml} (where literature permits).
Finally, \Cref{sec:conclusion} concludes the paper
with a summary of the main findings and a discussion of future research directions.

\section{Background}\label{sec:background}

Straightforwardly, edge \gls{abb:ml} can be defined as
performing some or all of the computation involved in
\gls{abb:ml} training and inference on edge devices%
~\cite{singhEdgeAISurvey2023}.
Therefore, understanding edge \gls{abb:ml} requires understanding
\glsxtrlong{abb:ml}, \glsxtrlong{abb:ec}, and the ways in which they interact.
This section discusses these three topics in order,
starting with a general overview of \gls{abb:ml},
followed by a brief discussion of \gls{abb:ec},
and finally, the algorithms used for \gls{abb:ml} on the edge.

\paragraph{Notations and conventions}\label{par:notations}

Throughout this document, we will make use of the notations presented in~\Cref{tab:notations}.
When we consider probabilities, expectations, or random variables on some space,
we implicitly assume a suitable \(\sigma-\)algebra and probability measure on that space.

\begin{table}[hbt]
  \begin{center}
    \begin{tabularx}{\linewidth}{cX}
      \toprule
      \textbf{Notation}                    & \textbf{Significance}                                                                                                      \\
      \midrule
      \(\mathcal{X}\)                      & Input space                                                                                                                \\
      \(\mathcal{Y}, \tilde{\mathcal{Y}}\) & Output space, and noisy output space respectively                                                                          \\
      \(\mathcal{H}\)                      & Hypothesis class                                                                                                           \\
      \(\Theta\)                           & Parameter space                                                                                                            \\
      \(\mathcal{R}, \mathcal{R}_{i}\)     & Global true risk, local true risk of client \(i\)                                                                          \\
      \(\hat{\mathcal{R}}_{\xi}\)          & Empirical risk with respect to \(\xi\)                                                                                     \\
      \(\theta^{\ast}\)                    & Bayes-optimal model (minimizer of \(\mathcal{R}\))                                                                         \\
      \(\hat{\theta}_{\xi}\)               & Minimizer of \(\hat{\mathcal{R}}_{\xi}\)                                                                                   \\
      \(\theta^{(t)},\theta_{i}^{(t)}\)    & Global model (resp. \(i^{\mathrm{th}}\) local model) at round \(t\)                                                        \\
      \(A \independent B\)                 & \(A\) is independent of \(B\)                                                                                              \\
      \(A \independent B \mid C\)          & \(A\) is independent of \(B\) given \(C\)                                                                                  \\
      \([n]\)                              & The set \(\{1, 2, \ldots, n\}\)                                                                                            \\
      \(\triangle^{n}\)                    & The simplex of \(\reals^{n}\), i.e., \(\left\{ \alpha \in \positive{\reals}^{n} \middle| \lpnorm[1]{\alpha} = 1 \right\}\) \\
      \bottomrule
    \end{tabularx}
  \end{center}
  \caption{Notations used in this document.}\label{tab:notations}
\end{table}

\subsection{General notions of machine learning}%
\label{sub:general_notions_of_ml}

The goal of \gls{abb:ml} is to learn from data algorithmically%
~\cite{%
  kulkarniElementaryIntroductionStatistical2011,%
  vapnikStatisticalLearningTheory1998,%
  abu-mostafaLearningDataShort2012%
}.
This vague goal can be formalized in different ways for different tasks.
For the sake of simplicity, we will focus on one particularly easy-to-define task,
namely, \emph{supervised learning}.
A more general formulation that covers most \gls{abb:ml} tasks is presented in%
~\cite{grohsMathematicalAspectsDeep2023}.

In supervised learning, the goal is to learn a function,
given a (finite) sample of input-output pairs generated from it,
where the input distribution is unknown, and the output is possibly noisy%
~\cite{raschkaMachineLearningPyTorch2022}.
One way to mathematically phrase this problem
is presented in this and the following paragraphs.
Given an \emph{input space} \(\mathcal{X}\),
an \emph{output space} \(\mathcal{Y}\),
a \emph{hypothesis class} \(\mathcal{H}\) of measurable functions
from \(\mathcal{X}\) to \(\mathcal{Y}\),
and a random pair \((X, Y)\) with values in \(\mathcal{X}\times\mathcal{Y}\),
with an unknown joint distribution \(\Pr_{(X, Y)}\),
the goal is to find a function \(f \in \mathcal{H}\),
that maximized the similarity between the distribution of \(f(X)\)
and that of \(Y\).

To make this problem completely unambiguous,
we need to specify a notion of similarity.
A common way to do so is to start with
a \emph{loss function} \(\ell: \mathcal{Y}^{2} \to \reals\),
intuitively thought of as a measure of dissimilarity between two outputs,
and using the functional
\begin{equation}\label{eq:risk}
  \mathcal{R}(f) \coloneqq \expect \left[ \ell(f(X), Y) \right],
\end{equation}
known as the \emph{risk} functional,
to measure \(f\)'s deviation from the true distribution.
The problem outlined above can then be expressed as finding the function
\(f^{\ast} \in \mathcal{H}\) (referred to as the \emph{Bayes optimal} function)
that minimizes \(\mathcal{R}\) i.e.,
\begin{equation}\label{eq:risk_minimization}
  f^{\ast} = \argmin_{f \in \mathcal{H}} \mathcal{R}(f).
\end{equation}
Solving \Cref{eq:risk_minimization} is often infeasible in practice
because the distribution \(\Pr_{(X, Y)}\) is unknown,
making the risk functional \Cref{eq:risk} uncomputable.
This problem can be circumvented in the presence of a sample
\[
  \xi = \left\{ (X_{1}, Y_{1}), (X_{2}, Y_{2}), \ldots, (X_{n}, Y_{n}) \right\}
\]
of \gls{abb:iid} random pairs such that
\((X_{i}, Y_{i}) \follows \Pr_{(X, Y)}, i \in [n]\).
In this case, the weak law of large numbers gives the following approximation
\begin{equation}\label{eq:empirical_risk}
  \hat{\mathcal{R}}_{\xi}(f) \coloneqq \frac{1}{n} \sum_{i=1}^{n} \ell(f(X_{i}), Y_{i}),
\end{equation}
of the risk functional,
where \(\hat{\mathcal{R}}_{\xi}\)  is referred to as the \emph{empirical risk functional}
(with respect to the sample \(\xi\)).
By solving the following \emph{\gls{abb:erm}} problem
\begin{equation}\label{eq:empirical_risk_minimization}
  \hat{f}_{\xi} = \argmin_{f \in \mathcal{H}} \hat{\mathcal{R}}_{\xi}(f)
\end{equation}
with a large enough sample size \(n\),
one can hope to obtain a good approximation of \(f^{\ast}\).
Put differently, one can hope that the solution \(\hat{f}_{\xi}\)
of \Cref{eq:empirical_risk_minimization} converges to \(f^{\ast}\) as \(n \to +\infty\).
In this case, we say that \gls{abb:erm} is \emph{consistent}%
~\cite{%
  vapnikPrinciplesRiskMinimization1991,%
  vapnikNatureStatisticalLearning2010%
}.

To further simplify it, and make it computationally feasible,
the learning problem is often formulated for a
finite-dimensionally parametrized hypothesis class.
For some \emph{parameter dimension} \(P \in \naturals\),
and a \emph{parameter space} \(\Theta \subset \reals^{P}\),
the hypothesis class is defined as
\[\mathcal{H} = \left\{ f_{\theta} \middle| \theta \in \Theta \right\},\]
reducing the problems of risk minimization and \gls{abb:erm} to
\begin{align}
  \theta^{\ast} =      & \argmin_{\theta \in \Theta} \mathcal{R}(\theta)\label{eq:parametric_risk_minimization} \\
  \hat{\theta}_{\xi} = & \argmin_{\theta \in \Theta} \hat{\mathcal{R}}_{\xi}(\theta)\label{eq:parametric_erm}
\end{align}
respectively, where the abuse of notation
\(\mathcal{R}(\theta) \coloneqq \mathcal{R}(f_{\theta})\) and
\(\hat{\mathcal{R}}_{\xi}(\theta) \coloneqq \hat{\mathcal{R}}_{\xi}(f_{\theta})\)
is used.
This form of the problem has the notable advantage of a finite-dimensional solution space,
allowing the use of standard optimization algorithms.
In particular, if \(\ell\) and \(\mathcal{H}\) are chosen such that
\(\hat{\mathcal{R}}_{\xi}\) is differentiable, convex, and bounded below,
then \Cref{eq:parametric_erm} can be solved using gradient-descent.
For these reasons, and the fact that most \gls{abb:ml} models are of this form
(e.g., linear regression, logistic regression, neural networks),
the rest of this work will refer to \Cref{eq:parametric_erm} as the \gls{abb:erm} problem.

\subsection{Edge computing}\label{sub:edge_computing}

The edge of the network (or simply the edge) can be defined as the interface
between the outer layers of the network,
represented by user devices commonly referred to as \emph{end devices}
(e.g., sensors, smartphones, and laptops),
and its \emph{core} of the network (often represented by a cloud server)%
~\cite{singhEdgeAISurvey2023}.
This definition can be vague, as the internet does not come with a clear predefined hierarchy.
As a result, the terminology used to describe the edge is often ambiguous,
particularly when it comes to the distinction
between \emph{end devices} and \emph{edge nodes}%
~\cite{wangEdgeAIConvergence2020}.
In this work, we adopt the terminology proposed by~\citet{wangEdgeAIConvergence2020}:
\begin{enumerate}
  \item \emph{End devices} or \emph{end nodes}, are the bottom layer of the network.
        Any mobile device (e.g., \gls{abb:iot} device, smartphone, or \gls{abb:av})
        with some (usually limited) computational power can be considered an end device.
        Due to their limited computational power,
        end devices are only capable of performing the most basic \gls{abb:ml} tasks,
        such as inference.
  \item \emph{Edge nodes}, edge servers, or edge devices,
        are the middle layer of 3-tiered end-edge-cloud architecture.
        They are usually more powerful than end devices,
        and are thus capable of performing more complex \gls{abb:ml} tasks,
        such as training or orchestration.
\end{enumerate}

\begin{figure}[hbt]
  \begin{center}
    \includegraphics[width=.8\linewidth]{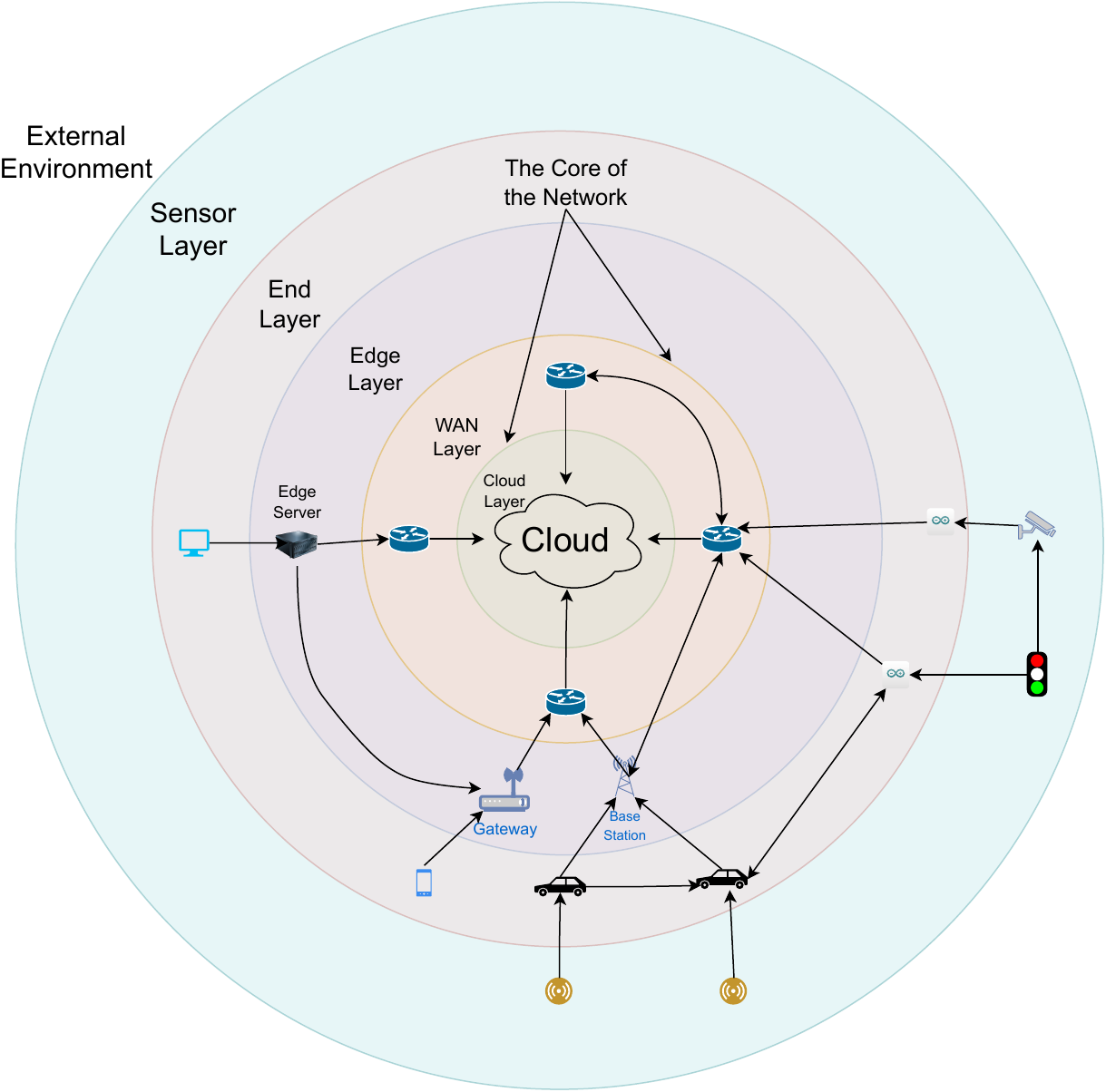}
  \end{center}
  \caption{Network architecture for edge computing~\cite{zhouEdgeIntelligencePaving2019}.}%
  \label{fig:edge_architecture}
\end{figure}

A more fine-grained hierarchy for the network graph is given in \Cref{fig:edge_architecture}.
Where the core is made up of the 2 innermost layers,
the edge is the \nth{3} layer, and end devices occupy the next layer.

The difficulty of defining and agreeing on a clear hierarchy for the edge
is partially behind the emergence of multiple paradigms for \gls{abb:ec}.
These include \emph{cloudlet computing} and \gls{abb:mec},
which rely on a 3-tiered architecture,
and \gls{abb:fc},
which organizes the network into an end-to-cloud \emph{continuum} rather than a hierarchy%
~\cite{singhEdgeAISurvey2023}.

\subsection{Edge machine learning}\label{sec:edge_ml}

\begin{figure*}[hbt]
  \begin{center}
    \includegraphics[width=.8\linewidth]{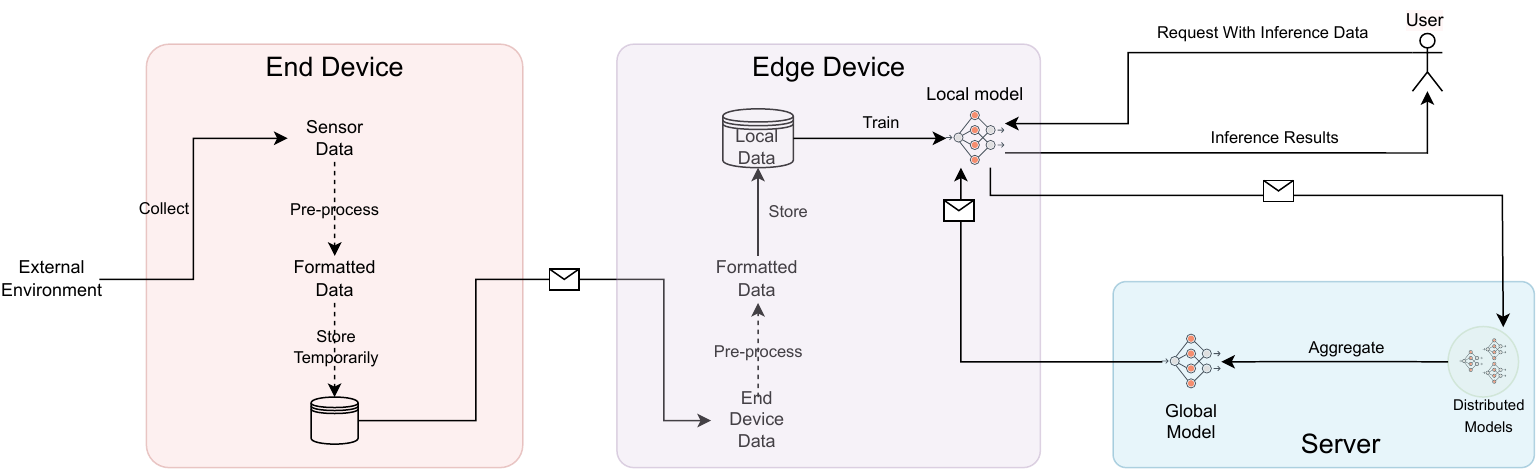}
  \end{center}
  \caption{An end-to-end \glsfmtshort{abb:ml} training process on the edge.}%
  \label{fig:edge_ml_process}
\end{figure*}

Having introduced both \glsxtrlong{abb:ml} in \Cref{sub:general_notions_of_ml}
and \glsxtrlong{abb:ec} in \Cref{sub:edge_computing},
we dedicate this section to discussing their combination.
We are particularly interested in \gls{abb:dl} on the edge,
since \glspl{abb:dnn} are both very popular, and easy to parallelize.
In all generality, the process of training a \gls{abb:ml} models on the edge
is depicted in \Cref{fig:edge_ml_process}.
End devices collect data from their environment,
optionally preprocessing or temporally storing it,
before sending it to an edge node.
The edge node aggregates, processes, and stores the data,
before training its local model on it,
possibly invoking the server for orchestration or heavy computations.
The server manages the training process
and performs the computations delegated to it by the edge nodes.
During each step of this process,
edge nodes can respond to client inference requests
using their current local model.

The number of proposed techniques to train \gls{abb:ml} models on the edge
is beyond what can be covered in a single section.
Therefore, we focus on two of the most popular techniques:
distributed learning and federated learning,
directing the reader's attention to~\cite{khouasTrainingMachineLearning2024}
for a review of other techniques.

\subsubsection{Distributed learning}\label{ssub:background:distl}

The process of training \gls{abb:ml} models by solving \Cref{eq:parametric_erm}
is a computationally intensive one, particularly, when \(n\) is large
(which needs to be the case for \gls{abb:erm} to be viable).
While this is a major inconvenience for cloud-based centralized \gls{abb:ml},
it is a complete deal-breaker for performing \gls{abb:ml} on resource-constrained edge devices.
The naive solution of designating a single powerful node to perform the training
suffers from the same drawback as all centralized computing: inefficient use of resources,
leading to unnecessarily high energy consumption and long training times%
~\cite{wangEdgeAIConvergence2020}.

The solution for the inadequacies of centralized computing is already well-known:
preform the computation in parallel on multiple devices.
Applying this solution to \gls{abb:ml} gives rise to \emph{distributed learning}%
~\cite{wangEdgeAIConvergence2020}.
Four main approaches exist for splitting the training workload
among devices, namely,
\begin{enumerate*}[label=(\arabic*)]
  \item \emph{data parallelism},
        in which each device is assigned a chunk of the dataset,
        on which it is responsible for training the model;
  \item \emph{model parallelism},
        in which the model is divided among devices,
        each of which is responsible for training a part of it
        on the entire dataset;
  \item \emph{hybrid parallelism},
        a combination of the previous two;
        applying one to part of the model and the other to the rest; and
  \item \emph{pipeline parallelism},
        which performs different stages of the training process
        concurrently on different batches of data%
\end{enumerate*}
~\cite{dehghaniDistributedMachineDistributed2023,wangEdgeAIConvergence2020}.

\begin{remark}
  The term \emph{distributed learning} is ambiguous,
  used by some authors~\cite{lyuCollaborativeFairnessFederated2020}
  as a synonym for \glsxtrlong{abb:fl},
  or the even more ambiguous \emph{collaborative learning}.
  Our use of the term is taken from%
  ~\cite{wangEdgeAIConvergence2020,dehghaniDistributedMachineDistributed2023},
  where it simply means dividing the computational workload among multiple devices
  for the sake of efficiency.
\end{remark}

From a \gls{abb:dq} perspective,
distributed learning is indistinguishable from centralized learning.
In all distributed learning approaches, the global dataset is public within the network.
This makes it easy to at least detect data quality issues.
The same cannot be said for \glsxtrlong{abb:fl},
which we will discuss in the next section.

\subsubsection{Federated learning}\label{ssub:background:fl}

\Glsxtrfull{abb:fl} is a collaborative framework for training \gls{abb:ml} models
using data and computational resources from different devices
without compromising the privacy of their local data%
~\cite{wangEdgeAIConvergence2020,xiaSurveyFederatedLearning2021}.
It allows multiple devices to collaborate on the training of a \gls{abb:nn}%
~\cite{mcmahanCommunicationEfficientLearningDeep2017}.
These devices can be wildly different in terms of hardware and software,
and their datasets are often incomparable in terms of size, diversity, and distribution%
~\cite{xiaSurveyFederatedLearning2021}.
In so-called \emph{centralized} \gls{abb:fl}%
~\cite{naikIntroductionFederatedLearning2024},
nodes are partitioned into two subsets (see \Cref{fig:edge_ml_process}):
\begin{enumerate*}[label=(\roman*)]
  \item \emph{clients}, which are the nodes that contribute their data to the training process and
  \item \emph{servers}, which are the nodes that orchestrate the training process%
        ~\cite{xiaSurveyFederatedLearning2021}.
\end{enumerate*}

\paragraph{Formulation of the problem}\label{par:background:fl:fomulation}

In federated learning, the target distribution \(\Pr_{(X, Y)}\)
is a \emph{mixture} of distributions.
A set of \(N\) clients, which we assimilate to the set of indices \([N]\) is given,
along with corresponding weights
\((\alpha_{1}, \alpha_{2}, \dots, \alpha_{N}) \in \triangle^{N}\),
and datasets \(\xi_{1}, \xi_{2}, \dots, \xi_{N}\),
where the elements of \(\xi_{i}\) are \gls{abb:iid} random pairs
sampled from the \emph{local} distribution \(\Pr_{(X_{i}, Y_{i})}\).
The so-called \emph{global} distribution \(\Pr_{(X, Y)}\) is then defined as
\begin{equation}
  \Pr_{(X, Y)} \coloneqq \sum_{i=1}^{N} \alpha_{i} \Pr_{(X_{i}, Y_{i})}.
\end{equation}
Given a common loss function \(\ell\),
the \emph{local} risk \(\mathcal{R}_{i}\) and empirical risk \(\hat{R}_{\xi_{i}}\)
of client \(i\) are defined analogously to \Cref{eq:risk,eq:empirical_risk},
with respect to local distribution
\(\Pr_{(X_{i}, Y_{i})}\) and sample \(\xi_{i}\) respectively.
The same is true for their \emph{global} counterparts \(\mathcal{R}\) and \(\hat{R}_{\xi}\),
but with respect to the global distribution and sample
\(\xi \coloneqq \bigcup_{i=1}^{N} \xi_{i}\)
respectively,
and one easily verifies that they can be expressed as convex combinations of the local versions
with coefficients proportional to sample sizes.
Federated learning is simply the problem of minimizing the global empirical risk,
which we also refer to as \gls{abb:ferm}.

\paragraph{The training process}\label{par:background:fl:training}

Limiting our attention to centralized \gls{abb:fl} with a single server,
a general algorithm for \gls{abb:fl} is given below~\cite{%
  mcmahanCommunicationEfficientLearningDeep2017,%
  xiaSurveyFederatedLearning2021,%
  wangEdgeAIConvergence2020%
}.
In the first stage, the server initializes the global model \(\theta^{(0)}\)
either randomly or using a pretrained parameter vector.
Next, the server and clients iteratively alternate between local training and global aggregation
for multiple \emph{rounds} until a stopping criterion is met.
At the beginning of round \(t\), the server selects a subset \(S_{t}\) of clients
according to some selection probability distribution,
to whom it broadcasts the current global model \(\theta^{(t-1)}\).
In the local training step each client \(i\in S_{t}\)
computes a local model \(\theta^{(t)}_{i}\) by solving \gls{abb:erm} locally:
\begin{equation}
  \theta^{(t)}_{i} = \argmin_{\theta \in \Theta} \hat{R}_{\xi_{i}}(\theta).
\end{equation}
This is usually done by running gradient descent on \(\hat{R}_{\xi_{i}}\)
with \(\theta^{(t-1)}\) as initialization
\begin{equation}
  \theta^{(t)}_{i} = \theta^{(t-1)} - \eta_{t}\nabla \hat{R}_{\xi_{i}}\left(\theta^{(t-1)}\right),
\end{equation}
and a learning rate \(\eta_{t}\), assigned by the server.
Each client then sends its local model \(\theta^{(t)}_{i}\) to the server,
where aggregation is performed
\begin{equation}
  \theta^{(t)} = A\left( \left\{ \theta^{(t)}_{i} \middle| i \in S_{t} \right\} \right)
\end{equation}
using a \gls{abb:gar} \(A\), which maps subsets of \(\Theta\) to elements of \(\Theta\).
A simple example of a \gls{abb:gar} is the convex combination i.e.,
\begin{equation}\label{eq:fedavg}
  \theta^{(t)} = \frac{1}{\sum_{i\in S_{t}}\alpha_{i}}
  \sum_{i\in S_{t}} \alpha_{i} \theta^{(t)}_{i},
\end{equation}
in which case the algorithm is called \gls{abb:fedavg}~\cite{mcmahanCommunicationEfficientLearningDeep2017}.
The global model \(\theta^{(T)}\) of the final round \(T\)
is the output of the algorithm.

\paragraph{Discussion}\label{par:background:fl:discussion}

Unlike distributed learning, \gls{abb:fl} is fundamentally different from centralized learning.
One of the starkest differences is that \gls{abb:fl} does not require data sharing,
which makes collaboration easier, but also, and crucially,
means that it does not reveal any information about local data,
except for what can be inferred from the local models (see \Cref{sub:privacy:model-reveals}).
This privacy guarantee is one of the main selling points of \gls{abb:fl},
and indeed one of the motivations cited for its introduction%
~\cite{mcmahanCommunicationEfficientLearningDeep2017}.
A second point of contrast is that \gls{abb:fl} does not rely on the \gls{abb:iid} assumption.
As the reader may have noticed, in formulating the \gls{abb:ferm} problem,
we only assumed \emph{locally \gls{abb:iid}} data.

These characteristics make \gls{abb:fl} a very attractive framework
for doing \gls{abb:ml} on the edge, justifying its wide adoption for many edge use cases
like healthcare, \gls{abb:iov}, and recommendation systems%
~\cite{wangEdgeAIConvergence2020,xiaSurveyFederatedLearning2021}.
Simultaneously, these same characteristics make \gls{abb:fl} meaningfully different
from both distributed learning and centralized learning from a data quality perspective,
which creates problems and opportunities
that are unique to distributed environments like the edge.
For this reason, we will focus on \gls{abb:fl} in the rest of this paper,
using it interchangeably with edge \gls{abb:ml}.

\section{Data quality for machine learning}\label{sec:data_quality}

\Glsxtrfull{abb:dq} is the central focus of this survey.
It is an extremely broad topic, not only in \gls{abb:ml} but also in
data science, data engineering, database management, information systems,
and multiple other adjacent disciplines,
as evidenced by the numerous works on the subject%
~\cite{%
  xuDataQualityMatters2023,%
  guptaDataQualityMachine2021,%
  olsonDataQualityAccuracy2003,%
  mahantiDataQualityDimensions2019,%
  renggliDataQualityDrivenView2021,%
  camachoQualityQualityOut2023,%
  gudivadaDataQualityConsiderations2017%
}
and the equally numerous approaches these works take to study it.
In this section, we explore \gls{abb:dq} for \gls{abb:ml} data
as it is covered in the existing literature,
with particular focus on the questions of:
\begin{enumerate*}[label=(\roman*)]
  \item what is data quality? (i.e., how it is defined),
  \item what characterizes a good dataset? (data quality dimensions), and
  \item how is data quality ensured? (existing solutions).
\end{enumerate*}
As such, this section is divided into three parts,
each corresponding to one of the above questions (in order).

\subsection{What is data quality?}\label{sub:what_is_data_quality}

A great body of research has been dedicated to making sense of the concept of data quality,
producing an equally great number of views on how to define it.
Among the few common threads between a majority of these works are the following two observations:
\begin{enumerate*}[label=(\arabic*)]
  \item data quality is context-dependent, i.e.,
        rather than being a function of the dataset alone,
        it is a function of the entire context of its use, and
  \item data quality is multidimensional, i.e.,
        multiple aspects of data must be considered when assessing it.
\end{enumerate*}
The first observation is demonstrated by the convergent pragmatic definition of data quality
as ``fitness for use''~\cite{sidiDataQualitySurvey2012},
``suitability for business purposes''~\cite{hagendorffLinkingHumanMachine2021},
or ``the degree to which the data of interest fulfills given requirements''%
~\cite{hassensteinDataQualityConcepts2022}.
The second is supported by the overwhelming trend of using multiple properties of a dataset
to assess its quality~\cite{%
  budachEffectsDataQuality2022,%
  mahantiDataQualityDimensions2019,%
  sidiDataQualitySurvey2012%
}.

For the purpose of this work, we adhere to both of the above-stated observations.
The first one boils down to the fact that
assigning quality to a dataset ---or anything else for that matter---
is a normative judgment, and therefore cannot be done without an underlying objective.
In the context of \gls{abb:ml}, and particularly for risk minimization,
the objective is to minimize the risk \(\mathcal{R}\).
Although this is usually approximated by \gls{abb:erm},
minimizing \(\mathcal{R}\) remains the ultimate goal, and thus the measure of quality.
This implies the following principle, which we will use
to guide us to a definition of data quality.

\begin{principle}\label{princple:performance_is_dq}\ \\
  A good dataset \(\xi\) can be defined as one that (when used for \gls{abb:erm})
  produces a good model, which in turn can be defined
  as a model \(\hat{\theta}_{\xi}\) with low \emph{true risk}
  \(\mathcal{R}\left( \hat{\theta}_{\xi} \right)\).
\end{principle}

Echoes of this principle can be found in the literature,
for example in~\cite{danilovDataQualityEstimation2023,camachoQualityQualityOut2023}.
Its main advantage is that it reduces the nebulous and poorly defined concept of data quality
to the comparatively well-defined concept of model quality,
measured by the completely unambiguous true risk.
The second observation can be captured by the use of multiple properties,
which together imply that \(\mathcal{R}\left( \hat{\theta}_{\xi} \right)\)
is small.
In the following, we will explore some of these properties.

\begin{forest}  for tree={%
  grow'=east,
  growth parent anchor=west,
  parent anchor=east,
  child anchor=west,
  calign=center,
  edge path={%
  \noexpand\path[\forestoption{edge},->, >={latex}]
  (!u.parent anchor) -- + (10pt,0pt) |-  (.child anchor) \forestoption{edge label};
  }
  }
  [Data quality dimensions, parent, l sep=10mm,
  [\nameref{sec:independence}, child]
    [\nameref{sec:equidistribution}, child]
    [\nameref{sec:noise}, child, l sep=10mm,
      [\nameref{par:noise:taxonomy:label-set}, gchild]
        [\nameref{par:noise:taxonomy:dependence}, gchild]%
    ]
    [\nameref{sec:fairness}, child, l sep=10mm,
      [\nameref{sub:fairness:social}, gchild, l sep=10mm,
          [\nameref{ssub:fairness:social:group}, g2child]
            [\nameref{ssub:fairness:social:individual}, g2child]
            [\nameref{ssub:fairness:social:subgroup}, g2child]%
        ]
        [\nameref{sub:fairness:coop}, gchild, l sep=10mm,
          [\nameref{sub:fairness:coop:egalitarian}, g2child]
            [\nameref{sub:fairness:coop:proportional}, g2child]
            [\nameref{sub:fairness:coop:core}, g2child]%
        ]
    ]
    [\nameref{sec:privacy}, child, l sep=10mm,
      [\nameref{ssub:privacy:attacks:membership}, gchild]
        [\nameref{ssub:privacy:attacks:representative}, gchild]
        [\nameref{ssub:privacy:attacks:reconstruction}, gchild]
        [\nameref{ssub:privacy:attacks:property}, gchild]
    ]
    [\nameref{sec:trust}, child, l sep=10mm,
      [\nameref{ssub:trust:attacks:model}, gchild]
        [\nameref{ssub:trust:attacks:data}, gchild]
    ]
  ]
\end{forest}
\subsection{Data quality-aware machine learning}\label{sub:dq:solutions}

Having examined the existing literature on defining data quality,
we now turn our attention to the question of how to ensure it \gls{abb:ml} systems.
While the definition and dimensions of \gls{abb:dq}
are relatively identical between centralized and federated learning,
many of the solutions to \gls{abb:dq} issues are specific to \gls{abb:fl}.
As such, we will distinguish between the two cases where necessary.

\begin{figure}[hbt]
  \begin{center}
    \resizebox{\linewidth}{!}{\begin{forest} for tree={%
  grow'=east,
  growth parent anchor=west,
  parent anchor=east,
  child anchor=west,
  calign=center,
  edge path={%
  \noexpand\path[\forestoption{edge},->, >={latex}]
  (!u.parent anchor) -- + (10pt,0pt) |-  (.child anchor) \forestoption{edge label};
  }
  }
  [\nameref{sub:dq:solutions}, parent, l sep=10mm,
  [\nameref{ssub:dq:solution:pre}, child, l sep=10mm,
      [Data cleaning, gchild]
        [Data augmentation, gchild]
        [Data transformation, gchild]%
    ]
    [\nameref{ssub:dq:solution:in}, child, l sep=10mm,
      [Learning algorithm modification, gchild, l sep=10mm,
          [Objective modification, g2child]
            [Hypothesis class modification, g2child]%
        ]
        [\gls{abb:fl} protocol modification, gchild, l sep=10mm,
          [\gls{abb:gar} modification, g2child]
            [Client selection modification, g2child]%
        ]%
    ]
    [\nameref{ssub:dq:solution:post}, child]
  ]
\end{forest}}
  \end{center}
  \caption{The different approaches to mitigating \gls{abb:dq} issues in \gls{abb:ml}.}%
  \label{fig:solutions-taxonomy}
\end{figure}
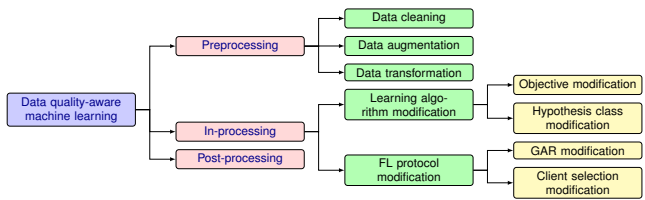

Broadly speaking, there are three approaches to mitigating data quality issues in \gls{abb:ml},
depending on the stage of the \gls{abb:ml} pipeline at which they are applied~\cite{%
  mehrabiSurveyBiasFairness2022,%
  whangDataCollectionQuality2023%
}
(or equivalently, the component of the \gls{abb:ml} system they target):
\begin{enumerate*}[label=(\roman*)]
  \item \emph{preprocessing},
        which is applied before training, with the goal \emph{enhancing} the data,
  \item \emph{in-processing},
        which is applied during training,
        with the goal of producing a \emph{robust} model, and
  \item \emph{post-processing},
        which is applied after training, (i.e., during inference),
        with the goal of \emph{correcting} the model's predictions, or more generally,
        removing inaccurate, discriminatory, dangerous, or otherwise inadequate predictions.
\end{enumerate*}
This distinction is used by~\cite{mehrabiSurveyBiasFairness2022,ferraguigSurveyBiasMitigation2021}
as a cross-domain classification of \gls{abb:ai} fairness approaches,
but we will adopt it here for the more general problem of \gls{abb:dq}.
Within each of these three broad families,
approaches can be further distinguished (see~\Cref{fig:solutions-taxonomy}).

\subsubsection{Preprocessing}\label{ssub:dq:solution:pre}

Preprocessing is a straightforward strategy to addressing \gls{abb:dq} issues.
Quite simply, it assumes that data problems should be addressed on the data level.
Generally speaking, a preprocessing technique operates on
a possibly low-quality dataset \(\xi\),
and produces a higher-quality dataset \(\xi^{\prime}\),
which can be a subset, superset, or a more general function of \(\xi\).
These cases are respectively known as \emph{data cleaning},
\emph{data augmentation}, and \emph{data transformation}~\cite{%
  ramirez-gallegoSurveyDataPreprocessing2017,%
  wernerdevargasImbalancedDataPreprocessing2023,%
  youHandlingMissingData2020%
}.

\subsubsection{In-processing}\label{ssub:dq:solution:in}

This family is the richest of the three,
because training is usually the most elaborate part of the \gls{abb:ml} pipeline.
This is particularly the case when \gls{abb:fl} is used.
It is therefore warranted to consider a subclassification
of in-processing methods, based on what element of the training process they modify.

An instance of the \gls{abb:erm} is given by the risk \(\mathcal{R}\)
and the hypothesis class \(\mathcal{H}\), which in turn,
in \gls{abb:dl}, is determined by a neural architecture,
and a parameter space \(\Theta\).
Each of these components can be modified to incorporate \gls{abb:dq} considerations.
The risk can be modified by
changing the loss function~\cite{zhangGeneralizedCrossEntropy2018},
replacing the expectation with a different aggregation function%
~\cite{linDistributionallyRobustOptimization2022},
or adding a \gls{abb:dq}-dependent regularization term%
~\cite{zhouFederatedLabelNoiseLearning2024}.
The hypothesis class can be modified by changing the neural architecture%
~\cite{goldbergerTrainingDeepNeuralnetworks2022}
or restricting the parameter space (typically by adding constraints)%
~\cite{zafarFairnessConstraintsMechanisms2017}.

\Glsxtrlong{abb:fl} adds another layer of complexity to the picture
by adding two steps to the training process: client selection and aggregation.
Both of these steps can be targeted for modification.
The client selection process can be modified by
assigning higher probability to clients with higher-quality data%
~\cite{taikDataQualityBasedScheduling2021,liSamplelevelDataSelection2021}.
The aggregation process can be modified by
using a robust \gls{abb:gar}~\cite{el-mhamdiStrategyproofnessGeometricMedian2023}.

\subsubsection{Post-processing}\label{ssub:dq:solution:post}

Post-processing is applied after the model has been trained (i.e., at inference time).
Out of the three families, post-processing is the only one
that is universally applicable.
In the black-box setting, that is when the data and the model
are inaccessible, unknown, or unmodifiable,
preprocessing and in-processing are not possible.
Post-processing, however, can still be applied~\cite{%
  bolukbasiManComputerProgrammer2016,%
  hardtEqualityOpportunitySupervised2016,%
  petersenPostprocessingIndividualFairness2021%
}.
We note that to the best of our knowledge, no edge-specific
post-processing techniques have been proposed in the literature.
Therefore, we will not elaborate on this family further.

\section{Statistical independence}\label{sec:independence}

The so-called \gls{abb:iid} assumption is the bedrock
for much of the classical theory (and practice) of \gls{abb:ml}
~\cite{vapnikNatureStatisticalLearning2010,liuOutOfDistributionGeneralizationSurvey2023}.
While convenient for theoretical analysis,
the \gls{abb:iid} assumption is not a realistic model of real-world data%
~\cite{liuOutOfDistributionGeneralizationSurvey2023},
an observation that is even more salient in \gls{abb:fl} on the edge%
~\cite{%
  xiaSurveyFederatedLearning2021,%
  wangNonIIDDataRebalancing2021,%
  liConvergenceFedAvgNonIID2020%
}.
In general, the \gls{abb:iid} assumption can fail in two ways,
the data can be either
\begin{enumerate*}[label=(\roman*)]
  \item  \emph{dependent}, or
  \item it can fail to be equidistributed according to
        the true distribution \(\Pr_{(X, Y)}\).
\end{enumerate*}
In this section, we discuss the first of these two issues,
leaving the second for \Cref{sec:equidistribution,sec:noise}.

\subsection{Machine learning on dependent data}\label{sub:independence:ml}

The majority of works in the literature on learning from dependent data%
~\cite{%
  zouGeneralizationPerformanceERM2009,%
  guoClassificationNoniSampling2011,%
  ziminLearningDependentData2018,%
  royEmpiricalRiskMinimization2021%
}
relies on the theory of stochastic processes.
The typical approach is to replace the independence part
of the \gls{abb:iid} assumption with a weaker condition like \emph{mixing}%
~\cite{guoClassificationNoniSampling2011}.
This is sometimes accompanied by a strengthening
of the \emph{identically distributed} part to a \emph{stationarity condition}%
~\cite{royEmpiricalRiskMinimization2021,dundarLearningClassifiersWhen}.
The alternate set of assumptions is then used to derive generalization bounds
analogous to those available in the \gls{abb:iid} case.

Notable deviations from this approach include the work of
\citet{lauerUniformRiskBounds2023} and \citet{dundarLearningClassifiersWhen}.
The former presents derivations of classical generalization bounds
such as those based on the \gls{abb:vc} dimension~\cite{vapnikStatisticalLearningTheory1998},
or the Rademacher complexity of the hypothesis class,
without assuming independence, nor replacing it with another condition.
The latter on the other hand, takes a more practical approach,
leveraging (actual or assumed) information about the structure of the dependence
to train linear classifiers on dependent data.

\subsection{Dependent data in federated learning}\label{sub:independence:fl}

The problem of data dependence in \gls{abb:fl} on edge devices remains largely unexplored.
In our review of the literature, we could not find any edge or \gls{abb:fl}-specific works
that address the issue of dependent data.
This is not indicative of a general lack of interest in
the effect of \gls{abb:iid} violations on \gls{abb:fl},
which is one of the most active areas of research in the field,
even more so in the case of edge \gls{abb:fl}~\cite{xiaSurveyFederatedLearning2021}.
Most of these works, however, disregard the question of dependence,
in favor of focusing on distribution mismatch~\cite{%
  wangNonIIDDataRebalancing2021,%
  liConvergenceFedAvgNonIID2020,%
  leeConvergenceFederatedLearning2023%
}.
The absence of works on dependent data in \gls{abb:fl},
coupled with the localized nature of the edge environment,
which can lead to strong dependence between the data points
of one device or a set of similar devices,
and the difficulty of detecting such dependence due to privacy constraints,
suggests that this is a promising direction for future research.

\section{Attribute skew}\label{sec:equidistribution}

The second (and arguably the most important) component
of the \gls{abb:iid} assumption is for all random pairs in the sample
to be drawn from the true distribution.
If this is not the case, the empirical risk is not guaranteed to converge to the true risk,
negating guarantees on the consistency of \gls{abb:erm}.
This can be due to a mismatch in the marginal distribution \(\Pr_{X}\)
or the conditional distribution \(\Pr_{Y \mid X}\).
In this section, we will discuss the first of these two possibilities,
discussing the second in the next section.

When multiple edge devices participate in \gls{abb:fl},
their local datasets are often heterogeneous~\cite{abrehaFederatedLearningEdge2022}.
For example, two wearable devices owned by people of different ages
and physical activity levels are extremely unlikely to have similarly distributed datasets.
Therefore, aggregating models trained on these datasets
will not necessarily yield an unbiased estimate of the model trained on their union%
~\cite{zhaoFederatedLearningNonIID2018}.

Addressing this issue is one of the most active areas of research in \gls{abb:fl}.
The corpus of literature on the topic is expansive, diverse, and rapidly growing,
covering a wide range of topics from theoretical guarantees%
~\cite{liConvergenceFedAvgNonIID2020,leeConvergenceFederatedLearning2023}
to practical solutions%
~\cite{wangNonIIDDataRebalancing2021,zhaoFederatedLearningNonIID2018},
and empirical evaluation thereof%
~\cite{wangWhyBatchNormalization2023,wangBatchNormalizationDamages2023}.
In this section, we provide a representative sample of the work in this area,
directing the reader's attention to the surveys by~\cite{%
  zhuFederatedLearningNonIID2021,%
  maStateoftheartSurveySolving2022,%
  luFederatedLearningNonIID2024%
}
for a more comprehensive overview.

\subsection{Practical solutions and empirical evaluation}\label{sub:equidistribution:practical}

Multiple authors have proposed, implemented, and tested novel algorithms
for performing \gls{abb:fl} on the edge with non-\gls{abb:iid} data.
Different surveys of the literature~\cite{%
  zhuFederatedLearningNonIID2021,%
  maStateoftheartSurveySolving2022,%
  luFederatedLearningNonIID2024%
}
introduced different categorizations of these methods.
Two recurring categories are data-based and algorithm-based methods,
corresponding roughly to preprocessing and in-processing methods in our taxonomy.
Both categories contain a variety of methods,
a subset of which we will discuss in this section.

\subsubsection{Data-based methods}\label{ssub:equidistribution:practical:data}

The motivating intuition behind data-based methods is that
modifying the local datasets to make them more homogeneous
will lead to better performance when training a global model.
Three main approaches to achieving this have been proposed:
data selection, data augmentation, and data selection%
~\cite{%
  zhuFederatedLearningNonIID2021,%
  maStateoftheartSurveySolving2022%
}.

\paragraph{Data sharing}\label{par:equidistribution:practical:sharing}

\citet{zhaoFederatedLearningNonIID2018} observed that the accuracy of models
trained through \gls{abb:fl} can decrease by up to \(55\%\) if the data is not \gls{abb:iid}.
To address this concern, they propose maintaining a small global shared dataset on the server,
a portion of which is shared with each edge device to augment their local datasets.
They find that a shared dataset as small as \(5\%\) of the total data
can improve the accuracy by as much as \(30\%\).
Later work by \citet{yoshidaHybridFLWirelessNetworks2020} combine this
approach with data selection to achieve \(13.5\%\) higher accuracy still.
\citet{tianAsynchronousFederatedLearning2021} applied this approach
to asynchronous edge \gls{abb:fl} using a shared dataset size of \(5-10\%\),
improving both the accuracy and the convergence speed of the model.

\paragraph{Data augmentation}\label{par:equidistribution:practical:augmentation}

\citet{abayMitigatingBiasFederated2020} tackle the problem of non-\gls{abb:iid} data
by augmenting the local datasets to match the global distribution.
More concretely, the server uses client-reported statistics on the label distribution
which it uses to coordinate the clients to generate augmentations for rare classes.
\citet{jeongCommunicationEfficientOnDeviceMachine2023}
on the other hand, propose an approach based on a \gls{abb:gan}
trained by the server on shared data,
which it distributes to the clients for local data augmentation.

\paragraph{Data selection}\label{par:equidistribution:practical:selection}

\citet{raiClientSelectionFederated2022} propose a client selection algorithm
based on a single metric called the \emph{irrelevance score}, which measures,
among other things, deviation from being \gls{abb:iid}.
\cite{wangOptimizingFederatedLearning2020} on the other hand,
propose a \(q-\)learning based client selection approach to increase convergence speed
with non-\gls{abb:iid} data.

\subsubsection{Algorithm-based methods}\label{ssub:equidistribution:practical:algorithm}

Algorithm-based methods are in-processing methods that amend the training algorithm
to make it more robust to non-\gls{abb:iid} data.
Due to the complex and multi-component nature of the \gls{abb:fl} training process,
these methods are more varied and technically sophisticated than data-based methods.

\paragraph{Meta-learning}\label{par:equidistribution:practical:meta}

\citet{jiangImprovingFederatedLearning2023} propose a model agnostic meta-learning approach%
~\citet{finnModelAgnosticMetaLearningFast2017}
to simultaneously improve personalization and convergence speed.
Another work by \citet{liFederatedMetalearningSpatialtemporal2022}
applies meta-learning to spatial and temporal data which is innately non-\gls{abb:iid}.
In a different vein, \citet{zhangFedPDFederatedLearning2021}
propose a meta-algorithm based on the primal-dual algorithm.
Their method produces strong convergence guarantees
even with non-\gls{abb:iid} data and non-convex objectives.
Furthermore, it achieves optimal communication efficiency,
requiring only \(\bigO(1)\) communication rounds in the \gls{abb:iid} case.

\paragraph{Multitask learning}\label{par:equidistribution:practical:multi}

Personalization can be cast as a multitask learning problem%
~\cite{zhuFederatedLearningNonIID2021}.
Works that have addressed \gls{abb:fl} on the edge with non-\gls{abb:iid} data
from this perspective include~\cite{smithFederatedMultiTaskLearning2017},
which considers each edge device as a separate task,
\cite{corinziaVariationalFederatedMultiTask2021},
which treats the edge system as a Bayesian network,
to handle strong \gls{abb:iid} violations,
and \cite{sattlerClusteredFederatedLearning2021},
which leverages the geometry of the loss function to
combine subsets of clients into clusters with better overall distributions.

\subsection{Theoretical analysis}\label{sub:equidistribution:theoretical}

The majority of literature on theoretical \gls{abb:fl} with non-\gls{abb:iid}
data focuses on convergence.
For example, \citet{liConvergenceFedAvgNonIID2020}
prove convergence theorems and rates for \gls{abb:fedavg} with non-\gls{abb:iid} data
under certain conditions on the smoothness and convexity of the loss function,
and the boundedness of the gradients.
\citet{leeConvergenceFederatedLearning2023} also provide convergence guarantees
for federated regression assuming similar rather than identical local distributions.

\section{Label noise}\label{sec:noise}

Label noise, or \emph{labeling error}, is a term used to qualify
any situation where the label present in a dataset is different from the target%
~\cite{frenayClassificationPresenceLabel2014}.
This can be due to human error,
communication and storage defects~\cite{baccourPervasiveAIIoT2022}
(very common on the edge),
and deliberate mislabeling~\cite{%
  frenayClassificationPresenceLabel2014,%
  songLearningNoisyLabels2023%
}.
It is not difficult to see how the presence of noise can be problematic,
particularly for highly expressive models such as \glspl{abb:dnn},
which are susceptible to overfitting.

\subsection{Types of label noise}\label{sub:noise:taxonomy}


In order to reason about label noise, it is usually modeled as
a stochastic process acting on the dataset, i.e.,
instead of a clean dataset
\[\xi=\left\{(X_{1}, Y_{1}), (X_{2}, Y_{2}), \dots, (X_{n}, Y_{n})\right\},\]
we have a noisy dataset
\[
  \tilde{\xi} = \left\{
  (X_{1}, \tilde{Y}_{1}), (X_{2}, \tilde{Y}_{2}), \dots, (X_{n}, \tilde{Y}_{n})\right
  \}
\]
where the observed label \(\tilde{Y}_{i}\) is a random variable
over \(\tilde{\mathcal{Y}} \subset \mathcal{Y}\),
which may or may not depend on \((X_i, Y_i)\)%
~\cite{hanSurveyLabelnoiseRepresentation2021}.
The properties of \(\tilde{Y}_i\) can be used as a basis
for classification of label noise types.

\paragraph{Label set-based classification}\label{par:noise:taxonomy:label-set}

One property of \(\tilde{Y}_i\) that can be used to classify label noise
is the set of modalities  \(\tilde{\mathcal{Y}}\).
Depending on whether \(\tilde{\mathcal{Y}} = \mathcal{Y}\) or
\(\tilde{\mathcal{Y}} \subsetneq \mathcal{Y}\),
we speak of \emph{closed-set} (or \emph{in-distribution}) noise,
and \emph{open-set} (or \emph{out-of-distribution}) noise~\cite{%
  wanUnlockingPowerOpen2023,%
  albertAddressingOutofDistributionLabel2022%
}.
The former means that we retain all the classes present in the original dataset,
while the latter means that the noise causes loss of knowledge of some classes.

\paragraph{Dependence-based classification}\label{par:noise:taxonomy:dependence}

Another relevant property for label noise classification is
the dependence of \(\tilde{Y}_i\) on \((X_{i}, Y_{i})\),
that is, the conditional distribution \(\Pr_{\tilde{Y}_{i} | (X_{i}, Y_{i})}\).
Four cases are possible for this distribution:
\begin{enumerate*}[label=(\alph*)]
  \item it can reduce to \(\Pr_{\tilde{Y}_{i} \mid Y_{i}}\),
        in which case \(\tilde{Y}_i\) the noise is said to be \emph{class-dependent}%
        ~\cite{alganLabelNoiseTypes2020},
  \item to \(\Pr_{\tilde{Y}_{i} \mid X_{i}}\),
        in which case the noise is \emph{instance-dependent}%
        ~\cite{hanSurveyLabelnoiseRepresentation2021},
  \item to \(\Pr_{\tilde{Y}_{i}}\), in which case the noise is completely random%
        ~\cite{frenayClassificationPresenceLabel2014},
  \item or it can be irreducible, depending on both \(X_i\) and \(Y_i\).
\end{enumerate*}
Each class of noise can be subdivided further,
and each is modeled and handled differently.
For a more detailed discussion of these types of noise,
we refer the reader to~\cite{%
  alganLabelNoiseTypes2020,%
  shanthiniTaxonomyImpactLabel2019,%
  hanSurveyLabelnoiseRepresentation2021,%
  liangReviewSurveyLearning2022%
}.

\subsection{Effects of label noise on machine learning}\label{sub:noise:effects}

Several studies have been conducted to confirm and quantify
the effects of label noise on \gls{abb:ml} models in diverse scenarios,
including deep supervised image classification~\cite{%
  alganImageClassificationDeep2021,%
  barryImpactDataQuality2023,%
  alganLabelNoiseTypes2020%
},
binary classification using tree-based models~\cite{shanthiniTaxonomyImpactLabel2019},
and federated training of \glspl{abb:svm} on the edge~\cite{zengNoiseUsefulExploiting2021}.

Most of these studies agree on the intuitive conclusion that
the presence of label noise is detrimental to the performance of \gls{abb:ml} models,
and further, that the size of the effect grows with the amount of noise present in the dataset%
~\cite{%
  alganImageClassificationDeep2021,%
  alganLabelNoiseTypes2020,%
  shanthiniTaxonomyImpactLabel2019%
}.
\citet{barryImpactDataQuality2023} even report a decrease in fairness
as label noise increases,
an observation that they make for \glspl{abb:dnn} and \glspl{abb:svm},
but not for naive Bayesian classifiers which they find to be robust to this effect.

However, other studies have found that label noise can be beneficial to \gls{abb:fl}.
\citet{zengNoiseUsefulExploiting2021} report that noisy communication channels
increase data diversity, thus, prioritizing clients with noisy channels in the selection process
can outperform random selection when training an \gls{abb:svm} on the MNIST dataset%
~\cite{dengMNISTDatabaseHandwritten2012}.
\citet{jafarigolParadoxNoiseEmpirical2023} present similarly positive results
for generalization, stability, and privacy
when training a \gls{abb:cnn} on the CIFAR-10 dataset%
~\cite{krizhevskyLearningMultipleLayers}
using \gls{abb:fl} with intentionally added label noise.

\section{Fairness}\label{sec:fairness}

Fairness is a critical ethical consideration in the design
of any decision-making process (automated or otherwise) that affects humans.
In the context of edge \gls{abb:fl},
we can distinguish between two types of fairness%
~\cite{vucinichCurrentStateChallenges2023},
broadly corresponding to the traditional notions
of \emph{outcome fairness} and \emph{process fairness}%
~\cite{grgic-hlacaDistributiveFairnessAlgorithmic2018}
from the literature on algorithmic fairness.
We refer to them as social fairness and cooperative fairness, respectively.
The latter is of particular interest in the context of \gls{abb:fl},
given its procedural divergence from traditional \gls{abb:ml}.
While social fairness is concerned with the model making fair predictions,
cooperative fairness is concerned with the training process producing a fair model.

\subsection{Social fairness}\label{sub:fairness:social}

A prerequisite for ensuring fair predictions is to have a clear definition of fairness.
The literature is teeming with attempts to formalize our intuitive notion of fairness,
leading to a plethora of definitions and metrics~\cite{weertsFairlearnAssessingImproving2023}.
\citet{mehrabiSurveyBiasFairness2022} provide a comprehensive survey of these notions,
dividing them into the three categories of group, individual, and subgroup fairness.
The first two of these categories directly aim to
capture a common intuition of fairness into a mathematical definition,
while the third aims to combine their strengths.
Group fairness formalizes the idea that
certain attributes of an individual, typically membership to a protected group
(e.g., ethnicity, gender, or religion),
should not influence the output of a fair model.
Individual fairness, on the other hand, formalizes the idea that
a fair model should behave similarly given inputs from similar individuals.
Within each of these categories,
methods can be grouped in different ways,
producing multiple taxonomies that are very deep and very wide.
The reader is referred to~\cite{%
  mehrabiSurveyBiasFairness2022,%
  castelnovoClarificationNuancesFairness2022,%
  vermaFairnessDefinitionsExplained2018%
}
for a more detailed discussion.

\subsubsection{Group fairness}\label{ssub:fairness:social:group}

Given a (categorical) sensitive attribute \(A = g(X)\), and a model \(f \in \mathcal{H}\)
a group fairness notion can be formulated as an independence requirement
between \(f(X)\) and \(A\)~\cite{castelnovoClarificationNuancesFairness2022}.
The simplest possible such requirement is for \(f(X)\) to be independent of \(A\):
\begin{equation}\label{eq:fairness:group:dp}
  f(X) \independent A,
\end{equation}
a notion referred to as \emph{demographic parity}%
~\cite{agarwalReductionsApproachFair2018,weertsFairlearnAssessingImproving2023},
which is problematic for many reasons,
not least of which is that it forbids the use
of the \emph{perfect} predictor \(f(X) = Y\),
if \(Y\) and \(A\) are correlated~\cite{%
  agarwalReductionsApproachFair2018,%
  castelnovoClarificationNuancesFairness2022%
}.
As such, more practical relaxations of it have been proposed.
They typically weaken the independence requirement
for example, by conditioning on a \emph{legitimate} variable \(h(X)\)
~\cite{corbett-daviesAlgorithmicDecisionMaking2017}:
\begin{equation}\label{eq:fairness:group:cdp}
  f(X) \independent A \mid h(X),
\end{equation}
or by conditioning on the true label \(Y\)~\cite{hardtEqualityOpportunitySupervised2016}:
\begin{equation}\label{eq:fairness:group:eodds}
  f(X) \independent A \mid Y,
\end{equation}
giving rise to the notions of \emph{conditional demographic parity}
and \emph{equalized odds} respectively.
The latter can be further relaxed in the binary classification setting
(\(\mathcal{Y} = \{0, 1\}\))
to only require equal true positive rates~\cite{doniniEmpiricalRiskMinimization2018},
a notion known as \emph{equal opportunity}~\cite{hardtEqualityOpportunitySupervised2016}.

\subsubsection{Individual fairness}\label{ssub:fairness:social:individual}

Similarity-based individual fairness notions are as diverse
as the underlying definitions of similarity.
In the context of fairness,
a reasonable definition of similarity is for two individuals to be similar
if they have identical non-sensitive attributes.
An individually fair model should therefore treat them identically,
as long as the sensitive attributes are not used in its training%
~\cite{grgic-hlacaCaseProcessFairness2016}.
This notion, referred to as \emph{fairness through unawareness}%
~\cite{castelnovoClarificationNuancesFairness2022,vermaFairnessDefinitionsExplained2018},
is too permissive, as non-sensitive attributes (like zip codes)
can be highly correlated with sensitive ones (like ethnicity).

The polar opposite of this notion is \emph{fairness through awareness},
which explicitly constrains the model to treat similar individuals similarly
instead of relying on ignorance.
To achieve this, \citet{dworkFairnessAwareness2012}
first formalize similarity between individuals
by assuming the input space \(\mathcal{X}\) is equipped with a metric \(d_{\mathcal{X}}\).
The similarity between outputs is formalized by considering that
a probabilistic model \(f\) that maps \(\mathcal{X}\)
to the set of probability measures over \(\mathcal{Y}\)
(mapping \(x\) to an estimate of \(\Pr_{Y \mid X = x}\))
instead of directly to \(\mathcal{Y}\).
The advantage of doing this is that probability measures are naturally
equipped with multiple metrics (collectively known as statistical distances),
any of which can be used to formalize similarity.
Fixing one such metric \(d_{\mathcal{Y}}\),
individual fairness can be defined as a continuity requirement on \(f\).
Most works~\cite{dworkFairnessAwareness2012,petersenPostprocessingIndividualFairness2021}
opt for the (rather strong) Lipschitz continuity requirement
\begin{equation}\label{eq:fairness:individual:lip}
  d_{\mathcal{Y}}\left(f(x), f(x^\prime)\right) \le \kappa d_{\mathcal{X}}(x, x^\prime),
\end{equation}
for some \(\kappa \ge 0\).

Yet another approach to individual fairness is to consider
that an individual is similar to its counterfactual self,
that is, the individual it would have been if the protected attribute were different.
This notion, referred to as \emph{counterfactual fairness}%
~\cite{kusnerCounterfactualFairness2017}
can be made rigorous using causal models,
and has the straightforward interpretation that
sensitive attributes do not causally influence the output of the model.

\subsubsection{Subgroup fairness}\label{ssub:fairness:social:subgroup}

The last family of fairness notions we will discuss is subgroup fairness.
This family is a stronger version of group fairness,
where multiple partitions of the population are considered.
For a set \(G \subset \powerset\left( \mathcal{X} \right)\) of protected groups,
we say that a model \(f\) satisfies \(\epsilon-\)statistical parity
with respect to \(G\)~\cite{kearnsPreventingFairnessGerrymandering2018}
if and only if, for all \(A \in G\), \(f\) satisfies approximate demographic parity
(where the tightness of the approximation is controlled by \(\epsilon\)).
Similarly, \(f\) satisfies \(\epsilon-\)false positive rate fairness
if it has similar false positive rates on \(A\) and its complement for all \(A \in G\)%
~\cite{kearnsPreventingFairnessGerrymandering2018,kearnsEmpiricalStudyRich2019}.

\subsection{Cooperative fairness}\label{sub:fairness:coop}

In addition to the traditional causes,
the \gls{abb:fl} protocol itself can be a source of unfairness%
~\cite{rafiFairnessPrivacyPreserving2024}.
The decentralized nature of the edge environment,
where data is distributed across multiple
---often very different--- devices,
with very little information being shared,
makes it extremely difficult to coordinate the learning process
in a way that is satisfactory to all clients.
This can in large part be attributed to the lack of
agreed-upon fairness metrics for this scenario~\cite{%
  rafiFairnessPrivacyPreserving2024,%
  lyuFairPrivacyPreservingFederated2020%
}.
Absent such metrics, defining what fair edge \gls{abb:fl} means is necessary.
In this section, we will introduce three fairness criteria proposed in the literature.

\subsubsection{Egalitarian fairness}\label{sub:fairness:coop:egalitarian}

Egalitarianism is one of the most straightforward notions of fairness
to consider in any cooperative setting.
Applying this to edge \gls{abb:fl} means that all clients benefit equally from the global model,
i.e., the global model must have (approximately) equal risk on all clients.
This notion can be formalized in multiple ways;
several works have explored egalitarian notions of fairness in \gls{abb:fl}
(see~\cite[Section 2.1]{donahueFairnessModelsharingGames2023}
and~\cite[Sections 1 and 2]{raychaudhuryFairnessFederatedLearning2022} for a list of references).

One way to formalize egalitarian fairness is to minimize the maximum risk across all clients.
A slight generalization of this, referred to as \emph{good intent fairness}, was proposed by%
~\citet{mohriAgnosticFederatedLearning2019}.
Instead of solving \gls{abb:ferm} with respect to a single \(\alpha\in\triangle^{N}\),
they minimize the maximum risk over all possible coefficients \(\alpha\in\Lambda\)
for some set \(\Lambda\subset\triangle^{N}\).
If \(\Lambda\) contains the vertices of \(\triangle^{N}\),
then the risk of the worst-performing client is minimized.

An alternative approach to egalitarian fairness
aims to exploit the notions of fairness introduced in \Cref{sub:fairness:social}.
For instance, applying demographic parity (or any social fairness notion)
with the sensitive attribute being the client ID gives a notion of fairness
that is applicable to edge \gls{abb:fl}.
Following this idea, \citet{zhangUnifiedGroupFairness2022}
propose the notion of \emph{unified group fairness} which aims to simultaneously minimize
accuracy disparity (defined as the standard deviation of the accuracy)
among clients and (potentially) other sensitive groups.

\subsubsection{Proportional fairness}\label{sub:fairness:coop:proportional}

A common criticism of egalitarian fairness is that it is agnostic to client contributions.
On top of being intuitively unfair, this framework induces undesirable incentives.
If the global model is dictated by the worst-performing client
(which is likely to have low-quality data),
other clients, particularly those with large amounts of high-quality data,
have nothing to gain by contributing.
Worse still, a malicious client (see \Cref{sub:trust:attacks})
can monopolize the training process by pretending to have low-quality data.
It is therefore necessary to consider other notions of fairness
that take into account the contributions of the clients.

A reasonable approach to this is to start with a notion of egalitarian fairness
and then weigh the clients according to their contributions.
\cite{donahueFairnessModelsharingGames2023}
apply this principle starting with \(\lambda-\)egalitarian fairness (\(\lambda \ge 1\)),
which is met when the global model \(\theta\) satisfies the condition
\begin{equation}\label{eq:fairness:proportional:λ}
  \frac{\mathcal{R}_{i}(\theta)}{\mathcal{R}_{j}(\theta)}\le \lambda
\end{equation}
for all clients \(i, j\).
By multiplying the risk of each client by a factor that depends on its contribution,
a simplistic quantification of which is the size of its dataset,
we get the notion of proportional fairness.
More concretely, the global model \(\theta\) is \(\alpha-\)proportionally fair (\(\alpha > 0\))
if the following inequality holds
\begin{equation}\label{eq:fairness:proportional:α}
  \mathcal{R}_{i}(\theta) \le \alpha
  \frac{\abs{\xi_{j}}}{\abs{\xi_{i}}} \mathcal{R}_{j}(\theta)
\end{equation}
for all clients \(i, j\) such that \(\abs{\xi_{j}} \le \abs{\xi_{i}}\).
Setting \(\alpha > 1\) (resp. \(\alpha < 1\))
favors the clients with more (resp. less) data.
While proportional fairness does account for client contributions, it is far from satisfactory.
On one hand, from a fairness perspective, clients with extremely large amounts of data
(usually controlled by large corporations) can dominate the training process.
On the other hand, a malicious client can still manipulate the training process
by inflating the number of samples they contribute.

To address these shortcomings, more sophisticated notions of fairness are needed.
For instance, \citet{xuReputationMechanismAll2020,lyuHowDemocratiseProtect2020} propose
\emph{collaborative fairness}, a measure of the correlation between contributions and rewards.
Contribution is measured by the test performance of the model uploaded by the client,
while reward is defined as the performance of the final model the client receives.
It should be noted that all clients are not assigned the same model.

\subsubsection{Core-stable fairness}\label{sub:fairness:coop:core}

\citet{raychaudhuryFairnessFederatedLearning2022} leverage the concept of the \emph{core}
from cooperative game theory~\cite{shapleyMarketGames1969}
to define a notion of fairness in \gls{abb:fl}.
The intuition behind this idea is that a model is fair if no subset of clients
can achieve a better outcome%
\footnote{%
  This is the definition of the core in cooperative game theory%
  ~\cite{shapleyMarketGames1969}.
  \citet{raychaudhuryFairnessFederatedLearning2022} use a slightly weaker definition,
  where the outcome is bounded by a factor that depends on the size of the coalition.
}
by forming a coalition among themselves.
Formally, if we denote by \(u_{i}(\theta)\) the utility of client \(i\)
under the global model \(\theta\) (defined as the complement of \(\mathcal{R}_{i}(\theta)\)),
then \(\theta\) is in the core if and only if for all \(\theta^{\prime}\in\Theta\)
and all subsets \(S\subset [N]\), the inequality
\begin{equation}\label{eq:fairness:core}
  \frac{|S|}{N}u_{i}(\theta^{\prime})\leq u_{i}(\theta)
\end{equation}
holds for at least one client \(i\in S\).

Models that are in the core have several desirable properties which make them
an attractive choice for a fairness criterion on the edge.
For instance, a model \(\theta\) in the core is \emph{proportional},
which means that the utility of all clients
is at least \(1/N\) of their best possible utility:
\[
  \forall \theta^{\prime}\in\Theta, \forall i\in[N],\,
  u_{i}(\theta)\geq \frac{1}{N}u_{i}(\theta^{\prime}).
\]
Such a model is also \emph{Pareto optimal},
meaning that no other model performs better for all clients.
Moreover, the core is \emph{scale-invariant},
which makes models in it robust to low local data quality.

Having established the qualities of the core, the question of feasibility arises.
That is, is the core non-empty? And if so,
is it possible to efficiently compute models in it?
Both questions are answered in the affirmative under mild conditions
on the continuity and convexity of the utility functions%
~\cite{raychaudhuryFairnessFederatedLearning2022}.
Furthermore, relaxations of the definition exist for cases
where the utility is non-convex (as is the case for \glspl{abb:dnn})
or when the clients are weighted differently.

\section{Privacy protection}\label{sec:privacy}

Privacy is one of the main selling points of \gls{abb:fl}.
By abolishing the need to share data, conventional wisdom says,
\gls{abb:fl} guarantees data privacy.
However, this guarantee rings hollow in the face of numerous privacy attacks%
~\cite{%
  carliniExtractingTrainingData2021,%
  pasquiniEludingSecureAggregation2022,%
  liModelArchitectureLevel2023,
  balleReconstructingTrainingData2022%
}.
While it is true that \gls{abb:fl} blocks traditional avenues of data theft
that rely on compromising a central server or intercepting data in transit,
it leaves open other attack angles that rely on what is shared:
model updates, the final model parameters, or, failing that, the model's predictions%
~\cite{shokriMembershipInferenceAttacks2017,carliniExtractingTrainingData2021}.
This state of affairs jeopardizes the privacy of individuals
whose data is used in edge \gls{abb:fl} systems,
and poses significant ethical and legal risks
to the organizations that deploy them~\cite{truongPrivacyPreservationFederated2021}.
In this section, we explore privacy threats to edge \gls{abb:fl},
proposed defenses, and their limitations.

\subsection{What models reveal about their training data}\label{sub:privacy:model-reveals}

The thesis that \gls{abb:fl} guarantees privacy is based on
the simplistic assumption that data is private unless explicitly shared,
a consequence of which is that sharing a model trained on data does not compromise its privacy.
This assumption is patently false, as the model's parameters contain ---at the very least---
information on the data distribution, otherwise the model would not be useful.
The relevant question is therefore not whether the model contains private information,
but rather
\begin{enumerate*}[label=(\alph*)]
  \item how much information it contains,\label{item:privacy:info-amount}
  \item and how much of that information can be extracted.\label{item:privacy:info-extraction}
\end{enumerate*}

Even this view, however, underestimates models' information leakage.
While classical \gls{abb:ml} wisdom, embodied by the bias-variance trade-off,
dictates that \glspl{abb:nn} must have as many parameters as necessary
to fit training data, but no more to avoid overfitting,
recent research has shown that they can be largely overparameterized,
yet have excellent generalization performance.
In fact, a neural network with zero empirical risk
(i.e., one that interpolates training data)
can have lower risk than one that slightly underfits the data%
~\cite{belkinReconcilingModernMachinelearning2019,belkinFitFearRemarkable2021}.
Due to this phenomenon, usually referred to as \emph{double descent},
large modern \glspl{abb:dnn} tend to operate in the so-called \emph{interpolation regime}.
That is to say, not only do \glspl{abb:nn} contain information about the training data,
in many cases, they contain \emph{all} the information about it.
This implies a worst-case answer to question~\ref{item:privacy:info-amount},
leaving question~\ref{item:privacy:info-extraction}
as the only hope for defending the privacy of \gls{abb:fl}.
Unfortunately, the answer to this question is also pessimistic.
As we demonstrate in \Cref{sub:privacy:attacks},
it is possible to extract a significant amount of information about the training data
from surprisingly little information about the model~\cite{%
  jereTaxonomyAttacksFederated2021,%
  rodriguez-barrosoSurveyFederatedLearning2023%
}.
Vanilla \gls{abb:fl} is in this sense no more private than centralized \gls{abb:ml}.
As such, modifications to \gls{abb:fl} are necessary to ensure privacy.

\subsection{Privacy attacks}\label{sub:privacy:attacks}

Privacy attacks against \gls{abb:fl} are numerous and diverse.
Instead of providing an exhaustive list of all known attacks,
it is more instructive to categorize them based on some relevant criteria,
and then discuss representative examples from each category, as well as potential defenses.
Many works in the literature have already provided such taxonomies%
~\cite{%
  yinComprehensiveSurveyPrivacypreserving2021,%
  rafiFairnessPrivacyPreserving2024,%
  mothukuriSurveySecurityPrivacy2021%
}.
\citet{yinComprehensiveSurveyPrivacypreserving2021} in particular,
proposed 5 classifications of privacy attacks, based on the attacker's
status (internal or external), approach (passive or active), timing (training or inference),
target (parameter updates, gradients, or the final model), and
goal (reconstruction, membership inference, etc.).
For our purposes, we will adopt the goal-based classification, as do many other works%
~\cite{%
  rafiFairnessPrivacyPreserving2024,%
  xiaSurveyFederatedLearning2021%
}.

\subsubsection{Membership inference}\label{ssub:privacy:attacks:membership}

A \gls{abb:mia} aims to determine whether a specific data point
is part of a particular client's dataset~\cite{%
  xiaSurveyFederatedLearning2021,%
  kulynychDisparateVulnerabilityMembership2022%
}.
This can be done in multiple ways,
a notable one being the \emph{binary classification-based} \gls{abb:mia},
where the attacker trains a binary classifier
to decide whether a given data point is part of the training set of a target model.
\citet{shokriMembershipInferenceAttacks2017} introduced an effective technique for doing so,
which they named \emph{shadow training}.
It works by training several \emph{shadow models} on \emph{shadow datasets},
then training an \emph{attack model} to predict whether a given data point
was part of some shadow dataset, by using information
about the corresponding shadow model~\cite{shokriMembershipInferenceAttacks2017}.

Depending on whether the attack model's input contains information
about the shadow models' final predictions or their intermediate representations,
shadow training-based \glspl{abb:mia} can be performed
both in the so-called \emph{white-box} setting, where the attacker has full knowledge
of the target model's parameters~\cite{huMembershipInferenceAttacks2022},
and in the \emph{black-box} setting,
where the attacker's knowledge is restricted to
the model's input-output behavior (through queries)%
~\cite{shokriMembershipInferenceAttacks2017}.
The latter fact is devastating to the claim of \gls{abb:fl}'s privacy,
since it's completely agnostic to the training process.
It is also worth noting that \citet{shokriMembershipInferenceAttacks2017}
demonstrated the feasibility of a black-box \gls{abb:mia} against several
\gls{abb:ml}-as-a-service platforms, including Google and Amazon,
on a hospital discharge dataset, for which membership is sensitive information.

It should be noted that not all \glspl{abb:mia} are classification-based.
An alternative family of so-called \emph{metric-based} \glspl{abb:mia} has been proposed,
which is based on the assumption that the target model performs better on data it was trained on,
so if its performance on a given data point (as measured by some metric)
is above a certain threshold, then that data point is likely to be part of the training set.
These methods are much simpler and computationally cheaper
than classification-based \glspl{abb:mia}.
The reader can refer to~\cite{huMembershipInferenceAttacks2022}
for a comprehensive survey of the existing literature on \glspl{abb:mia}.

\subsubsection{Property inference}\label{ssub:privacy:attacks:property}

The goal of a \gls{abb:pia} is to learn properties of the training data
that are not directly relevant to the model's task%
~\cite{wangPoisoningAssistedPropertyInference2023}.
For example, an attacker might seek information about the race distribution
from a model trained for gender recognition~\cite{yinComprehensiveSurveyPrivacypreserving2021}.
\Glspl{abb:pia} against \gls{abb:fl} can be performed either \emph{actively} or \emph{passively}%
~\cite{melisExploitingUnintendedFeature2019,kimExploringClusteredFederated2023}.

In the passive case, the attacker simply observes model updates and
uses them to infer the desired property.
This can be done by training a \emph{batch property classifier}
to predict whether a given gradient was produced from a batch with the property of interest.
This classifier needs to be trained on an auxiliary dataset,
distributed similarly to the victim's training data,
The classifier is then used to map intercepted gradients
to the probability of the batch having the target property,
which can be extended to the entire dataset by averaging the probabilities%
~\cite{melisExploitingUnintendedFeature2019}.
In the active case, the adversary is a participant in the \gls{abb:fl} process.
This opens the door for a simple yet more effective attack based on \emph{multitask learning}.
By simply participating in the training process using a local risk
that combines performance on the main task and the property of interest,
a discriminative representation (for the property) can be learned%
~\cite{melisExploitingUnintendedFeature2019}.

We note that both passive and active \glspl{abb:pia}
assume \emph{``honest-but-curious''} adversaries,
since neither requires the attacker to deviate from the training protocol%
~\cite{melisExploitingUnintendedFeature2019}.
It stands to reason that even more powerful attacks can be mounted by stronger adversaries.
This is demonstrated by attacks like \emph{property inference from poisoning}%
~\cite{mahloujifarPropertyInferencePoisoning2022},
and \emph{poisoning-assisted property inference}%
~\cite{wangPoisoningAssistedPropertyInference2023},
both of which are based on data poisoning (see \Cref{ssub:trust:attacks:data}).
\citet{mahloujifarPropertyInferencePoisoning2022} show that an attacker
can force a near Bayes-optimal training algorithm to leak a property,
by poisoning its local data to introduce correlation between the label and the property.
\citet{wangPoisoningAssistedPropertyInference2023} on the other hand,
present an attack that changes the labels to distort the global model's decision boundary,
thereby forcing clients to divulge additional information about the target property.
The increased power and danger of these attacks is validated both theoretically and empirically%
~\cite{wangPoisoningAssistedPropertyInference2023,mahloujifarPropertyInferencePoisoning2022}.

\subsubsection{Class representative inference}\label{ssub:privacy:attacks:representative}

A \gls{abb:cria} seeks to draw samples from the training distribution
that correspond to a specific target label,
without them necessarily being actual training data points%
~\cite{liuThreatsAttacksDefenses2022,yinComprehensiveSurveyPrivacypreserving2021}.
\citet{hitajDeepModelsGAN2017} propose one such attack based on \glspl{abb:gan},
which they prove resilient to traditional privacy-preserving mechanisms like differential privacy.
\citet{wangInferringClassRepresentatives2019} take this further,
attacking specific clients by compromising the server.
It should be noted that, as highlighted by~\cite{melisExploitingUnintendedFeature2019},
the danger of \glspl{abb:cria} is far from being obvious,
nor is the fact that they are even preventable.
On the contrary, \citet{hitajDeepModelsGAN2017} argue that they are not.

\subsubsection{Data reconstruction}\label{ssub:privacy:attacks:reconstruction}

Also known as \emph{model inversion attacks}%
\footnote{%
  The term \emph{model inversion} is sometimes used to refer to \glspl{abb:cria}%
  ~\cite{elmestariPreservingDataPrivacy2024},
  or privacy attacks in general~\cite{moQueryingLittleEnough2021}.
},
\glspl{abb:dra} aim directly at reconstructing elements (features, labels, or both)
of the training dataset.
A successful \gls{abb:dra} is the most catastrophic form of privacy leakage,
providing perfect knowledge of a subset of the training dataset,
making it equivalent to a data breach~\cite{liuThreatsAttacksDefenses2022}.
The theoretical feasibility of such attacks is a natural consequence of large \glspl{abb:nn}
operating in the interpolation regime~\cite{belkinReconcilingModernMachinelearning2019}
(making them essentially mangled copies of the training data),
combined with previous work on reconstructing databases from aggregate statistics%
~\cite{dinurRevealingInformationPreserving2003}.

One of the earliest successful attempts at devising a \gls{abb:dra}
is \gls{abb:dlg}, wherein the attacker (server or client)
uses gradients to reconstruct a specific client's local dataset.
This is done by solving an optimization problem that
to find the batch that produces the closest possible gradient to the target client's%
~\cite{zhuDeepLeakageGradients2020,zhuDeepLeakageGradients2019}.
Testing on a variety of image classification and language modeling datasets,
\citet{zhuDeepLeakageGradients2020} found that \gls{abb:dlg}
can reconstruct images to the pixel and sentences to the token.
\citet{haimReconstructingTrainingData2022} propose another optimization-based method
for reconstructing data from \(\relu-\)activated \gls{abb:mlp} binary classifiers
by relying on the \emph{``implicit bias''} of gradient-based optimization algorithms.
This was later extended by~\cite{buzagloDeconstructingDataReconstruction2023}
to work in other scenarios, including multi-class classification and other activation functions.

Both theoretical and empirical evidence suggest that \glspl{abb:dra} are a real threat.
On the theoretical side, \citet{wangReconstructingTrainingData2023}
present a tensor-decomposition-based attack
that provably reconstructs data from a single gradient.
On the empirical side, multiple mainstream models
have been shown to be vulnerable to black-box \glspl{abb:dra}.
These include generative language models like GPT-2~\cite{carliniExtractingTrainingData2021},
diffusion models like Stable-diffusion and Imagen~\cite{carliniExtractingTrainingData2023},
and text embeddings models~\cite{morrisTextEmbeddingsReveal2023}.
The latter work is particularly concerning, as it only requires API access to the target model.

\subsection{Defending against privacy attacks on the edge}\label{sub:privacy:defenses}

The number, scale, and success of the enumerated privacy attacks against \gls{abb:fl} systems,
particularly one considers the exacerbating effect of the edge environment,
calls into question the supposed privacy guarantees of edge \gls{abb:fl}.
It is therefore imperative, for ethical and legal reasons%
~\cite{truongPrivacyPreservationFederated2021},
to augment edge \gls{abb:fl} systems with privacy-preserving mechanisms,
the effectiveness of which must be provable (theoretical guarantees)
and verifiable (empirical evaluation).
In this section, we review state-of-the-art privacy defenses for \gls{abb:fl} systems.
To facilitate our review, we categorize the defenses into two broad categories:
\begin{enumerate*}[label=(\arabic*)]
  \item encryption-based defenses, and
  \item perturbation-based defenses (differential privacy).
\end{enumerate*}
By necessity, our review is not exhaustive.
We refer the reader to~\cite{
  xiaSurveyFederatedLearning2021,%
  jereTaxonomyAttacksFederated2021,%
  yinComprehensiveSurveyPrivacypreserving2021%
}
for a more comprehensive survey of privacy-preserving \gls{abb:fl},
and to~\cite{truongPrivacyPreservationFederated2021}
for a survey of privacy-preserving \gls{abb:fl} in the context of GDPR compliance.

\subsubsection{Encryption-based defenses}\label{ssub:privacy:defenses:encryption}

The most straightforward way to protect the confidentiality of data is to encrypt it.
Decades of research in cryptography have produced a variety of encryption schemes
that can provide an arbitrary level of security.
However, it is not immediately clear how encryption can be used
to protect the privacy of data in \gls{abb:fl} (or in \gls{abb:ml} in general)
~\cite{gilad-bachrachCryptoNetsApplyingNeural2016}

\paragraph{Homomorphic encryption}\label{par:privacy:defenses:encryption:homomorphic}

\Gls{abb:he} allows computations to be performed on encrypted data.
An encryption scheme is homomorphic with respect to a set of operations
if it \emph{commutes} with those operations,
meaning that applying an operation to cipher data
yields the encrypted result of applying the operation to the plaintext data%
~\cite{yiHomomorphicEncryption2014}.
This eliminates the need to decrypt the data before processing it,
and prevents the party performing the computation from learning the result,
which is desirable in cases where the server is curious.
Early works on \gls{abb:he} in the context of \gls{abb:fl}
used additively homomorphic encryption to allow the aggregation of encrypted model updates,
preventing a curious server (or an eavesdropper) from using successive gradients
for an inference attack~\cite{%
  phongPrivacyPreservingDeepLearning2018,%
  zhangPEFLPrivacyEnhancedFederated2019%
}.
Later works extended this idea using fully homomorphic encryption%
~\cite{haoEfficientPrivacyEnhancedFederated2020},
and fast homomorphic encryption schemes~\cite{zhangBatchCryptEfficientHomomorphic2020}.

\paragraph{Secure multi-party computation}\label{par:privacy:defenses:encryption:smpc}

\Gls{abb:smc} refers to a family of cryptographic protocols that allow multiple parties
to perform a computation that requires input from all of them,
without revealing any information about the input of any participant to any other
(except for what can be inferred from the output of the computation)%
~\cite{lindellSecureMultipartyComputation2021}.
In the context of \gls{abb:fl}, \gls{abb:smc} can be used to aggregate model updates.
In this direction, \citet{bonawitzPracticalSecureAggregation2017}
proposed \emph{secure aggregation},
a protocol that is provably secure against honest-but-curious and active adversaries,
at the cost of a quadratic increase in computation, communication, and storage.

\subsubsection{Differential privacy}\label{ssub:privacy:defenses:dp}

Unlike encryption-based defenses,
which aim to preserve privacy (by encrypting updates) while keeping the data intact,
perturbation-based defenses allow small, deliberate corruptions to the data
in order to provide a formal notion of privacy known as \gls{abb:dp}%
~\cite{xiaSurveyFederatedLearning2021}.
Several notions of \gls{abb:dp} exist,
\(\epsilon-\)\gls{abb:dp} and \(\epsilon,\delta-\)\gls{abb:dp} being the most common,
but they all share the same basic idea:
adding or removing a single data point from the dataset
should not affect the output distribution too much%
~\cite{dworkDifferentialPrivacySurvey2008,ouadrhiriDifferentialPrivacyDeep2022}.

The standard approach to \gls{abb:dp} in edge \gls{abb:fl} is to add noise to the model updates%
~\cite{ouadrhiriDifferentialPrivacyDeep2022}.
The distribution of the noise depends on the \emph{privacy budget} \(\epsilon,\delta\),
the \emph{sensitivity} of model updates, and the \emph{mechanism} used to add the noise
(e.g., Laplacian,
Gaussian~\cite{dworkDifferentialPrivacySurvey2008,ouadrhiriDifferentialPrivacyDeep2022},
or Skellam~\cite{agarwalSkellamMechanismDifferentially2021}).
A full survey of \gls{abb:dp} in the context of \gls{abb:fl}
is provided by~\cite{ouadrhiriDifferentialPrivacyDeep2022}.

\subsubsection{Limitations of encryption-based defenses}%
\label{ssub:privacy:defenses:encryption:limitations}

Perturbation-based privacy-preserving edge \gls{abb:fl}
suffers from an inherent trade-off between privacy and accuracy%
~\cite{kimFederatedLearningLocal2021}.
Encryption-based defenses, on the other hand, appear to have no such trade-off
because they preserve \(100\%\) of the transmitted information.
From this observation, two conclusions can be drawn.
Either
\begin{enumerate*}[label=(\arabic*)]
  \item encryption-based defenses are inherently superior to perturbation-based defenses, or
  \item encryption-based defenses are not as effective as they appear.
\end{enumerate*}

In the case of \gls{abb:he}, the latter conclusion is more likely.
As a matter of fact, the model learned by the server under \gls{abb:he}
is theoretically identical to the one learned under vanilla \gls{abb:fl}.
The only difference is that individual updates are not shared.
This implies that all the black-box attacks enumerated in \Cref{sub:privacy:attacks},
as well as the attacks on the final model parameters,
are still possible against an edge \gls{abb:fl} system secured with \gls{abb:he}.
As for \gls{abb:smc}, successful attacks against secure aggregation
have been demonstrated in the literature~\cite{pasquiniEludingSecureAggregation2022},
supporting the second conclusion in this case as well.
This pessimistic analysis is made rigorous by~\cite{el-mhamdiImpossibleSafetyLarge2023},
where the authors prove the existence of a trade-off
between model accuracy and differential privacy,
which becomes less favorable as the heterogeneity of the data increases.
As a result, the \emph{only} way to ensure \gls{abb:dp},
is to sacrifice model accuracy.

\section{Reputation, trust, and security}\label{sec:trust}

As established in \Cref{sec:data_quality}, rather than being an intrinsic property of the data,
\gls{abb:dq} is strongly dependent on extrinsic and contextual factors.
Having introduced two such factors,
namely fairness and privacy in \Cref{sec:fairness,sec:privacy},
we now turn our attention to a third one: trustworthiness.
Instead of being a property of the data,
trustworthiness is a property of the data \emph{source},
it allows for different treatments of identical data points depending on their provenance.
Despite the extrinsic nature of trustworthiness,
it is a ubiquitous dimension in the \gls{abb:dq} literature~\cite{%
  restucciaQualityInformationMobile2017,%
  hassensteinDataQualityConcepts2022,%
  wangAccuracyWhatData1996,%
  sidiDataQualitySurvey2012%
}.
It is even more salient for \gls{abb:fl} on the edge,
where the data is private, and the clients are potentially anonymous.
This section is dedicated to the study of untrusted clients in \gls{abb:fl}
(also known as \emph{Byzantine clients} for reasons that will become clear),
and the defenses that can be deployed against them.

\subsection{The Byzantine generals' problem}\label{sub:trust:byzantine}

The main reason trust is so important in \gls{abb:fl} is its distributed nature.
All distributed computing schemes suffer from a fundamental obstacle:
it is often impossible to detect a problem that exhibits different symptoms to different nodes.
This limitation, infamous in the literature as the \emph{Byzantine generals' problem},
was introduced by~\citet{lamportByzantineGeneralsProblem1982} in 1982.

The problem is usually stated abstractly using a military analogy
of a group of Byzantine generals who must coordinate their attack on a city (hence the name).
The generals must agree on whether to attack only by exchanging messengers.
Each general has an initial opinion, and the goal is to reach a consensus.
However, a certain (unknown) number of generals are traitors,
who may send false messages to prevent consensus.
Crucially, they can send different messages to different generals.
The problem is to design a protocol that allows the honest generals to agree on a plan
(no matter which one).

If at least one third of the generals are traitors,
the Byzantine generals' problem is impossible to solve~\cite{lamportByzantineGeneralsProblem1982}.
It is worth noting that this is not a technical challenge, it is an impossibility result.
\citet{lamportByzantineGeneralsProblem1982} proved that no distributed algorithm
can defeat more than \(n\) traitors in a \(3n + 1\) general network.
This is important to keep in mind when examining Byzantine defenses in \gls{abb:fl},
which can be extremely effective, but are still subject to this fundamental threshold.

\subsubsection{Impossibility results}\label{ssub:trust:defenses:impossibility}

Despite the plethora of defense strategies we have enumerated
(which represent a vanishingly small fraction of the literature),
the problem of ensuring Byzantine fault tolerance on the edge is far from solved.
\citet{el-mhamdiImpossibleSafetyLarge2023} have shown that securing \gls{abb:fl}
against poisoning attacks is impossible if the datasets of honest clients are heterogeneous.
They manage this by leveraging an equivalence result between robust mean estimation
and robust \gls{abb:fl}~\cite{el-mhamdiCollaborativeLearningJungle2021},
which they combine with a proof of impossibility for robust mean estimation to conclude.
Given existing results on the possibility of Byzantine-robust \gls{abb:fl}
with homogeneous honest clients~\cite{%
  karimireddyLearningHistoryByzantine2021,%
  el-mhamdiCollaborativeLearningJungle2021%
},
\citet{el-mhamdiImpossibleSafetyLarge2023} conclude that heterogeneity is the key obstacle.

\section{Conclusion and future research directions}\label{sec:conclusion}

In this survey, we discussed the problems data can pose for \gls{abb:ml} on the edge.
Starting with a general formulation of the learning problem in terms of \gls{abb:erm}
and \gls{abb:ferm}, we introduced a characterization of data quality in terms of
six dimensions justified by the statistical properties
of the learning problem and social/ethical considerations.
For each of these dimensions, we discussed, to the extent literature permits,
the challenges they pose, the methods to address them, and the limitations of these methods.
In doing this, we remained focused on the specific challenges
that arise in the context of \gls{abb:fl} on the edge
and the solutions that are tailored to this context.
We conclude by discussing some open problems and potential avenues for future exploration.

\subsection{Directions for further research}\label{sub:conclusion:further}

Multiple \gls{abb:dq}-related questions remain unanswered,
even more so in the context of \gls{abb:fl}.
These include:
\begin{enumerate}[label=(\arabic*)]
  \item \textbf{Other data quality considerations.}
        The list of dimensions discussed in this survey is not exhaustive.
        This is particularly true of extrinsic dimensions
        but is also true of intrinsic dimensions,
        which can be subdivided into more specific categories.
        Examples of subtleties that have not been discussed include:
        \begin{enumerate}[label=(\alph*)]
          \item Shortcuts~\cite{wuOnePixelShortcutLearning2023,geirhosShortcutLearningDeep2020},
                a special form of bias in the conditional distribution.
          \item Temporal and spatial dependencies~\cite{%
                  wangConceptDriftBasedCheckpointRestart2023,%
                  liFederatedMetalearningSpatialtemporal2022%
                }, which often lead to dependent and attribute skewed data.
          \item Missing data~\cite{guastellaEdgeBasedMissingData2021,youHandlingMissingData2020},
                which can reduce the effective sample size and introduce bias.
          \item The interplay between \gls{abb:dq} dimensions.
                The dimensions discussed in this survey are not orthogonal.
                Many of them are correlated, while others are in conflict,
                and it is not always clear which is the case.
                For example, some works suggest that differential privacy
                and social fairness are equivalent~\cite{dworkFairnessAwareness2012},
                while others find a trade-off between them~\cite{rafiFairnessPrivacyPreserving2024}.
                Other dimensions that can be related include:
                cooperative fairness, noise resilience, and Byzantine fault tolerance.
        \end{enumerate}
  \item \textbf{Understanding the effect of data quality on the learning process.}
        While \gls{abb:dq}'s influence on the learning process
        is both intuitive and empirically validated, it is not yet well understood.
        Despite attempts to rigorously study it both theoretically%
        ~\cite{izzoTheoryAlgorithmsDatacentric2023}
        and experimentally~\cite{qiImpactsDirtyData2021},
        a unified theory of data quality does not yet exist for \gls{abb:ml},
        let alone for \gls{abb:fl}.
        Characterizing the relationship between data
        and different aspects of the learning process
        such as the loss landscape, convergence and convergence rate,
        quality of the local minima, and generalization error,
        as well as the changes this relationship undergoes in the distributed setting,
        is an important, fertile, and largely underexplored area of research.
        If pursued, this line of research has the potential to facilitate
        the establishment of universally accepted definitions of data quality,
        and motivate sound practices for its assessment and improvement.
  \item \textbf{\gls{abb:dq}-robust learning algorithms.}
        Most existing methods for addressing \gls{abb:dq} issues rely on in-processing.
        This trend is likely to continue in the future,
        as learning algorithms are the most complex, sophisticated, and flexible
        components of the \gls{abb:ml} pipeline.
        While researchers exert very little control over the data collection process,
        the learning algorithm is completely under their control.
        Focusing on this component is also motivated by our observation of biological learning,
        which can extract useful models from noisy, biased, dependent, and untrusted data.
        Problems that lie in this space include \gls{abb:ood},
        where methods like \gls{abb:dro} and \gls{abb:irm} have already been proposed
        ~\cite{zhouModelAgnosticSample2022}, algorithmic robustness,
        and algorithmic stability.
\end{enumerate}

\bibliography{references}

\begin{thebibliography}{233}
\expandafter\ifx\csname natexlab\endcsname\relax\def\natexlab#1{#1}\fi
\providecommand{\url}[1]{\texttt{#1}}
\providecommand{\href}[2]{#2}
\providecommand{\path}[1]{#1}
\providecommand{\DOIprefix}{doi:}
\providecommand{\ArXivprefix}{arXiv:}
\providecommand{\URLprefix}{URL: }
\providecommand{\Pubmedprefix}{pmid:}
\providecommand{\doi}[1]{\href{http://dx.doi.org/#1}{\path{#1}}}
\providecommand{\Pubmed}[1]{\href{pmid:#1}{\path{#1}}}
\providecommand{\bibinfo}[2]{#2}
\ifx\xfnm\relax \def\xfnm[#1]{\unskip,\space#1}\fi
\bibitem[{Murshed et~al.(2022)Murshed, Murphy, Hou, Khan, Ananthanarayanan, and Hussain}]{murshedMachineLearningNetwork2022}
\bibinfo{author}{M.~G.~S. Murshed}, \bibinfo{author}{C.~Murphy}, \bibinfo{author}{D.~Hou}, \bibinfo{author}{N.~Khan}, \bibinfo{author}{G.~Ananthanarayanan}, \bibinfo{author}{F.~Hussain},
\newblock \bibinfo{title}{Machine {{Learning}} at the {{Network Edge}}: {{A Survey}}},
\newblock \bibinfo{journal}{ACM Computing Surveys} \bibinfo{volume}{54} (\bibinfo{year}{2022}) \bibinfo{pages}{1--37}. \href{http://arxiv.org/abs/1908.00080}{{\tt arXiv:1908.00080}}.
\bibitem[{Singh and Gill(2023)}]{singhEdgeAISurvey2023}
\bibinfo{author}{R.~Singh}, \bibinfo{author}{S.~S. Gill},
\newblock \bibinfo{title}{Edge {{AI}}: {{A}} survey},
\newblock \bibinfo{journal}{Internet of Things and Cyber-Physical Systems} \bibinfo{volume}{3} (\bibinfo{year}{2023}) \bibinfo{pages}{71--92}.
\bibitem[{Wang et~al.(2020)Wang, Han, Leung, Niyato, Yan, and Chen}]{wangEdgeAIConvergence2020}
\bibinfo{author}{X.~Wang}, \bibinfo{author}{Y.~Han}, \bibinfo{author}{V.~C.~M. Leung}, \bibinfo{author}{D.~Niyato}, \bibinfo{author}{X.~Yan}, \bibinfo{author}{X.~Chen}, \bibinfo{title}{Edge {{AI}}: {{Convergence}} of {{Edge Computing}} and {{Artificial Intelligence}}}, \bibinfo{publisher}{Springer}, \bibinfo{address}{Singapore}, \bibinfo{year}{2020}.
\bibitem[{Taik et~al.(2021)Taik, Moudoud, and Cherkaoui}]{taikDataQualityBasedScheduling2021}
\bibinfo{author}{A.~Taik}, \bibinfo{author}{H.~Moudoud}, \bibinfo{author}{S.~Cherkaoui},
\newblock \bibinfo{title}{Data-{{Quality Based Scheduling}} for {{Federated Edge Learning}}},
\newblock in: \bibinfo{booktitle}{2021 {{IEEE}} 46th {{Conference}} on {{Local Computer Networks}} ({{LCN}})}, \bibinfo{year}{2021}, pp. \bibinfo{pages}{17--23}. \href{http://arxiv.org/abs/2201.11247}{{\tt arXiv:2201.11247}}.
\bibitem[{Mahanti(2019)}]{mahantiDataQualityDimensions2019}
\bibinfo{author}{R.~Mahanti}, \bibinfo{title}{Data Quality: Dimensions, Measurement, Strategy, Management, and Governance}, \bibinfo{publisher}{ASQ Quality Press}, \bibinfo{address}{Milwaukee, Wisconsin}, \bibinfo{year}{2019}.
\bibitem[{Hassenstein and Vanella(2022)}]{hassensteinDataQualityConcepts2022}
\bibinfo{author}{M.~J. Hassenstein}, \bibinfo{author}{P.~Vanella},
\newblock \bibinfo{title}{Data {{Quality}}---{{Concepts}} and {{Problems}}},
\newblock \bibinfo{journal}{Encyclopedia} \bibinfo{volume}{2} (\bibinfo{year}{2022}) \bibinfo{pages}{498--510}.
\bibitem[{Camacho et~al.(2023)Camacho, Wasielewska, Espinosa, and {Fuentes-Garc{\'i}a}}]{camachoQualityQualityOut2023}
\bibinfo{author}{J.~Camacho}, \bibinfo{author}{K.~Wasielewska}, \bibinfo{author}{P.~Espinosa}, \bibinfo{author}{M.~{Fuentes-Garc{\'i}a}},
\newblock \bibinfo{title}{Quality {{In}} / {{Quality Out}}: {{Data}} quality more relevant than model choice in anomaly detection with the {{UGR}}'16},
\newblock in: \bibinfo{booktitle}{{{NOMS}} 2023-2023 {{IEEE}}/{{IFIP Network Operations}} and {{ Management Symposium}}}, \bibinfo{year}{2023}, pp. \bibinfo{pages}{1--5}.
\bibitem[{Sidi et~al.(2012)Sidi, Shariat~Panahy, Affendey, Jabar, Ibrahim, and Mustapha}]{sidiDataQualitySurvey2012}
\bibinfo{author}{F.~Sidi}, \bibinfo{author}{P.~H. Shariat~Panahy}, \bibinfo{author}{L.~S. Affendey}, \bibinfo{author}{M.~A. Jabar}, \bibinfo{author}{H.~Ibrahim}, \bibinfo{author}{A.~Mustapha},
\newblock \bibinfo{title}{Data quality: {{A}} survey of data quality dimensions},
\newblock in: \bibinfo{booktitle}{2012 {{International Conference}} on {{Information Retrieval}} \& {{Knowledge Management}}}, \bibinfo{year}{2012}, pp. \bibinfo{pages}{300--304}.
\bibitem[{Han et~al.(2021)Han, Yao, Liu, Niu, Tsang, Kwok, and Sugiyama}]{hanSurveyLabelnoiseRepresentation2021}
\bibinfo{author}{B.~Han}, \bibinfo{author}{Q.~Yao}, \bibinfo{author}{T.~Liu}, \bibinfo{author}{G.~Niu}, \bibinfo{author}{I.~W. Tsang}, \bibinfo{author}{J.~T. Kwok}, \bibinfo{author}{M.~Sugiyama}, \bibinfo{title}{A {{Survey}} of {{Label-noise Representation Learning}}: {{Past}}, {{ Present}} and {{Future}}}, \bibinfo{year}{2021}. \href{http://arxiv.org/abs/2011.04406}{{\tt arXiv:2011.04406}}.
\bibitem[{Jere et~al.(2021)Jere, Farnan, and Koushanfar}]{jereTaxonomyAttacksFederated2021}
\bibinfo{author}{M.~S. Jere}, \bibinfo{author}{T.~Farnan}, \bibinfo{author}{F.~Koushanfar},
\newblock \bibinfo{title}{A {{Taxonomy}} of {{Attacks}} on {{Federated Learning}}},
\newblock \bibinfo{journal}{IEEE Security \& Privacy} \bibinfo{volume}{19} (\bibinfo{year}{2021}) \bibinfo{pages}{20--28}.
\bibitem[{Xia et~al.(2021)Xia, Ye, Tao, Wu, and Li}]{xiaSurveyFederatedLearning2021}
\bibinfo{author}{Q.~Xia}, \bibinfo{author}{W.~Ye}, \bibinfo{author}{Z.~Tao}, \bibinfo{author}{J.~Wu}, \bibinfo{author}{Q.~Li},
\newblock \bibinfo{title}{A survey of federated learning for edge computing: {{Research}} problems and solutions},
\newblock \bibinfo{journal}{High-Confidence Computing} \bibinfo{volume}{1} (\bibinfo{year}{2021}) \bibinfo{pages}{100008}.
\bibitem[{Yin et~al.(2021)Yin, Zhu, and Hu}]{yinComprehensiveSurveyPrivacypreserving2021}
\bibinfo{author}{X.~Yin}, \bibinfo{author}{Y.~Zhu}, \bibinfo{author}{J.~Hu},
\newblock \bibinfo{title}{A {{Comprehensive Survey}} of {{Privacy-preserving Federated Learning }}: {{A Taxonomy}}, {{Review}}, and {{Future Directions}}},
\newblock \bibinfo{journal}{ACM Computing Surveys} \bibinfo{volume}{54} (\bibinfo{year}{2021}) \bibinfo{pages}{131:1--131:36}.
\bibitem[{Ferraguig et~al.(2021)Ferraguig, Djebrouni, Bouchenak, and Marangozova}]{ferraguigSurveyBiasMitigation2021}
\bibinfo{author}{L.~Ferraguig}, \bibinfo{author}{Y.~Djebrouni}, \bibinfo{author}{S.~Bouchenak}, \bibinfo{author}{V.~Marangozova},
\newblock \bibinfo{title}{Survey of {{Bias Mitigation}} in {{Federated Learning}}},
\newblock in: \bibinfo{booktitle}{Conf{\'e}rence Francophone d'informatique En {{Parall{\'e}lisme}} , {{Architecture}} et {{Syst{\`e}me}}}, \bibinfo{address}{Lyon (virtuel), France}, \bibinfo{year}{2021}.
\bibitem[{Truong et~al.(2021)Truong, Sun, Wang, Guitton, and Guo}]{truongPrivacyPreservationFederated2021}
\bibinfo{author}{N.~Truong}, \bibinfo{author}{K.~Sun}, \bibinfo{author}{S.~Wang}, \bibinfo{author}{F.~Guitton}, \bibinfo{author}{Y.~Guo},
\newblock \bibinfo{title}{Privacy preservation in federated learning: {{An}} insightful survey from the {{GDPR}} perspective},
\newblock \bibinfo{journal}{Computers \& Security} \bibinfo{volume}{110} (\bibinfo{year}{2021}) \bibinfo{pages}{102402}.
\bibitem[{Zhu et~al.(2021)Zhu, Xu, Liu, and Jin}]{zhuFederatedLearningNonIID2021}
\bibinfo{author}{H.~Zhu}, \bibinfo{author}{J.~Xu}, \bibinfo{author}{S.~Liu}, \bibinfo{author}{Y.~Jin},
\newblock \bibinfo{title}{Federated learning on non-{{IID}} data: {{A}} survey},
\newblock \bibinfo{journal}{Neurocomputing} \bibinfo{volume}{465} (\bibinfo{year}{2021}) \bibinfo{pages}{371--390}.
\bibitem[{Abreha et~al.(2022)Abreha, Hayajneh, and Serhani}]{abrehaFederatedLearningEdge2022}
\bibinfo{author}{H.~G. Abreha}, \bibinfo{author}{M.~Hayajneh}, \bibinfo{author}{M.~A. Serhani},
\newblock \bibinfo{title}{Federated {{Learning}} in {{Edge Computing}}: {{A Systematic Survey}}},
\newblock \bibinfo{journal}{Sensors} \bibinfo{volume}{22} (\bibinfo{year}{2022}) \bibinfo{pages}{450}.
\bibitem[{Liu et~al.(2022)Liu, Xu, and Wang}]{liuThreatsAttacksDefenses2022}
\bibinfo{author}{P.~Liu}, \bibinfo{author}{X.~Xu}, \bibinfo{author}{W.~Wang},
\newblock \bibinfo{title}{Threats, attacks and defenses to federated learning: Issues, taxonomy and perspectives},
\newblock \bibinfo{journal}{Cybersecurity} \bibinfo{volume}{5} (\bibinfo{year}{2022}) \bibinfo{pages}{4}.
\bibitem[{Mehrabi et~al.(2022)Mehrabi, Morstatter, Saxena, Lerman, and Galstyan}]{mehrabiSurveyBiasFairness2022}
\bibinfo{author}{N.~Mehrabi}, \bibinfo{author}{F.~Morstatter}, \bibinfo{author}{N.~Saxena}, \bibinfo{author}{K.~Lerman}, \bibinfo{author}{A.~Galstyan},
\newblock \bibinfo{title}{A {{Survey}} on {{Bias}} and {{Fairness}} in {{Machine Learning}}},
\newblock \bibinfo{journal}{ACM Computing Surveys} \bibinfo{volume}{54} (\bibinfo{year}{2022}) \bibinfo{pages}{1--35}.
\bibitem[{Xia et~al.(2023)Xia, Chen, Yu, and Ma}]{xiaPoisoningAttacksFederated2023}
\bibinfo{author}{G.~Xia}, \bibinfo{author}{J.~Chen}, \bibinfo{author}{C.~Yu}, \bibinfo{author}{J.~Ma},
\newblock \bibinfo{title}{Poisoning {{Attacks}} in {{Federated Learning}}: {{A Survey}}},
\newblock \bibinfo{journal}{IEEE Access} \bibinfo{volume}{11} (\bibinfo{year}{2023}) \bibinfo{pages}{10708--10722}.
\bibitem[{Shi et~al.(2022)Shi, Wan, Hu, Lu, and Yu~Zhang}]{shiChallengesApproachesMitigating2022}
\bibinfo{author}{J.~Shi}, \bibinfo{author}{W.~Wan}, \bibinfo{author}{S.~Hu}, \bibinfo{author}{J.~Lu}, \bibinfo{author}{L.~Yu~Zhang},
\newblock \bibinfo{title}{Challenges and {{Approaches}} for {{Mitigating Byzantine Attacks}} in {{Federated Learning}}},
\newblock in: \bibinfo{booktitle}{2022 {{IEEE International Conference}} on {{Trust}}, {{Security}} and {{Privacy}} in {{Computing}} and {{Communications}} ({{ TrustCom}})}, \bibinfo{year}{2022}, pp. \bibinfo{pages}{139--146}.
\bibitem[{{Rodr{\'i}guez-Barroso} et~al.(2023){Rodr{\'i}guez-Barroso}, {Jim{\'e}nez-L{\'o}pez}, Luz{\'o}n, Herrera, and {Mart{\'i}nez-C {\'a}mara}}]{rodriguez-barrosoSurveyFederatedLearning2023}
\bibinfo{author}{N.~{Rodr{\'i}guez-Barroso}}, \bibinfo{author}{D.~{Jim{\'e}nez-L{\'o}pez}}, \bibinfo{author}{M.~V. Luz{\'o}n}, \bibinfo{author}{F.~Herrera}, \bibinfo{author}{E.~{Mart{\'i}nez-C {\'a}mara}},
\newblock \bibinfo{title}{Survey on federated learning threats: {{Concepts}}, taxonomy on attacks and defences, experimental study and challenges},
\newblock \bibinfo{journal}{Information Fusion} \bibinfo{volume}{90} (\bibinfo{year}{2023}) \bibinfo{pages}{148--173}.
\bibitem[{Abyane et~al.(2023)Abyane, Zhu, Souza, Ma, and Hemmati}]{abyaneUnderstandingQualityChallenges2023}
\bibinfo{author}{A.~E. Abyane}, \bibinfo{author}{D.~Zhu}, \bibinfo{author}{R.~Souza}, \bibinfo{author}{L.~Ma}, \bibinfo{author}{H.~Hemmati},
\newblock \bibinfo{title}{Towards understanding quality challenges of the federated learning for neural networks: A first look from the lens of robustness},
\newblock \bibinfo{journal}{Empirical Software Engineering} \bibinfo{volume}{28} (\bibinfo{year}{2023}) \bibinfo{pages}{44}.
\bibitem[{Whang et~al.(2023)Whang, Roh, Song, and Lee}]{whangDataCollectionQuality2023}
\bibinfo{author}{S.~E. Whang}, \bibinfo{author}{Y.~Roh}, \bibinfo{author}{H.~Song}, \bibinfo{author}{J.-G. Lee},
\newblock \bibinfo{title}{Data collection and quality challenges in deep learning: A data-centric {{AI}} perspective},
\newblock \bibinfo{journal}{The VLDB Journal} \bibinfo{volume}{32} (\bibinfo{year}{2023}) \bibinfo{pages}{791--813}.
\bibitem[{Gallegos et~al.(2023)Gallegos, Rossi, Barrow, Tanjim, Kim, Dernoncourt, Yu, Zhang, and Ahmed}]{gallegosBiasFairnessLarge2023}
\bibinfo{author}{I.~O. Gallegos}, \bibinfo{author}{R.~A. Rossi}, \bibinfo{author}{J.~Barrow}, \bibinfo{author}{M.~M. Tanjim}, \bibinfo{author}{S.~Kim}, \bibinfo{author}{F.~Dernoncourt}, \bibinfo{author}{T.~Yu}, \bibinfo{author}{R.~Zhang}, \bibinfo{author}{N.~K. Ahmed}, \bibinfo{title}{Bias and {{Fairness}} in {{Large Language Models}}: {{A Survey}}}, \bibinfo{year}{2023}. \href{http://arxiv.org/abs/2309.00770}{{\tt arXiv:2309.00770}}.
\bibitem[{Lee et~al.(2023)Lee, Bang, Lovenia, Cahyawijaya, Dai, and Fung}]{leeSurveySocialBias2023}
\bibinfo{author}{N.~Lee}, \bibinfo{author}{Y.~Bang}, \bibinfo{author}{H.~Lovenia}, \bibinfo{author}{S.~Cahyawijaya}, \bibinfo{author}{W.~Dai}, \bibinfo{author}{P.~Fung}, \bibinfo{title}{Survey of {{Social Bias}} in {{Vision-Language Models}}}, \bibinfo{year}{2023}. \href{http://arxiv.org/abs/2309.14381}{{\tt arXiv:2309.14381}}.
\bibitem[{Gong et~al.(2023)Gong, Liu, Xue, Li, and Meng}]{gongSurveyDatasetQuality2023}
\bibinfo{author}{Y.~Gong}, \bibinfo{author}{G.~Liu}, \bibinfo{author}{Y.~Xue}, \bibinfo{author}{R.~Li}, \bibinfo{author}{L.~Meng},
\newblock \bibinfo{title}{A survey on dataset quality in machine learning},
\newblock \bibinfo{journal}{Information and Software Technology} \bibinfo{volume}{162} (\bibinfo{year}{2023}) \bibinfo{pages}{107268}.
\bibitem[{Wang et~al.(2023)Wang, Pal, Yang, Kant, Zhu, and Guo}]{wangCollaborativeMachineLearning2023}
\bibinfo{author}{J.~Wang}, \bibinfo{author}{A.~Pal}, \bibinfo{author}{Q.~Yang}, \bibinfo{author}{K.~Kant}, \bibinfo{author}{K.~Zhu}, \bibinfo{author}{S.~Guo},
\newblock \bibinfo{title}{Collaborative {{Machine Learning}}: {{Schemes}}, {{Robustness}}, and {{Privacy}}},
\newblock \bibinfo{journal}{IEEE Transactions on Neural Networks and Learning Systems} \bibinfo{volume}{34} (\bibinfo{year}{2023}) \bibinfo{pages}{9625--9642}.
\bibitem[{Rafi et~al.(2024)Rafi, Noor, Hussain, and Chae}]{rafiFairnessPrivacyPreserving2024}
\bibinfo{author}{T.~H. Rafi}, \bibinfo{author}{F.~A. Noor}, \bibinfo{author}{T.~Hussain}, \bibinfo{author}{D.-K. Chae},
\newblock \bibinfo{title}{Fairness and privacy preserving in federated learning: {{A}} survey},
\newblock \bibinfo{journal}{Information Fusion} \bibinfo{volume}{105} (\bibinfo{year}{2024}) \bibinfo{pages}{102198}.
\bibitem[{S{\'a}nchez~S{\'a}nchez et~al.(2024)S{\'a}nchez~S{\'a}nchez, Huertas~Celdr{\'a}n, Xie, Bovet, Mart{\'i}nez~P{\' e}rez, and Stiller}]{sanchezsanchezFederatedTrustSolutionTrustworthy2024}
\bibinfo{author}{P.~M. S{\'a}nchez~S{\'a}nchez}, \bibinfo{author}{A.~Huertas~Celdr{\'a}n}, \bibinfo{author}{N.~Xie}, \bibinfo{author}{G.~Bovet}, \bibinfo{author}{G.~Mart{\'i}nez~P{\' e}rez}, \bibinfo{author}{B.~Stiller},
\newblock \bibinfo{title}{{{FederatedTrust}}: {{A}} solution for trustworthy federated learning},
\newblock \bibinfo{journal}{Future Generation Computer Systems} \bibinfo{volume}{152} (\bibinfo{year}{2024}) \bibinfo{pages}{83--98}.
\bibitem[{Gong et~al.(2019)Gong, Zhong, and Hu}]{gongDiversityMachineLearning2019}
\bibinfo{author}{Z.~Gong}, \bibinfo{author}{P.~Zhong}, \bibinfo{author}{W.~Hu},
\newblock \bibinfo{title}{Diversity in {{Machine Learning}}},
\newblock \bibinfo{journal}{IEEE Access} \bibinfo{volume}{7} (\bibinfo{year}{2019}) \bibinfo{pages}{64323--64350}.
\bibitem[{Karkouch et~al.(2016)Karkouch, Mousannif, Al~Moatassime, and Noel}]{karkouchDataQualityInternet2016}
\bibinfo{author}{A.~Karkouch}, \bibinfo{author}{H.~Mousannif}, \bibinfo{author}{H.~Al~Moatassime}, \bibinfo{author}{T.~Noel},
\newblock \bibinfo{title}{Data quality in internet of things: {{A}} state-of-the-art survey},
\newblock \bibinfo{journal}{Journal of Network and Computer Applications} \bibinfo{volume}{73} (\bibinfo{year}{2016}) \bibinfo{pages}{57--81}.
\bibitem[{Kulkarni and Harman(2011)}]{kulkarniElementaryIntroductionStatistical2011}
\bibinfo{author}{S.~Kulkarni}, \bibinfo{author}{G.~Harman}, \bibinfo{title}{An Elementary Introduction to Statistical Learning Theory}, Wiley Series in Probability and Statistics, \bibinfo{publisher}{Wiley}, \bibinfo{address}{Hoboken, N.J}, \bibinfo{year}{2011}.
\bibitem[{Vapnik(1998)}]{vapnikStatisticalLearningTheory1998}
\bibinfo{author}{V.~N. Vapnik}, \bibinfo{title}{Statistical Learning Theory}, Adaptive and Learning Systems for Signal Processing, Communications, and Control, \bibinfo{publisher}{Wiley}, \bibinfo{address}{New York}, \bibinfo{year}{1998}.
\bibitem[{{Abu-Mostafa} et~al.(2012){Abu-Mostafa}, {Magdon-Ismail}, and Lin}]{abu-mostafaLearningDataShort2012}
\bibinfo{author}{Y.~S. {Abu-Mostafa}}, \bibinfo{author}{M.~{Magdon-Ismail}}, \bibinfo{author}{H.-T. Lin}, \bibinfo{title}{Learning from Data: A Short Course}, \bibinfo{publisher}{AMLbook.com}, \bibinfo{address}{S.l.}, \bibinfo{year}{2012}.
\bibitem[{Grohs(2023)}]{grohsMathematicalAspectsDeep2023}
\bibinfo{author}{P.~Grohs}, \bibinfo{title}{Mathematical {{Aspects}} of {{Deep Learning}}}, \bibinfo{publisher}{Cambridge University Press}, \bibinfo{address}{Cambridge}, \bibinfo{year}{2023}.
\bibitem[{Raschka et~al.(2022)Raschka, Liu, Mirjalili, and Dzhulgakov}]{raschkaMachineLearningPyTorch2022}
\bibinfo{author}{S.~Raschka}, \bibinfo{author}{Y.~Liu}, \bibinfo{author}{V.~Mirjalili}, \bibinfo{author}{D.~Dzhulgakov}, \bibinfo{title}{Machine Learning with {{PyTorch}} and {{Scikit-Learn}}: {{Develop}} Machine Learning and Deep Learning Models with {{Python}}}, Expert Insight, \bibinfo{publisher}{Packt}, \bibinfo{address}{Birmingham Mumbai}, \bibinfo{year}{2022}.
\bibitem[{Vapnik(1991)}]{vapnikPrinciplesRiskMinimization1991}
\bibinfo{author}{V.~Vapnik},
\newblock \bibinfo{title}{Principles of {{Risk Minimization}} for {{Learning Theory}}},
\newblock in: \bibinfo{booktitle}{Advances in {{Neural Information Processing Systems}}}, volume~\bibinfo{volume}{4}, \bibinfo{publisher}{Morgan-Kaufmann}, \bibinfo{year}{1991}, pp. \bibinfo{pages}{831--838}.
\bibitem[{Vapnik(2010)}]{vapnikNatureStatisticalLearning2010}
\bibinfo{author}{V.~N. Vapnik}, \bibinfo{title}{The Nature of Statistical Learning Theory}, Statistics for Engineering and Information Science, \bibinfo{edition}{second edition} ed., \bibinfo{publisher}{Springer}, \bibinfo{address}{New York, NY}, \bibinfo{year}{2010}.
\bibitem[{Zhou et~al.(2019)Zhou, Chen, Li, Zeng, Luo, and Zhang}]{zhouEdgeIntelligencePaving2019}
\bibinfo{author}{Z.~Zhou}, \bibinfo{author}{X.~Chen}, \bibinfo{author}{E.~Li}, \bibinfo{author}{L.~Zeng}, \bibinfo{author}{K.~Luo}, \bibinfo{author}{J.~Zhang},
\newblock \bibinfo{title}{Edge {{Intelligence}}: {{Paving}} the {{Last Mile}} of {{Artificial Intelligence With Edge Computing}}},
\newblock \bibinfo{journal}{Proceedings of the IEEE} \bibinfo{volume}{107} (\bibinfo{year}{2019}) \bibinfo{pages}{1738--1762}.
\bibitem[{Khouas et~al.(2024)Khouas, Bouadjenek, Hacid, and Aryal}]{khouasTrainingMachineLearning2024}
\bibinfo{author}{A.~R. Khouas}, \bibinfo{author}{M.~R. Bouadjenek}, \bibinfo{author}{H.~Hacid}, \bibinfo{author}{S.~Aryal}, \bibinfo{title}{Training {{Machine Learning}} models at the {{Edge}}: {{A Survey}}}, \bibinfo{year}{2024}. \href{http://arxiv.org/abs/2403.02619}{{\tt arXiv:2403.02619}}.
\bibitem[{Dehghani and Yazdanparast(2023)}]{dehghaniDistributedMachineDistributed2023}
\bibinfo{author}{M.~Dehghani}, \bibinfo{author}{Z.~Yazdanparast},
\newblock \bibinfo{title}{From distributed machine to distributed deep learning: A comprehensive survey},
\newblock \bibinfo{journal}{Journal of Big Data} \bibinfo{volume}{10} (\bibinfo{year}{2023}) \bibinfo{pages}{158}.
\bibitem[{Lyu et~al.(2020)Lyu, Xu, Wang, and Yu}]{lyuCollaborativeFairnessFederated2020}
\bibinfo{author}{L.~Lyu}, \bibinfo{author}{X.~Xu}, \bibinfo{author}{Q.~Wang}, \bibinfo{author}{H.~Yu},
\newblock \bibinfo{title}{Collaborative {{Fairness}} in {{Federated Learning}}},
\newblock in: \bibinfo{editor}{Q.~Yang}, \bibinfo{editor}{L.~Fan}, \bibinfo{editor}{H.~Yu} (Eds.), \bibinfo{booktitle}{Federated {{Learning}}: {{Privacy}} and {{Incentive}}}, \bibinfo{publisher}{Springer International Publishing}, \bibinfo{address}{Cham}, \bibinfo{year}{2020}, pp. \bibinfo{pages}{189--204}.
\bibitem[{McMahan et~al.(2017)McMahan, Moore, Ramage, Hampson, and y~Arcas}]{mcmahanCommunicationEfficientLearningDeep2017}
\bibinfo{author}{B.~McMahan}, \bibinfo{author}{E.~Moore}, \bibinfo{author}{D.~Ramage}, \bibinfo{author}{S.~Hampson}, \bibinfo{author}{B.~A. y~Arcas},
\newblock \bibinfo{title}{Communication-{{Efficient Learning}} of {{Deep Networks}} from {{ Decentralized Data}}},
\newblock in: \bibinfo{booktitle}{Proceedings of the 20th {{International Conference}} on {{ Artificial Intelligence}} and {{Statistics}}}, \bibinfo{publisher}{PMLR}, \bibinfo{year}{2017}, pp. \bibinfo{pages}{1273--1282}.
\bibitem[{Naik and Naik(2024)}]{naikIntroductionFederatedLearning2024}
\bibinfo{author}{D.~Naik}, \bibinfo{author}{N.~Naik},
\newblock \bibinfo{title}{An {{Introduction}} to~{{Federated Learning}}: {{Working}}, {{Types}} , {{Benefits}} and~{{Limitations}}},
\newblock in: \bibinfo{editor}{N.~Naik}, \bibinfo{editor}{P.~Jenkins}, \bibinfo{editor}{P.~Grace}, \bibinfo{editor}{L.~Yang}, \bibinfo{editor}{S.~Prajapat} (Eds.), \bibinfo{booktitle}{Advances in {{Computational Intelligence Systems}}}, \bibinfo{publisher}{Springer Nature Switzerland}, \bibinfo{address}{Cham}, \bibinfo{year}{2024}, pp. \bibinfo{pages}{3--17}.
\bibitem[{Xu et~al.(2023)Xu, Yao, Xu, Gu, Xu, and Ma}]{xuDataQualityMatters2023}
\bibinfo{author}{S.~Xu}, \bibinfo{author}{Y.~Yao}, \bibinfo{author}{F.~Xu}, \bibinfo{author}{T.~Gu}, \bibinfo{author}{J.~Xu}, \bibinfo{author}{X.~Ma},
\newblock \bibinfo{title}{Data {{Quality Matters}}: {{A Case Study}} of {{Obsolete Comment Detection}}},
\newblock in: \bibinfo{booktitle}{2023 {{IEEE}}/{{ACM}} 45th {{International Conference}} on {{ Software Engineering}} ({{ICSE}})}, \bibinfo{year}{2023}, pp. \bibinfo{pages}{781--793}.
\bibitem[{Gupta et~al.(2021)Gupta, Mujumdar, Patel, Masuda, Panwar, Bandyopadhyay, Mehta, Guttula, Afzal, Sharma~Mittal, and Munigala}]{guptaDataQualityMachine2021}
\bibinfo{author}{N.~Gupta}, \bibinfo{author}{S.~Mujumdar}, \bibinfo{author}{H.~Patel}, \bibinfo{author}{S.~Masuda}, \bibinfo{author}{N.~Panwar}, \bibinfo{author}{S.~Bandyopadhyay}, \bibinfo{author}{S.~Mehta}, \bibinfo{author}{S.~Guttula}, \bibinfo{author}{S.~Afzal}, \bibinfo{author}{R.~Sharma~Mittal}, \bibinfo{author}{V.~Munigala},
\newblock \bibinfo{title}{Data {{Quality}} for {{Machine Learning Tasks}}},
\newblock in: \bibinfo{booktitle}{Proceedings of the 27th {{ACM SIGKDD Conference}} on {{Knowledge Discovery}} \& {{Data Mining}}}, {{KDD}} '21, \bibinfo{publisher}{Association for Computing Machinery}, \bibinfo{address}{New York, NY, USA}, \bibinfo{year}{2021}, pp. \bibinfo{pages}{4040--4041}.
\bibitem[{Olson(2003)}]{olsonDataQualityAccuracy2003}
\bibinfo{author}{J.~E. Olson}, \bibinfo{title}{Data Quality: The Accuracy Dimension}, \bibinfo{publisher}{Morgan Kaufmann}, \bibinfo{address}{San Francisco}, \bibinfo{year}{2003}.
\bibitem[{Renggli et~al.(2021)Renggli, Rimanic, G{\"u}rel, Karla{\v s}, Wu, and Zhang}]{renggliDataQualityDrivenView2021}
\bibinfo{author}{C.~Renggli}, \bibinfo{author}{L.~Rimanic}, \bibinfo{author}{N.~M. G{\"u}rel}, \bibinfo{author}{B.~Karla{\v s}}, \bibinfo{author}{W.~Wu}, \bibinfo{author}{C.~Zhang}, \bibinfo{title}{A {{Data Quality-Driven View}} of {{MLOps}}}, \bibinfo{year}{2021}. \href{http://arxiv.org/abs/2102.07750}{{\tt arXiv:2102.07750}}.
\bibitem[{Gudivada et~al.(2017)Gudivada, Apon, and Ding}]{gudivadaDataQualityConsiderations2017}
\bibinfo{author}{V.~N. Gudivada}, \bibinfo{author}{A.~Apon}, \bibinfo{author}{J.~Ding},
\newblock \bibinfo{title}{Data {{Quality Considerations}} for {{Big Data}} and {{Machine Learning}}: {{Going Beyond Data Cleaning}} and {{Transformations}}}  (\bibinfo{year}{2017}).
\bibitem[{Hagendorff(2021)}]{hagendorffLinkingHumanMachine2021}
\bibinfo{author}{T.~Hagendorff},
\newblock \bibinfo{title}{Linking {{Human And Machine Behavior}}: {{A New Approach}} to {{ Evaluate Training Data Quality}} for {{Beneficial Machine Learning}}},
\newblock \bibinfo{journal}{Minds and Machines} \bibinfo{volume}{31} (\bibinfo{year}{2021}) \bibinfo{pages}{563--593}.
\bibitem[{Budach et~al.(2022)Budach, Feuerpfeil, Ihde, Nathansen, Noack, Patzlaff, Naumann, and Harmouch}]{budachEffectsDataQuality2022}
\bibinfo{author}{L.~Budach}, \bibinfo{author}{M.~Feuerpfeil}, \bibinfo{author}{N.~Ihde}, \bibinfo{author}{A.~Nathansen}, \bibinfo{author}{N.~Noack}, \bibinfo{author}{H.~Patzlaff}, \bibinfo{author}{F.~Naumann}, \bibinfo{author}{H.~Harmouch}, \bibinfo{title}{The {{Effects}} of {{Data Quality}} on {{Machine Learning Performance }}}, \bibinfo{year}{2022}. \href{http://arxiv.org/abs/2207.14529}{{\tt arXiv:2207.14529}}.
\bibitem[{Danilov et~al.(2023)Danilov, Kotik, Shifrin, Strunina, Pronkina, Tsukanova, Nepomnyashiy, Konovalov, Danilov, and Potapov}]{danilovDataQualityEstimation2023}
\bibinfo{author}{G.~Danilov}, \bibinfo{author}{K.~Kotik}, \bibinfo{author}{M.~Shifrin}, \bibinfo{author}{Y.~Strunina}, \bibinfo{author}{T.~Pronkina}, \bibinfo{author}{T.~Tsukanova}, \bibinfo{author}{V.~Nepomnyashiy}, \bibinfo{author}{N.~Konovalov}, \bibinfo{author}{V.~Danilov}, \bibinfo{author}{A.~Potapov},
\newblock \bibinfo{title}{Data {{Quality Estimation Via Model Performance}}: {{Machine Learning }} as a {{Validation Tool}}},
\newblock in: \bibinfo{booktitle}{Healthcare {{Transformation}} with {{Informatics}} and {{ Artificial Intelligence}}}, \bibinfo{publisher}{IOS Press}, \bibinfo{year}{2023}, pp. \bibinfo{pages}{369--372}.
\bibitem[{Newey and McFadden(1994)}]{neweyChapter36Large1994}
\bibinfo{author}{W.~K. Newey}, \bibinfo{author}{D.~McFadden},
\newblock \bibinfo{title}{Chapter 36 {{Large}} sample estimation and hypothesis testing},
\newblock in: \bibinfo{booktitle}{Handbook of {{Econometrics}}}, volume~\bibinfo{volume}{4}, \bibinfo{publisher}{Elsevier}, \bibinfo{year}{1994}, pp. \bibinfo{pages}{2111--2245}.
\bibitem[{{Ram{\'i}rez-Gallego} et~al.(2017){Ram{\'i}rez-Gallego}, Krawczyk, Garc{\'i}a, Wo{\'z}niak, and Herrera}]{ramirez-gallegoSurveyDataPreprocessing2017}
\bibinfo{author}{S.~{Ram{\'i}rez-Gallego}}, \bibinfo{author}{B.~Krawczyk}, \bibinfo{author}{S.~Garc{\'i}a}, \bibinfo{author}{M.~Wo{\'z}niak}, \bibinfo{author}{F.~Herrera},
\newblock \bibinfo{title}{A survey on data preprocessing for data stream mining: {{Current}} status and future directions},
\newblock \bibinfo{journal}{Neurocomputing} \bibinfo{volume}{239} (\bibinfo{year}{2017}) \bibinfo{pages}{39--57}.
\bibitem[{{Werner~de~Vargas} et~al.(2023){Werner~de~Vargas}, Schneider~Aranda, { dos Santos~Costa}, {da Silva~Pereira}, and Vict{\'o}ria~Barbosa}]{wernerdevargasImbalancedDataPreprocessing2023}
\bibinfo{author}{V.~{Werner~de~Vargas}}, \bibinfo{author}{J.~A. Schneider~Aranda}, \bibinfo{author}{R.~{ dos Santos~Costa}}, \bibinfo{author}{P.~R. {da Silva~Pereira}}, \bibinfo{author}{J.~L. Vict{\'o}ria~Barbosa},
\newblock \bibinfo{title}{Imbalanced data preprocessing techniques for machine learning: A systematic mapping study},
\newblock \bibinfo{journal}{Knowledge and Information Systems} \bibinfo{volume}{65} (\bibinfo{year}{2023}) \bibinfo{pages}{31--57}.
\bibitem[{You et~al.(2020)You, Ma, Ding, Kochenderfer, and Leskovec}]{youHandlingMissingData2020}
\bibinfo{author}{J.~You}, \bibinfo{author}{X.~Ma}, \bibinfo{author}{Y.~Ding}, \bibinfo{author}{M.~J. Kochenderfer}, \bibinfo{author}{J.~Leskovec},
\newblock \bibinfo{title}{Handling {{Missing Data}} with {{Graph Representation Learning}}},
\newblock in: \bibinfo{booktitle}{Advances in {{Neural Information Processing Systems}}}, volume~\bibinfo{volume}{33}, \bibinfo{publisher}{Curran Associates, Inc.}, \bibinfo{year}{2020}, pp. \bibinfo{pages}{19075--19087}.
\bibitem[{Zhang and Sabuncu(2018)}]{zhangGeneralizedCrossEntropy2018}
\bibinfo{author}{Z.~Zhang}, \bibinfo{author}{M.~Sabuncu},
\newblock \bibinfo{title}{Generalized {{Cross Entropy Loss}} for {{Training Deep Neural Networks}} with {{Noisy Labels}}},
\newblock in: \bibinfo{booktitle}{Advances in {{Neural Information Processing Systems}}}, volume~\bibinfo{volume}{31}, \bibinfo{publisher}{Curran Associates, Inc.}, \bibinfo{year}{2018}.
\bibitem[{Lin et~al.(2022)Lin, Fang, and Gao}]{linDistributionallyRobustOptimization2022}
\bibinfo{author}{F.~Lin}, \bibinfo{author}{X.~Fang}, \bibinfo{author}{Z.~Gao},
\newblock \bibinfo{title}{Distributionally {{Robust Optimization}}: {{A}} review on theory and applications},
\newblock \bibinfo{journal}{Numerical Algebra, Control and Optimization} \bibinfo{volume}{12} (\bibinfo{year}{Mon Feb 28 19:00:00 EST 2022}) \bibinfo{pages}{159--212}.
\bibitem[{Zhou and Wang(2024)}]{zhouFederatedLabelNoiseLearning2024}
\bibinfo{author}{X.~Zhou}, \bibinfo{author}{X.~Wang},
\newblock \bibinfo{title}{Federated {{Label-Noise Learning}} with {{Local Diversity Product Regularization}}},
\newblock \bibinfo{journal}{Proceedings of the AAAI Conference on Artificial Intelligence} \bibinfo{volume}{38} (\bibinfo{year}{2024}) \bibinfo{pages}{17141--17149}.
\bibitem[{Goldberger and {Ben-Reuven}(2022)}]{goldbergerTrainingDeepNeuralnetworks2022}
\bibinfo{author}{J.~Goldberger}, \bibinfo{author}{E.~{Ben-Reuven}},
\newblock \bibinfo{title}{Training deep neural-networks using a noise adaptation layer},
\newblock in: \bibinfo{booktitle}{International {{Conference}} on {{Learning Representations}}}, \bibinfo{year}{2022}.
\bibitem[{Zafar et~al.(2017)Zafar, Valera, Rogriguez, and Gummadi}]{zafarFairnessConstraintsMechanisms2017}
\bibinfo{author}{M.~B. Zafar}, \bibinfo{author}{I.~Valera}, \bibinfo{author}{M.~G. Rogriguez}, \bibinfo{author}{K.~P. Gummadi},
\newblock \bibinfo{title}{Fairness {{Constraints}}: {{Mechanisms}} for {{Fair Classification}}},
\newblock in: \bibinfo{booktitle}{Proceedings of the 20th {{International Conference}} on {{ Artificial Intelligence}} and {{Statistics}}}, \bibinfo{publisher}{PMLR}, \bibinfo{year}{2017}, pp. \bibinfo{pages}{962--970}.
\bibitem[{Li et~al.(2021)Li, Zhang, Tan, Qin, Wang, and Li}]{liSamplelevelDataSelection2021}
\bibinfo{author}{A.~Li}, \bibinfo{author}{L.~Zhang}, \bibinfo{author}{J.~Tan}, \bibinfo{author}{Y.~Qin}, \bibinfo{author}{J.~Wang}, \bibinfo{author}{X.-Y. Li},
\newblock \bibinfo{title}{Sample-level {{Data Selection}} for {{Federated Learning}}},
\newblock in: \bibinfo{booktitle}{{{IEEE INFOCOM}} 2021 - {{IEEE Conference}} on {{Computer Communications}}}, \bibinfo{publisher}{IEEE}, \bibinfo{address}{Vancouver, BC, Canada}, \bibinfo{year}{2021}, pp. \bibinfo{pages}{1--10}.
\bibitem[{{El-Mhamdi} et~al.(2023){El-Mhamdi}, Farhadkhani, Guerraoui, and Hoang}]{el-mhamdiStrategyproofnessGeometricMedian2023}
\bibinfo{author}{E.-M. {El-Mhamdi}}, \bibinfo{author}{S.~Farhadkhani}, \bibinfo{author}{R.~Guerraoui}, \bibinfo{author}{L.-N. Hoang},
\newblock \bibinfo{title}{On the {{Strategyproofness}} of the {{Geometric Median}}},
\newblock in: \bibinfo{booktitle}{Proceedings of {{The}} 26th {{International Conference}} on {{ Artificial Intelligence}} and {{Statistics}}}, \bibinfo{publisher}{PMLR}, \bibinfo{year}{2023}, pp. \bibinfo{pages}{2603--2640}.
\bibitem[{Bolukbasi et~al.(2016)Bolukbasi, Chang, Zou, Saligrama, and Kalai}]{bolukbasiManComputerProgrammer2016}
\bibinfo{author}{T.~Bolukbasi}, \bibinfo{author}{K.-W. Chang}, \bibinfo{author}{J.~Y. Zou}, \bibinfo{author}{V.~Saligrama}, \bibinfo{author}{A.~T. Kalai},
\newblock \bibinfo{title}{Man is to {{Computer Programmer}} as {{Woman}} is to {{Homemaker}}? { {Debiasing Word Embeddings}}},
\newblock in: \bibinfo{booktitle}{Advances in {{Neural Information Processing Systems}}}, volume~\bibinfo{volume}{29}, \bibinfo{publisher}{Curran Associates, Inc.}, \bibinfo{year}{2016}.
\bibitem[{Hardt et~al.(2016)Hardt, Price, Price, and Srebro}]{hardtEqualityOpportunitySupervised2016}
\bibinfo{author}{M.~Hardt}, \bibinfo{author}{E.~Price}, \bibinfo{author}{E.~Price}, \bibinfo{author}{N.~Srebro},
\newblock \bibinfo{title}{Equality of {{Opportunity}} in {{Supervised Learning}}},
\newblock in: \bibinfo{booktitle}{Advances in {{Neural Information Processing Systems}}}, volume~\bibinfo{volume}{29}, \bibinfo{publisher}{Curran Associates, Inc.}, \bibinfo{year}{2016}.
\bibitem[{Petersen et~al.(2021)Petersen, Mukherjee, Sun, and Yurochkin}]{petersenPostprocessingIndividualFairness2021}
\bibinfo{author}{F.~Petersen}, \bibinfo{author}{D.~Mukherjee}, \bibinfo{author}{Y.~Sun}, \bibinfo{author}{M.~Yurochkin},
\newblock \bibinfo{title}{Post-processing for {{Individual Fairness}}},
\newblock in: \bibinfo{booktitle}{Advances in {{Neural Information Processing Systems}}}, volume~\bibinfo{volume}{34}, \bibinfo{publisher}{Curran Associates, Inc.}, \bibinfo{year}{2021}, pp. \bibinfo{pages}{25944--25955}.
\bibitem[{Liu et~al.(2023)Liu, Shen, He, Zhang, Xu, Yu, and Cui}]{liuOutOfDistributionGeneralizationSurvey2023}
\bibinfo{author}{J.~Liu}, \bibinfo{author}{Z.~Shen}, \bibinfo{author}{Y.~He}, \bibinfo{author}{X.~Zhang}, \bibinfo{author}{R.~Xu}, \bibinfo{author}{H.~Yu}, \bibinfo{author}{P.~Cui}, \bibinfo{title}{Towards {{Out-Of-Distribution Generalization}}: {{A Survey}}}, \bibinfo{year}{2023}. \href{http://arxiv.org/abs/2108.13624}{{\tt arXiv:2108.13624}}.
\bibitem[{Wang et~al.(2021)Wang, {Mu{\~n}oz-Gonz{\'a}lez}, Eklund, and Raza}]{wangNonIIDDataRebalancing2021}
\bibinfo{author}{H.~Wang}, \bibinfo{author}{L.~{Mu{\~n}oz-Gonz{\'a}lez}}, \bibinfo{author}{D.~Eklund}, \bibinfo{author}{S.~Raza},
\newblock \bibinfo{title}{Non-{{IID}} data re-balancing at {{IoT}} edge with peer-to-peer federated learning for anomaly detection},
\newblock in: \bibinfo{booktitle}{Proceedings of the 14th {{ACM Conference}} on {{Security}} and {{ Privacy}} in {{Wireless}} and {{Mobile Networks}}}, {{WiSec}} '21, \bibinfo{publisher}{Association for Computing Machinery}, \bibinfo{address}{New York, NY, USA}, \bibinfo{year}{2021}, pp. \bibinfo{pages}{153--163}.
\bibitem[{Li et~al.(2020)Li, Huang, Yang, Wang, and Zhang}]{liConvergenceFedAvgNonIID2020}
\bibinfo{author}{X.~Li}, \bibinfo{author}{K.~Huang}, \bibinfo{author}{W.~Yang}, \bibinfo{author}{S.~Wang}, \bibinfo{author}{Z.~Zhang}, \bibinfo{title}{On the {{Convergence}} of {{FedAvg}} on {{Non-IID Data}}}, \bibinfo{year}{2020}. \href{http://arxiv.org/abs/1907.02189}{{\tt arXiv:1907.02189}}.
\bibitem[{Zou et~al.(2009)Zou, Li, and Xu}]{zouGeneralizationPerformanceERM2009}
\bibinfo{author}{B.~Zou}, \bibinfo{author}{L.~Li}, \bibinfo{author}{Z.~Xu},
\newblock \bibinfo{title}{The generalization performance of {{ERM}} algorithm with~strongly mixing observations},
\newblock \bibinfo{journal}{Machine Learning} \bibinfo{volume}{75} (\bibinfo{year}{2009}) \bibinfo{pages}{275--295}.
\bibitem[{Guo and Shi(2011)}]{guoClassificationNoniSampling2011}
\bibinfo{author}{Z.-C. Guo}, \bibinfo{author}{L.~Shi},
\newblock \bibinfo{title}{Classification with non-i.i.d. sampling},
\newblock \bibinfo{journal}{Mathematical and Computer Modelling} \bibinfo{volume}{54} (\bibinfo{year}{2011}) \bibinfo{pages}{1347--1364}.
\bibitem[{Zimin(2018)}]{ziminLearningDependentData2018}
\bibinfo{author}{A.~Zimin}, \bibinfo{title}{Learning from Dependent Data}, \bibinfo{type}{Thesis}, Institute of Science and Technology Austria, \bibinfo{year}{2018}.
\bibitem[{Roy et~al.(2021)Roy, Balasubramanian, and Erdogdu}]{royEmpiricalRiskMinimization2021}
\bibinfo{author}{A.~Roy}, \bibinfo{author}{K.~Balasubramanian}, \bibinfo{author}{M.~A. Erdogdu},
\newblock \bibinfo{title}{On {{Empirical Risk Minimization}} with {{Dependent}} and {{ Heavy-Tailed Data}}},
\newblock in: \bibinfo{booktitle}{Advances in {{Neural Information Processing Systems}}}, volume~\bibinfo{volume}{34}, \bibinfo{publisher}{Curran Associates, Inc.}, \bibinfo{year}{2021}, pp. \bibinfo{pages}{8913--8926}.
\bibitem[{Dundar et~al.(????)Dundar, Krishnapuram, Bi, and Rao}]{dundarLearningClassifiersWhen}
\bibinfo{author}{M.~Dundar}, \bibinfo{author}{B.~Krishnapuram}, \bibinfo{author}{J.~Bi}, \bibinfo{author}{R.~B. Rao},
\newblock \bibinfo{title}{Learning {{Classifiers When The Training Data Is Not IID}}},
\newblock in: \bibinfo{booktitle}{International {{Joint Conference}} on {{Artificial Intelligence}}}, ????
\bibitem[{Lauer(2023)}]{lauerUniformRiskBounds2023}
\bibinfo{author}{F.~Lauer}, \bibinfo{title}{Uniform {{Risk Bounds}} for {{Learning}} with {{Dependent Data Sequences}}}, \bibinfo{year}{2023}. \href{http://arxiv.org/abs/2303.11650}{{\tt arXiv:2303.11650}}.
\bibitem[{Lee(2023)}]{leeConvergenceFederatedLearning2023}
\bibinfo{author}{H.~Lee},
\newblock \bibinfo{title}{Towards {{Convergence}} in {{Federated Learning}} via {{Non-IID Analysis}} in a {{Distributed Solar Energy Grid}}},
\newblock \bibinfo{journal}{Electronics} \bibinfo{volume}{12} (\bibinfo{year}{2023}) \bibinfo{pages}{1580}.
\bibitem[{Zhao et~al.(2018)Zhao, Li, Lai, Suda, Civin, and Chandra}]{zhaoFederatedLearningNonIID2018}
\bibinfo{author}{Y.~Zhao}, \bibinfo{author}{M.~Li}, \bibinfo{author}{L.~Lai}, \bibinfo{author}{N.~Suda}, \bibinfo{author}{D.~Civin}, \bibinfo{author}{V.~Chandra},
\newblock \bibinfo{title}{Federated {{Learning}} with {{Non-IID Data}}}  (\bibinfo{year}{2018}). \href{http://arxiv.org/abs/1806.00582}{{\tt arXiv:1806.00582}}.
\bibitem[{Wang et~al.(2023{\natexlab{a}})Wang, Shi, and Chang}]{wangWhyBatchNormalization2023}
\bibinfo{author}{Y.~Wang}, \bibinfo{author}{Q.~Shi}, \bibinfo{author}{T.-H. Chang},
\newblock \bibinfo{title}{Why {{Batch Normalization Damage Federated Learning}} on {{Non-IID Data}}?},
\newblock \bibinfo{journal}{IEEE transactions on neural networks and learning systems} \bibinfo{volume}{PP} (\bibinfo{year}{2023}{\natexlab{a}}).
\bibitem[{Wang et~al.(2023{\natexlab{b}})Wang, Shi, and Chang}]{wangBatchNormalizationDamages2023}
\bibinfo{author}{Y.~Wang}, \bibinfo{author}{Q.~Shi}, \bibinfo{author}{T.-H. Chang},
\newblock \bibinfo{title}{Batch {{Normalization Damages Federated Learning}} on {{NON-IID Data} }: {{Analysis}} and {{Remedy}}},
\newblock in: \bibinfo{booktitle}{{{ICASSP}} 2023 - 2023 {{IEEE International Conference}} on {{ Acoustics}}, {{Speech}} and {{Signal Processing}} ({{ICASSP}})}, \bibinfo{year}{2023}{\natexlab{b}}, pp. \bibinfo{pages}{1--5}.
\bibitem[{Ma et~al.(2022)Ma, Zhu, Lin, Chen, and Qin}]{maStateoftheartSurveySolving2022}
\bibinfo{author}{X.~Ma}, \bibinfo{author}{J.~Zhu}, \bibinfo{author}{Z.~Lin}, \bibinfo{author}{S.~Chen}, \bibinfo{author}{Y.~Qin},
\newblock \bibinfo{title}{A state-of-the-art survey on solving non-{{IID}} data in {{Federated Learning}}},
\newblock \bibinfo{journal}{Future Generation Computer Systems} \bibinfo{volume}{135} (\bibinfo{year}{2022}) \bibinfo{pages}{244--258}.
\bibitem[{Lu et~al.(2024)Lu, Pan, Dai, Si, and Zhang}]{luFederatedLearningNonIID2024}
\bibinfo{author}{Z.~Lu}, \bibinfo{author}{H.~Pan}, \bibinfo{author}{Y.~Dai}, \bibinfo{author}{X.~Si}, \bibinfo{author}{Y.~Zhang},
\newblock \bibinfo{title}{Federated {{Learning With Non-IID Data}}: {{A Survey}}},
\newblock \bibinfo{journal}{IEEE Internet of Things Journal}  (\bibinfo{year}{2024}) \bibinfo{pages}{1--1}.
\bibitem[{Yoshida et~al.(2020)Yoshida, Nishio, Morikura, Yamamoto, and Yonetani}]{yoshidaHybridFLWirelessNetworks2020}
\bibinfo{author}{N.~Yoshida}, \bibinfo{author}{T.~Nishio}, \bibinfo{author}{M.~Morikura}, \bibinfo{author}{K.~Yamamoto}, \bibinfo{author}{R.~Yonetani},
\newblock \bibinfo{title}{Hybrid-{{FL}} for {{Wireless Networks}}: {{Cooperative Learning Mechanism Using Non-IID Data}}},
\newblock in: \bibinfo{booktitle}{{{ICC}} 2020 - 2020 {{IEEE International Conference}} on {{ Communications}} ({{ICC}})}, \bibinfo{year}{2020}, pp. \bibinfo{pages}{1--7}.
\bibitem[{Tian et~al.(2021)Tian, Chen, Yu, and Liao}]{tianAsynchronousFederatedLearning2021}
\bibinfo{author}{P.~Tian}, \bibinfo{author}{Z.~Chen}, \bibinfo{author}{W.~Yu}, \bibinfo{author}{W.~Liao},
\newblock \bibinfo{title}{Towards asynchronous federated learning based threat detection: {{A DC-Adam}} approach},
\newblock \bibinfo{journal}{Computers \& Security} \bibinfo{volume}{108} (\bibinfo{year}{2021}) \bibinfo{pages}{102344}.
\bibitem[{Abay et~al.(2020)Abay, Zhou, Baracaldo, Rajamoni, Chuba, and Ludwig}]{abayMitigatingBiasFederated2020}
\bibinfo{author}{A.~Abay}, \bibinfo{author}{Y.~Zhou}, \bibinfo{author}{N.~Baracaldo}, \bibinfo{author}{S.~Rajamoni}, \bibinfo{author}{E.~Chuba}, \bibinfo{author}{H.~Ludwig}, \bibinfo{title}{Mitigating {{Bias}} in {{Federated Learning}}}, \bibinfo{year}{2020}.
\bibitem[{Jeong et~al.(2023)Jeong, Oh, Kim, Park, Bennis, and Kim}]{jeongCommunicationEfficientOnDeviceMachine2023}
\bibinfo{author}{E.~Jeong}, \bibinfo{author}{S.~Oh}, \bibinfo{author}{H.~Kim}, \bibinfo{author}{J.~Park}, \bibinfo{author}{M.~Bennis}, \bibinfo{author}{S.-L. Kim}, \bibinfo{title}{Communication-{{Efficient On-Device Machine Learning}}: {{Federated Distillation}} and {{Augmentation}} under {{Non-IID Private Data}}}, \bibinfo{year}{2023}. \href{http://arxiv.org/abs/1811.11479}{{\tt arXiv:1811.11479}}.
\bibitem[{Rai et~al.(2022)Rai, Kumari, and Prasad}]{raiClientSelectionFederated2022}
\bibinfo{author}{S.~Rai}, \bibinfo{author}{A.~Kumari}, \bibinfo{author}{D.~K. Prasad},
\newblock \bibinfo{title}{Client {{Selection}} in {{Federated Learning}} under {{Imperfections} } in {{Environment}}},
\newblock \bibinfo{journal}{AI} \bibinfo{volume}{3} (\bibinfo{year}{2022}) \bibinfo{pages}{124--145}.
\bibitem[{Wang et~al.(2020)Wang, Kaplan, Niu, and Li}]{wangOptimizingFederatedLearning2020}
\bibinfo{author}{H.~Wang}, \bibinfo{author}{Z.~Kaplan}, \bibinfo{author}{D.~Niu}, \bibinfo{author}{B.~Li},
\newblock \bibinfo{title}{Optimizing {{Federated Learning}} on {{Non-IID Data}} with {{ Reinforcement Learning}}},
\newblock in: \bibinfo{booktitle}{{{IEEE INFOCOM}} 2020 - {{IEEE Conference}} on {{Computer Communications}}}, \bibinfo{year}{2020}, pp. \bibinfo{pages}{1698--1707}.
\bibitem[{Jiang et~al.(2023)Jiang, Kone{\v c}n{\'y}, Rush, and Kannan}]{jiangImprovingFederatedLearning2023}
\bibinfo{author}{Y.~Jiang}, \bibinfo{author}{J.~Kone{\v c}n{\'y}}, \bibinfo{author}{K.~Rush}, \bibinfo{author}{S.~Kannan}, \bibinfo{title}{Improving {{Federated Learning Personalization}} via {{Model Agnostic Meta Learning}}}, \bibinfo{year}{2023}. \href{http://arxiv.org/abs/1909.12488}{{\tt arXiv:1909.12488}}.
\bibitem[{Finn et~al.(2017)Finn, Abbeel, and Levine}]{finnModelAgnosticMetaLearningFast2017}
\bibinfo{author}{C.~Finn}, \bibinfo{author}{P.~Abbeel}, \bibinfo{author}{S.~Levine},
\newblock \bibinfo{title}{Model-{{Agnostic Meta-Learning}} for {{Fast Adaptation}} of {{Deep Networks}}},
\newblock in: \bibinfo{booktitle}{Proceedings of the 34th {{International Conference}} on {{Machine Learning}}}, \bibinfo{publisher}{PMLR}, \bibinfo{year}{2017}, pp. \bibinfo{pages}{1126--1135}.
\bibitem[{Li and Wang(2022)}]{liFederatedMetalearningSpatialtemporal2022}
\bibinfo{author}{W.~Li}, \bibinfo{author}{S.~Wang},
\newblock \bibinfo{title}{Federated meta-learning for spatial-temporal prediction},
\newblock \bibinfo{journal}{Neural Computing and Applications} \bibinfo{volume}{34} (\bibinfo{year}{2022}) \bibinfo{pages}{10355--10374}.
\bibitem[{Zhang et~al.(2021)Zhang, Hong, Dhople, Yin, and Liu}]{zhangFedPDFederatedLearning2021}
\bibinfo{author}{X.~Zhang}, \bibinfo{author}{M.~Hong}, \bibinfo{author}{S.~Dhople}, \bibinfo{author}{W.~Yin}, \bibinfo{author}{Y.~Liu},
\newblock \bibinfo{title}{{{FedPD}}: {{A Federated Learning Framework With Adaptivity}} to {{ Non-IID Data}}},
\newblock \bibinfo{journal}{IEEE Transactions on Signal Processing} \bibinfo{volume}{69} (\bibinfo{year}{2021}) \bibinfo{pages}{6055--6070}.
\bibitem[{Smith et~al.(2017)Smith, Chiang, Sanjabi, and Talwalkar}]{smithFederatedMultiTaskLearning2017}
\bibinfo{author}{V.~Smith}, \bibinfo{author}{C.-K. Chiang}, \bibinfo{author}{M.~Sanjabi}, \bibinfo{author}{A.~S. Talwalkar},
\newblock \bibinfo{title}{Federated {{Multi-Task Learning}}},
\newblock in: \bibinfo{booktitle}{Advances in {{Neural Information Processing Systems}}}, volume~\bibinfo{volume}{30}, \bibinfo{publisher}{Curran Associates, Inc.}, \bibinfo{year}{2017}.
\bibitem[{Corinzia et~al.(2021)Corinzia, Beuret, and Buhmann}]{corinziaVariationalFederatedMultiTask2021}
\bibinfo{author}{L.~Corinzia}, \bibinfo{author}{A.~Beuret}, \bibinfo{author}{J.~M. Buhmann}, \bibinfo{title}{Variational {{Federated Multi-Task Learning}}}, \bibinfo{year}{2021}. \href{http://arxiv.org/abs/1906.06268}{{\tt arXiv:1906.06268}}.
\bibitem[{Sattler et~al.(2021)Sattler, M{\"u}ller, and Samek}]{sattlerClusteredFederatedLearning2021}
\bibinfo{author}{F.~Sattler}, \bibinfo{author}{K.-R. M{\"u}ller}, \bibinfo{author}{W.~Samek},
\newblock \bibinfo{title}{Clustered {{Federated Learning}}: {{Model-Agnostic Distributed Multitask Optimization Under Privacy Constraints}}},
\newblock \bibinfo{journal}{IEEE Transactions on Neural Networks and Learning Systems} \bibinfo{volume}{32} (\bibinfo{year}{2021}) \bibinfo{pages}{3710--3722}.
\bibitem[{Frenay and Verleysen(2014)}]{frenayClassificationPresenceLabel2014}
\bibinfo{author}{B.~Frenay}, \bibinfo{author}{M.~Verleysen},
\newblock \bibinfo{title}{Classification in the {{Presence}} of {{Label Noise}}: {{A Survey}}},
\newblock \bibinfo{journal}{IEEE Transactions on Neural Networks and Learning Systems} \bibinfo{volume}{25} (\bibinfo{year}{2014}) \bibinfo{pages}{845--869}.
\bibitem[{Baccour et~al.(2022)Baccour, Mhaisen, Abdellatif, Erbad, Mohamed, Hamdi, and Guizani}]{baccourPervasiveAIIoT2022}
\bibinfo{author}{E.~Baccour}, \bibinfo{author}{N.~Mhaisen}, \bibinfo{author}{A.~A. Abdellatif}, \bibinfo{author}{A.~Erbad}, \bibinfo{author}{A.~Mohamed}, \bibinfo{author}{M.~Hamdi}, \bibinfo{author}{M.~Guizani},
\newblock \bibinfo{title}{Pervasive {{AI}} for {{IoT}} applications: {{A Survey}} on {{ Resource-efficient Distributed Artificial Intelligence}}},
\newblock \bibinfo{journal}{IEEE Communications Surveys \& Tutorials} \bibinfo{volume}{24} (\bibinfo{year}{2022}) \bibinfo{pages}{2366--2418}. \href{http://arxiv.org/abs/2105.01798}{{\tt arXiv:2105.01798}}.
\bibitem[{Song et~al.(2023)Song, Kim, Park, Shin, and Lee}]{songLearningNoisyLabels2023}
\bibinfo{author}{H.~Song}, \bibinfo{author}{M.~Kim}, \bibinfo{author}{D.~Park}, \bibinfo{author}{Y.~Shin}, \bibinfo{author}{J.-G. Lee},
\newblock \bibinfo{title}{Learning {{From Noisy Labels With Deep Neural Networks}}: {{A Survey} }},
\newblock \bibinfo{journal}{IEEE Transactions on Neural Networks and Learning Systems} \bibinfo{volume}{34} (\bibinfo{year}{2023}) \bibinfo{pages}{8135--8153}.
\bibitem[{Wan et~al.(2023)Wan, Wang, Xie, Huang, Chen, and Li}]{wanUnlockingPowerOpen2023}
\bibinfo{author}{W.~Wan}, \bibinfo{author}{X.~Wang}, \bibinfo{author}{M.~Xie}, \bibinfo{author}{S.~Huang}, \bibinfo{author}{S.~Chen}, \bibinfo{author}{S.~Li}, \bibinfo{title}{Unlocking the {{Power}} of {{Open Set}} : {{A New Perspective}} for { {Open-set Noisy Label Learning}}}, \bibinfo{year}{2023}. \href{http://arxiv.org/abs/2305.04203}{{\tt arXiv:2305.04203}}.
\bibitem[{Albert et~al.(2022)Albert, Ortego, Arazo, O'Connor, and McGuinness}]{albertAddressingOutofDistributionLabel2022}
\bibinfo{author}{P.~Albert}, \bibinfo{author}{D.~Ortego}, \bibinfo{author}{E.~Arazo}, \bibinfo{author}{N.~E. O'Connor}, \bibinfo{author}{K.~McGuinness},
\newblock \bibinfo{title}{Addressing {{Out-of-Distribution Label Noise}} in {{Webly-Labelled Data}}},
\newblock in: \bibinfo{booktitle}{Proceedings of the {{IEEE}}/{{CVF Winter Conference}} on {{ Applications}} of {{Computer Vision}}}, \bibinfo{year}{2022}, pp. \bibinfo{pages}{392--401}.
\bibitem[{Algan and Ulusoy(2020)}]{alganLabelNoiseTypes2020}
\bibinfo{author}{G.~Algan}, \bibinfo{author}{{\.I}.~Ulusoy}, \bibinfo{title}{Label {{Noise Types}} and {{Their Effects}} on {{Deep Learning}}}, \bibinfo{year}{2020}. \href{http://arxiv.org/abs/2003.10471}{{\tt arXiv:2003.10471}}.
\bibitem[{Shanthini et~al.(2019)Shanthini, Vinodhini, Chandrasekaran, and Supraja}]{shanthiniTaxonomyImpactLabel2019}
\bibinfo{author}{A.~Shanthini}, \bibinfo{author}{G.~Vinodhini}, \bibinfo{author}{R.~M. Chandrasekaran}, \bibinfo{author}{P.~Supraja},
\newblock \bibinfo{title}{A taxonomy on impact of label noise and feature noise using machine learning techniques},
\newblock \bibinfo{journal}{Soft Computing} \bibinfo{volume}{23} (\bibinfo{year}{2019}) \bibinfo{pages}{8597--8607}.
\bibitem[{Liang et~al.(2022)Liang, Liu, and Yao}]{liangReviewSurveyLearning2022}
\bibinfo{author}{X.~Liang}, \bibinfo{author}{X.~Liu}, \bibinfo{author}{L.~Yao},
\newblock \bibinfo{title}{Review--{{A Survey}} of {{Learning}} from {{Noisy Labels}}},
\newblock \bibinfo{journal}{ECS Sensors Plus} \bibinfo{volume}{1} (\bibinfo{year}{2022}) \bibinfo{pages}{021401}.
\bibitem[{Algan and Ulusoy(2021)}]{alganImageClassificationDeep2021}
\bibinfo{author}{G.~Algan}, \bibinfo{author}{I.~Ulusoy},
\newblock \bibinfo{title}{Image classification with deep learning in the presence of noisy labels: {{A}} survey},
\newblock \bibinfo{journal}{Knowledge-Based Systems} \bibinfo{volume}{215} (\bibinfo{year}{2021}) \bibinfo{pages}{106771}.
\bibitem[{Barry et~al.(2023)Barry, Han, and Demartini}]{barryImpactDataQuality2023}
\bibinfo{author}{A.~Barry}, \bibinfo{author}{L.~Han}, \bibinfo{author}{G.~Demartini},
\newblock \bibinfo{title}{On the {{Impact}} of {{Data Quality}} on {{Image Classification Fairness}}},
\newblock in: \bibinfo{booktitle}{Proceedings of the 46th {{International ACM SIGIR Conference}} on {{Research}} and {{Development}} in {{Information Retrieval}}}, {{SIGIR}} '23, \bibinfo{publisher}{Association for Computing Machinery}, \bibinfo{address}{New York, NY, USA}, \bibinfo{year}{2023}, pp. \bibinfo{pages}{2225--2229}.
\bibitem[{Zeng et~al.(2021)Zeng, Liu, Tang, and Chen}]{zengNoiseUsefulExploiting2021}
\bibinfo{author}{Z.~Zeng}, \bibinfo{author}{Y.~Liu}, \bibinfo{author}{W.~Tang}, \bibinfo{author}{F.~Chen},
\newblock \bibinfo{title}{Noise {{Is Useful}}: {{Exploiting Data Diversity}} for {{Edge Intelligence}}},
\newblock \bibinfo{journal}{IEEE Wireless Communications Letters} \bibinfo{volume}{10} (\bibinfo{year}{2021}) \bibinfo{pages}{957--961}.
\bibitem[{Deng(2012)}]{dengMNISTDatabaseHandwritten2012}
\bibinfo{author}{L.~Deng},
\newblock \bibinfo{title}{The {{MNIST Database}} of {{Handwritten Digit Images}} for {{Machine Learning Research}} [{{Best}} of the {{Web}}]},
\newblock \bibinfo{journal}{IEEE Signal Processing Magazine} \bibinfo{volume}{29} (\bibinfo{year}{2012}) \bibinfo{pages}{141--142}.
\bibitem[{Jafarigol and Trafalis(2023)}]{jafarigolParadoxNoiseEmpirical2023}
\bibinfo{author}{E.~Jafarigol}, \bibinfo{author}{T.~Trafalis}, \bibinfo{title}{The {{Paradox}} of {{Noise}}: {{An Empirical Study}} of {{ Noise-Infusion Mechanisms}} to {{Improve Generalization}}, {{Stability }}, and {{Privacy}} in {{Federated Learning}}}, \bibinfo{year}{2023}. \href{http://arxiv.org/abs/2311.05790}{{\tt arXiv:2311.05790}}.
\bibitem[{Krizhevsky(????)}]{krizhevskyLearningMultipleLayers}
\bibinfo{author}{A.~Krizhevsky},
\newblock \bibinfo{title}{Learning {{Multiple Layers}} of {{Features}} from {{Tiny Images}}}  (????).
\bibitem[{Yao(2019)}]{yaoDeepLearningNoisy2019}
\bibinfo{author}{J.~Yao}, \bibinfo{title}{Deep Learning with Noisy Supervision}, \bibinfo{type}{Thesis}, \bibinfo{year}{2019}.
\bibitem[{Tuor et~al.(2021)Tuor, Wang, Ko, Liu, and Leung}]{tuorOvercomingNoisyIrrelevant2021}
\bibinfo{author}{T.~Tuor}, \bibinfo{author}{S.~Wang}, \bibinfo{author}{B.~J. Ko}, \bibinfo{author}{C.~Liu}, \bibinfo{author}{K.~K. Leung},
\newblock \bibinfo{title}{Overcoming {{Noisy}} and {{Irrelevant Data}} in {{Federated Learning} }},
\newblock in: \bibinfo{booktitle}{2020 25th {{International Conference}} on {{Pattern Recognition}} ({{ICPR}})}, \bibinfo{year}{2021}, pp. \bibinfo{pages}{5020--5027}.
\bibitem[{Yang et~al.(2022)Yang, Qian, Wang, Zhou, and Zhu}]{yangClientSelectionFederated2022}
\bibinfo{author}{M.~Yang}, \bibinfo{author}{H.~Qian}, \bibinfo{author}{X.~Wang}, \bibinfo{author}{Y.~Zhou}, \bibinfo{author}{H.~Zhu},
\newblock \bibinfo{title}{Client {{Selection}} for {{Federated Learning With Label Noise}}},
\newblock \bibinfo{journal}{IEEE Transactions on Vehicular Technology} \bibinfo{volume}{71} (\bibinfo{year}{2022}) \bibinfo{pages}{2193--2197}.
\bibitem[{Ji et~al.(2024)Ji, Zhu, Xi, Gadyatskaya, Song, Cai, and Liu}]{jiFedFixerMitigatingHeterogeneous2024}
\bibinfo{author}{X.~Ji}, \bibinfo{author}{Z.~Zhu}, \bibinfo{author}{W.~Xi}, \bibinfo{author}{O.~Gadyatskaya}, \bibinfo{author}{Z.~Song}, \bibinfo{author}{Y.~Cai}, \bibinfo{author}{Y.~Liu},
\newblock \bibinfo{title}{{{FedFixer}}: {{Mitigating Heterogeneous Label Noise}} in {{Federated Learning}}},
\newblock \bibinfo{journal}{Proceedings of the AAAI Conference on Artificial Intelligence} \bibinfo{volume}{38} (\bibinfo{year}{2024}) \bibinfo{pages}{12830--12838}.
\bibitem[{Tam et~al.(2023)Tam, Li, Han, Xu, and Fu}]{tamFederatedNoisyClient2023}
\bibinfo{author}{K.~Tam}, \bibinfo{author}{L.~Li}, \bibinfo{author}{B.~Han}, \bibinfo{author}{C.~Xu}, \bibinfo{author}{H.~Fu},
\newblock \bibinfo{title}{Federated {{Noisy Client Learning}}},
\newblock \bibinfo{journal}{IEEE Transactions on Neural Networks and Learning Systems}  (\bibinfo{year}{2023}) \bibinfo{pages}{1--14}.
\bibitem[{Gamberger et~al.(1996)Gamberger, Lavra{\v c}, and D{\v z}eroski}]{gambergerNoiseEliminationInductive1996}
\bibinfo{author}{D.~Gamberger}, \bibinfo{author}{N.~Lavra{\v c}}, \bibinfo{author}{S.~D{\v z}eroski},
\newblock \bibinfo{title}{Noise elimination in inductive concept learning: {{A}} case study in medical diagnosis},
\newblock in: \bibinfo{editor}{S.~Arikawa}, \bibinfo{editor}{A.~K. Sharma} (Eds.), \bibinfo{booktitle}{Algorithmic {{Learning Theory}}}, \bibinfo{publisher}{Springer}, \bibinfo{address}{Berlin, Heidelberg}, \bibinfo{year}{1996}, pp. \bibinfo{pages}{199--212}.
\bibitem[{Pleiss et~al.(2020)Pleiss, Zhang, Elenberg, and Weinberger}]{pleissIdentifyingMislabeledData2020}
\bibinfo{author}{G.~Pleiss}, \bibinfo{author}{T.~Zhang}, \bibinfo{author}{E.~Elenberg}, \bibinfo{author}{K.~Q. Weinberger},
\newblock \bibinfo{title}{Identifying {{Mislabeled Data}} using the {{Area Under}} the {{Margin Ranking}}},
\newblock in: \bibinfo{booktitle}{Advances in {{Neural Information Processing Systems}}}, volume~\bibinfo{volume}{33}, \bibinfo{publisher}{Curran Associates, Inc.}, \bibinfo{year}{2020}, pp. \bibinfo{pages}{17044--17056}.
\bibitem[{Gamberger et~al.(1999)Gamberger, Lavrac, and Groselj}]{gambergerExperimentsNoiseLtering1999}
\bibinfo{author}{D.~Gamberger}, \bibinfo{author}{N.~Lavrac}, \bibinfo{author}{C.~Groselj},
\newblock \bibinfo{title}{Experiments with noise ltering in a medical domain},
\newblock in: \bibinfo{booktitle}{Proceedings of the {{Sixteenth International Conference}} on {{ Machine Learning}} \{(\vphantom\}{{ICML}}\vphantom\{\} 1999), {{ Bled}}, {{Slovenia}}, {{June}} 27 - 30, 1999}, \bibinfo{publisher}{Morgan Kaufmann}, \bibinfo{year}{1999}, pp. \bibinfo{pages}{143--151}.
\bibitem[{Brodley and Friedl(1999)}]{brodleyIdentifyingMislabeledTraining1999}
\bibinfo{author}{C.~E. Brodley}, \bibinfo{author}{M.~A. Friedl},
\newblock \bibinfo{title}{Identifying {{Mislabeled Training Data}}},
\newblock \bibinfo{journal}{Journal of Artificial Intelligence Research} \bibinfo{volume}{11} (\bibinfo{year}{1999}) \bibinfo{pages}{131--167}.
\bibitem[{Zhu et~al.(2003)Zhu, Wu, and Chen}]{zhuEliminatingClassNoise2003}
\bibinfo{author}{X.~Zhu}, \bibinfo{author}{X.~Wu}, \bibinfo{author}{Q.~Chen},
\newblock \bibinfo{title}{Eliminating {{Class Noise}} in {{Large Datasets}}},
\newblock in: \bibinfo{booktitle}{Proceedings, {{Twentieth International Conference}} on {{Machine Learning}}}, volume~\bibinfo{volume}{2}, \bibinfo{year}{2003}, pp. \bibinfo{pages}{920--927}.
\bibitem[{Ma et~al.(2021)Ma, Pei, Zhou, Zhu, Wang, and Ji}]{maFederatedDataCleaning2021}
\bibinfo{author}{L.~Ma}, \bibinfo{author}{Q.~Pei}, \bibinfo{author}{L.~Zhou}, \bibinfo{author}{H.~Zhu}, \bibinfo{author}{L.~Wang}, \bibinfo{author}{Y.~Ji},
\newblock \bibinfo{title}{Federated {{Data Cleaning}}: {{Collaborative}} and {{ Privacy-Preserving Data Cleaning}} for {{Edge Intelligence}}},
\newblock \bibinfo{journal}{IEEE Internet of Things Journal} \bibinfo{volume}{8} (\bibinfo{year}{2021}) \bibinfo{pages}{6757--6770}.
\bibitem[{Azadi et~al.(2016)Azadi, Feng, Jegelka, and Darrell}]{azadiAuxiliaryImageRegularization2016}
\bibinfo{author}{S.~Azadi}, \bibinfo{author}{J.~Feng}, \bibinfo{author}{S.~Jegelka}, \bibinfo{author}{T.~Darrell}, \bibinfo{title}{Auxiliary {{Image Regularization}} for {{Deep CNNs}} with {{Noisy Labels}}}, \bibinfo{year}{2016}. \href{http://arxiv.org/abs/1511.07069}{{\tt arXiv:1511.07069}}.
\bibitem[{Yu et~al.(2022)Yu, Zhu, Li, Hong, Zhang, Ye, Huang, and He}]{yuRegularizationPenaltyOptimization2022}
\bibinfo{author}{R.~Yu}, \bibinfo{author}{H.~Zhu}, \bibinfo{author}{K.~Li}, \bibinfo{author}{L.~Hong}, \bibinfo{author}{R.~Zhang}, \bibinfo{author}{N.~Ye}, \bibinfo{author}{S.-L. Huang}, \bibinfo{author}{X.~He},
\newblock \bibinfo{title}{Regularization {{Penalty Optimization}} for {{Addressing Data Quality Variance}} in {{OoD Algorithms}}},
\newblock \bibinfo{journal}{Proceedings of the AAAI Conference on Artificial Intelligence} \bibinfo{volume}{36} (\bibinfo{year}{2022}) \bibinfo{pages}{8945--8953}.
\bibitem[{Liu and Tao(2016)}]{liuClassificationNoisyLabels2016}
\bibinfo{author}{T.~Liu}, \bibinfo{author}{D.~Tao},
\newblock \bibinfo{title}{Classification with {{Noisy Labels}} by {{Importance Reweighting}}},
\newblock \bibinfo{journal}{IEEE Transactions on Pattern Analysis and Machine Intelligence} \bibinfo{volume}{38} (\bibinfo{year}{2016}) \bibinfo{pages}{447--461}.
\bibitem[{Chen et~al.(2023)Chen, Ang, Chen, and Wang}]{chenRobustFederatedLearning2023}
\bibinfo{author}{L.~Chen}, \bibinfo{author}{F.~Ang}, \bibinfo{author}{Y.~Chen}, \bibinfo{author}{W.~Wang},
\newblock \bibinfo{title}{Robust {{Federated Learning With Noisy Labeled Data Through Loss Function Correction}}},
\newblock \bibinfo{journal}{IEEE Transactions on Network Science and Engineering} \bibinfo{volume}{10} (\bibinfo{year}{2023}) \bibinfo{pages}{1501--1511}.
\bibitem[{Manwani and Sastry(2013)}]{manwaniNoiseToleranceRisk2013}
\bibinfo{author}{N.~Manwani}, \bibinfo{author}{P.~S. Sastry},
\newblock \bibinfo{title}{Noise {{Tolerance Under Risk Minimization}}},
\newblock \bibinfo{journal}{IEEE Transactions on Cybernetics} \bibinfo{volume}{43} (\bibinfo{year}{2013}) \bibinfo{pages}{1146--1151}.
\bibitem[{Xu and Lyu(2020)}]{xuReputationMechanismAll2020}
\bibinfo{author}{X.~Xu}, \bibinfo{author}{L.~Lyu}, \bibinfo{title}{A {{Reputation Mechanism Is All You Need}}: {{Collaborative Fairness} } and {{Adversarial Robustness}} in {{Federated Learning}}}, \bibinfo{year}{2020}.
\bibitem[{Chen et~al.(2020)Chen, Yang, Qin, Yu, Chan, and Shen}]{chenDealingLabelQuality2020}
\bibinfo{author}{Y.~Chen}, \bibinfo{author}{X.~Yang}, \bibinfo{author}{X.~Qin}, \bibinfo{author}{H.~Yu}, \bibinfo{author}{P.~Chan}, \bibinfo{author}{Z.~Shen},
\newblock \bibinfo{title}{Dealing with {{Label Quality Disparity}} in {{Federated Learning}}},
\newblock in: \bibinfo{editor}{Q.~Yang}, \bibinfo{editor}{L.~Fan}, \bibinfo{editor}{H.~Yu} (Eds.), \bibinfo{booktitle}{Federated {{Learning}}: {{Privacy}} and {{Incentive}}}, \bibinfo{publisher}{Springer International Publishing}, \bibinfo{address}{Cham}, \bibinfo{year}{2020}, pp. \bibinfo{pages}{108--121}.
\bibitem[{Vucinich and Zhu(2023)}]{vucinichCurrentStateChallenges2023}
\bibinfo{author}{S.~Vucinich}, \bibinfo{author}{Q.~Zhu},
\newblock \bibinfo{title}{The {{Current State}} and {{Challenges}} of {{Fairness}} in {{ Federated Learning}}},
\newblock \bibinfo{journal}{IEEE Access} \bibinfo{volume}{11} (\bibinfo{year}{2023}) \bibinfo{pages}{80903--80914}.
\bibitem[{{Grgi{\'c}-Hla{\v c}a} et~al.(2018){Grgi{\'c}-Hla{\v c}a}, Zafar, Gummadi, and Weller}]{grgic-hlacaDistributiveFairnessAlgorithmic2018}
\bibinfo{author}{N.~{Grgi{\'c}-Hla{\v c}a}}, \bibinfo{author}{M.~Zafar}, \bibinfo{author}{K.~Gummadi}, \bibinfo{author}{A.~Weller},
\newblock \bibinfo{title}{Beyond distributive fairness in algorithmic decision making: {{ Feature}} selection for procedurally fair learning},
\newblock in: \bibinfo{booktitle}{32nd {{AAAI Conference}} on {{Artificial Intelligence}}, {{AAAI}} 2018}, \bibinfo{year}{2018}, pp. \bibinfo{pages}{51--60}.
\bibitem[{Weerts et~al.(2023)Weerts, Dud{\'i}k, Edgar, Jalali, Lutz, and Madaio}]{weertsFairlearnAssessingImproving2023}
\bibinfo{author}{H.~Weerts}, \bibinfo{author}{M.~Dud{\'i}k}, \bibinfo{author}{R.~Edgar}, \bibinfo{author}{A.~Jalali}, \bibinfo{author}{R.~Lutz}, \bibinfo{author}{M.~Madaio},
\newblock \bibinfo{title}{Fairlearn: {{Assessing}} and {{Improving Fairness}} of {{AI Systems}}},
\newblock \bibinfo{journal}{Journal of Machine Learning Research} \bibinfo{volume}{24} (\bibinfo{year}{2023}) \bibinfo{pages}{1--8}.
\bibitem[{Castelnovo et~al.(2022)Castelnovo, Crupi, Greco, Regoli, Penco, and Cosentini}]{castelnovoClarificationNuancesFairness2022}
\bibinfo{author}{A.~Castelnovo}, \bibinfo{author}{R.~Crupi}, \bibinfo{author}{G.~Greco}, \bibinfo{author}{D.~Regoli}, \bibinfo{author}{I.~G. Penco}, \bibinfo{author}{A.~C. Cosentini},
\newblock \bibinfo{title}{A clarification of the nuances in the fairness metrics landscape},
\newblock \bibinfo{journal}{Scientific Reports} \bibinfo{volume}{12} (\bibinfo{year}{2022}) \bibinfo{pages}{4209}.
\bibitem[{Verma and Rubin(2018)}]{vermaFairnessDefinitionsExplained2018}
\bibinfo{author}{S.~Verma}, \bibinfo{author}{J.~Rubin},
\newblock \bibinfo{title}{Fairness definitions explained},
\newblock in: \bibinfo{booktitle}{Proceedings of the {{International Workshop}} on {{Software Fairness}}}, \bibinfo{publisher}{ACM}, \bibinfo{address}{Gothenburg Sweden}, \bibinfo{year}{2018}, pp. \bibinfo{pages}{1--7}.
\bibitem[{Agarwal et~al.(2018)Agarwal, Beygelzimer, Dudik, Langford, and Wallach}]{agarwalReductionsApproachFair2018}
\bibinfo{author}{A.~Agarwal}, \bibinfo{author}{A.~Beygelzimer}, \bibinfo{author}{M.~Dudik}, \bibinfo{author}{J.~Langford}, \bibinfo{author}{H.~Wallach},
\newblock \bibinfo{title}{A {{Reductions Approach}} to {{Fair Classification}}},
\newblock in: \bibinfo{booktitle}{Proceedings of the 35th {{International Conference}} on {{Machine Learning}}}, \bibinfo{publisher}{PMLR}, \bibinfo{year}{2018}, pp. \bibinfo{pages}{60--69}.
\bibitem[{{Corbett-Davies} et~al.(2017){Corbett-Davies}, Pierson, Feller, Goel, and Huq}]{corbett-daviesAlgorithmicDecisionMaking2017}
\bibinfo{author}{S.~{Corbett-Davies}}, \bibinfo{author}{E.~Pierson}, \bibinfo{author}{A.~Feller}, \bibinfo{author}{S.~Goel}, \bibinfo{author}{A.~Huq},
\newblock \bibinfo{title}{Algorithmic {{Decision Making}} and the {{Cost}} of {{Fairness}}},
\newblock in: \bibinfo{booktitle}{Proceedings of the 23rd {{ACM SIGKDD International Conference}} on {{Knowledge Discovery}} and {{Data Mining}}}, {{KDD}} '17, \bibinfo{publisher}{Association for Computing Machinery}, \bibinfo{address}{New York, NY, USA}, \bibinfo{year}{2017}, pp. \bibinfo{pages}{797--806}.
\bibitem[{Donini et~al.(2018)Donini, Oneto, {Ben-David}, { Shawe-Taylor}, and Pontil}]{doniniEmpiricalRiskMinimization2018}
\bibinfo{author}{M.~Donini}, \bibinfo{author}{L.~Oneto}, \bibinfo{author}{S.~{Ben-David}}, \bibinfo{author}{J.~S. { Shawe-Taylor}}, \bibinfo{author}{M.~Pontil},
\newblock \bibinfo{title}{Empirical {{Risk Minimization Under Fairness Constraints}}},
\newblock in: \bibinfo{booktitle}{Advances in {{Neural Information Processing Systems}}}, volume~\bibinfo{volume}{31}, \bibinfo{publisher}{Curran Associates, Inc.}, \bibinfo{year}{2018}.
\bibitem[{{Grgic-Hlaca} et~al.(2016){Grgic-Hlaca}, Zafar, Gummadi, and Weller}]{grgic-hlacaCaseProcessFairness2016}
\bibinfo{author}{N.~{Grgic-Hlaca}}, \bibinfo{author}{M.~B. Zafar}, \bibinfo{author}{K.~Gummadi}, \bibinfo{author}{A.~Weller},
\newblock \bibinfo{title}{The {{Case}} for {{Process Fairness}} in {{Learning}}: {{Feature Selection}} for {{Fair Decision Making}}},
\newblock \bibinfo{year}{2016}.
\bibitem[{Dwork et~al.(2012)Dwork, Hardt, Pitassi, Reingold, and Zemel}]{dworkFairnessAwareness2012}
\bibinfo{author}{C.~Dwork}, \bibinfo{author}{M.~Hardt}, \bibinfo{author}{T.~Pitassi}, \bibinfo{author}{O.~Reingold}, \bibinfo{author}{R.~Zemel},
\newblock \bibinfo{title}{Fairness through awareness},
\newblock in: \bibinfo{booktitle}{Proceedings of the 3rd {{Innovations}} in {{Theoretical Computer Science Conference}}}, {{ITCS}} '12, \bibinfo{publisher}{Association for Computing Machinery}, \bibinfo{address}{New York, NY, USA}, \bibinfo{year}{2012}, pp. \bibinfo{pages}{214--226}.
\bibitem[{Kusner et~al.(2017)Kusner, Loftus, Russell, and Silva}]{kusnerCounterfactualFairness2017}
\bibinfo{author}{M.~J. Kusner}, \bibinfo{author}{J.~Loftus}, \bibinfo{author}{C.~Russell}, \bibinfo{author}{R.~Silva},
\newblock \bibinfo{title}{Counterfactual {{Fairness}}},
\newblock in: \bibinfo{booktitle}{Advances in {{Neural Information Processing Systems}}}, volume~\bibinfo{volume}{30}, \bibinfo{publisher}{Curran Associates, Inc.}, \bibinfo{year}{2017}.
\bibitem[{Kearns et~al.(2018)Kearns, Neel, Roth, and Wu}]{kearnsPreventingFairnessGerrymandering2018}
\bibinfo{author}{M.~Kearns}, \bibinfo{author}{S.~Neel}, \bibinfo{author}{A.~Roth}, \bibinfo{author}{Z.~S. Wu},
\newblock \bibinfo{title}{Preventing {{Fairness Gerrymandering}}: {{Auditing}} and {{Learning}} for {{Subgroup Fairness}}},
\newblock in: \bibinfo{booktitle}{Proceedings of the 35th {{International Conference}} on {{Machine Learning}}}, \bibinfo{publisher}{PMLR}, \bibinfo{year}{2018}, pp. \bibinfo{pages}{2564--2572}.
\bibitem[{Kearns et~al.(2019)Kearns, Neel, Roth, and Wu}]{kearnsEmpiricalStudyRich2019}
\bibinfo{author}{M.~Kearns}, \bibinfo{author}{S.~Neel}, \bibinfo{author}{A.~Roth}, \bibinfo{author}{Z.~S. Wu},
\newblock \bibinfo{title}{An {{Empirical Study}} of {{Rich Subgroup Fairness}} for {{Machine Learning}}},
\newblock in: \bibinfo{booktitle}{Proceedings of the {{Conference}} on {{Fairness}}, {{ Accountability}}, and {{Transparency}}}, {{FAT}}* '19, \bibinfo{publisher}{Association for Computing Machinery}, \bibinfo{address}{New York, NY, USA}, \bibinfo{year}{2019}, pp. \bibinfo{pages}{100--109}.
\bibitem[{Lyu et~al.(2020)Lyu, Yu, Nandakumar, Li, Ma, Jin, Yu, and Ng}]{lyuFairPrivacyPreservingFederated2020}
\bibinfo{author}{L.~Lyu}, \bibinfo{author}{J.~Yu}, \bibinfo{author}{K.~Nandakumar}, \bibinfo{author}{Y.~Li}, \bibinfo{author}{X.~Ma}, \bibinfo{author}{J.~Jin}, \bibinfo{author}{H.~Yu}, \bibinfo{author}{K.~S. Ng},
\newblock \bibinfo{title}{Towards {{Fair}} and {{Privacy-Preserving Federated Deep Models}}},
\newblock \bibinfo{journal}{IEEE Transactions on Parallel and Distributed Systems} \bibinfo{volume}{31} (\bibinfo{year}{2020}) \bibinfo{pages}{2524--2541}.
\bibitem[{Donahue and Kleinberg(2023)}]{donahueFairnessModelsharingGames2023}
\bibinfo{author}{K.~Donahue}, \bibinfo{author}{J.~Kleinberg},
\newblock \bibinfo{title}{Fairness in model-sharing games},
\newblock in: \bibinfo{booktitle}{Proceedings of the {{ACM Web Conference}} 2023}, {{WWW}} '23, \bibinfo{publisher}{Association for Computing Machinery}, \bibinfo{address}{New York, NY, USA}, \bibinfo{year}{2023}, pp. \bibinfo{pages}{3775--3783}.
\bibitem[{Ray~Chaudhury et~al.(2022)Ray~Chaudhury, Li, Kang, Li, and Mehta}]{raychaudhuryFairnessFederatedLearning2022}
\bibinfo{author}{B.~Ray~Chaudhury}, \bibinfo{author}{L.~Li}, \bibinfo{author}{M.~Kang}, \bibinfo{author}{B.~Li}, \bibinfo{author}{R.~Mehta},
\newblock \bibinfo{title}{Fairness in {{Federated Learning}} via {{Core-Stability}}},
\newblock \bibinfo{journal}{Advances in Neural Information Processing Systems} \bibinfo{volume}{35} (\bibinfo{year}{2022}) \bibinfo{pages}{5738--5750}.
\bibitem[{Mohri et~al.(2019)Mohri, Sivek, and Suresh}]{mohriAgnosticFederatedLearning2019}
\bibinfo{author}{M.~Mohri}, \bibinfo{author}{G.~Sivek}, \bibinfo{author}{A.~T. Suresh},
\newblock \bibinfo{title}{Agnostic {{Federated Learning}}},
\newblock in: \bibinfo{booktitle}{Proceedings of the 36th {{International Conference}} on {{Machine Learning}}}, \bibinfo{publisher}{PMLR}, \bibinfo{year}{2019}, pp. \bibinfo{pages}{4615--4625}.
\bibitem[{Zhang et~al.(2022)Zhang, Kuang, Liu, Chen, Wu, Wu, Lu, Shao, and Xiao}]{zhangUnifiedGroupFairness2022}
\bibinfo{author}{F.~Zhang}, \bibinfo{author}{K.~Kuang}, \bibinfo{author}{Y.~Liu}, \bibinfo{author}{L.~Chen}, \bibinfo{author}{C.~Wu}, \bibinfo{author}{F.~Wu}, \bibinfo{author}{J.~Lu}, \bibinfo{author}{Y.~Shao}, \bibinfo{author}{J.~Xiao}, \bibinfo{title}{Unified {{Group Fairness}} on {{Federated Learning}}}, \bibinfo{year}{2022}. \href{http://arxiv.org/abs/2111.04986}{{\tt arXiv:2111.04986}}.
\bibitem[{Lyu et~al.(2020)Lyu, Li, Nandakumar, Yu, and Ma}]{lyuHowDemocratiseProtect2020}
\bibinfo{author}{L.~Lyu}, \bibinfo{author}{Y.~Li}, \bibinfo{author}{K.~Nandakumar}, \bibinfo{author}{J.~Yu}, \bibinfo{author}{X.~Ma},
\newblock \bibinfo{title}{How to {{Democratise}} and {{Protect AI}}: {{Fair}} and {{ Differentially Private Decentralised Deep Learning}}},
\newblock \bibinfo{journal}{IEEE Transactions on Dependable and Secure Computing}  (\bibinfo{year}{2020}) \bibinfo{pages}{1--1}.
\bibitem[{Shapley and Shubik(1969)}]{shapleyMarketGames1969}
\bibinfo{author}{L.~S. Shapley}, \bibinfo{author}{M.~Shubik},
\newblock \bibinfo{title}{On market games},
\newblock \bibinfo{journal}{Journal of Economic Theory} \bibinfo{volume}{1} (\bibinfo{year}{1969}) \bibinfo{pages}{9--25}.
\bibitem[{Brunet et~al.(2019)Brunet, {Alkalay-Houlihan}, Anderson, and Zemel}]{brunetUnderstandingOriginsBias2019}
\bibinfo{author}{M.-E. Brunet}, \bibinfo{author}{C.~{Alkalay-Houlihan}}, \bibinfo{author}{A.~Anderson}, \bibinfo{author}{R.~Zemel},
\newblock \bibinfo{title}{Understanding the {{Origins}} of {{Bias}} in {{Word Embeddings}}},
\newblock in: \bibinfo{booktitle}{Proceedings of the 36th {{International Conference}} on {{Machine Learning}}}, \bibinfo{publisher}{PMLR}, \bibinfo{year}{2019}, pp. \bibinfo{pages}{803--811}.
\bibitem[{Calmon et~al.(2017)Calmon, Wei, Vinzamuri, Natesan~Ramamurthy, and Varshney}]{calmonOptimizedPreProcessingDiscrimination2017}
\bibinfo{author}{F.~Calmon}, \bibinfo{author}{D.~Wei}, \bibinfo{author}{B.~Vinzamuri}, \bibinfo{author}{K.~Natesan~Ramamurthy}, \bibinfo{author}{K.~R. Varshney},
\newblock \bibinfo{title}{Optimized {{Pre-Processing}} for {{Discrimination Prevention}}},
\newblock in: \bibinfo{booktitle}{Advances in {{Neural Information Processing Systems}}}, volume~\bibinfo{volume}{30}, \bibinfo{publisher}{Curran Associates, Inc.}, \bibinfo{year}{2017}.
\bibitem[{Abhishek and Abdelaziz(2023)}]{abhishekMachineLearningImbalanced2023}
\bibinfo{author}{K.~Abhishek}, \bibinfo{author}{M.~Abdelaziz}, \bibinfo{title}{Machine Learning for Imbalanced Data: Tackle Imbalanced Datasets Using Machine Learning and Deep Learning Techniques}, \bibinfo{edition}{[first edition]} ed., \bibinfo{publisher}{Packt Publishing Ltd.}, \bibinfo{address}{Birmingham, UK}, \bibinfo{year}{2023}.
\bibitem[{Duan et~al.(2019)Duan, Liu, Chen, Tan, Ren, Qiao, and Liang}]{duanAstraeaSelfBalancingFederated2019}
\bibinfo{author}{M.~Duan}, \bibinfo{author}{D.~Liu}, \bibinfo{author}{X.~Chen}, \bibinfo{author}{Y.~Tan}, \bibinfo{author}{J.~Ren}, \bibinfo{author}{L.~Qiao}, \bibinfo{author}{L.~Liang},
\newblock \bibinfo{title}{Astraea: {{Self-Balancing Federated Learning}} for {{Improving Classification Accuracy}} of {{Mobile Deep Learning Applications}}},
\newblock in: \bibinfo{booktitle}{2019 {{IEEE}} 37th {{International Conference}} on {{Computer Design}} ({{ICCD}})}, \bibinfo{publisher}{IEEE}, \bibinfo{address}{Abu Dhabi, United Arab Emirates}, \bibinfo{year}{2019}, pp. \bibinfo{pages}{246--254}.
\bibitem[{Kamishima et~al.(2012)Kamishima, Akaho, Asoh, and Sakuma}]{kamishimaFairnessAwareClassifierPrejudice2012}
\bibinfo{author}{T.~Kamishima}, \bibinfo{author}{S.~Akaho}, \bibinfo{author}{H.~Asoh}, \bibinfo{author}{J.~Sakuma},
\newblock \bibinfo{title}{Fairness-{{Aware Classifier}} with {{Prejudice Remover Regularizer}}},
\newblock in: \bibinfo{editor}{P.~A. Flach}, \bibinfo{editor}{T.~De~Bie}, \bibinfo{editor}{N.~Cristianini} (Eds.), \bibinfo{booktitle}{Machine {{Learning}} and {{Knowledge Discovery}} in {{Databases}}}, \bibinfo{publisher}{Springer}, \bibinfo{address}{Berlin, Heidelberg}, \bibinfo{year}{2012}, pp. \bibinfo{pages}{35--50}.
\bibitem[{Ezzeldin et~al.(2023)Ezzeldin, Yan, He, Ferrara, and Avestimehr}]{ezzeldinFairFedEnablingGroup2023}
\bibinfo{author}{Y.~H. Ezzeldin}, \bibinfo{author}{S.~Yan}, \bibinfo{author}{C.~He}, \bibinfo{author}{E.~Ferrara}, \bibinfo{author}{A.~S. Avestimehr},
\newblock \bibinfo{title}{{{FairFed}}: {{Enabling Group Fairness}} in {{Federated Learning}}},
\newblock \bibinfo{journal}{Proceedings of the AAAI Conference on Artificial Intelligence} \bibinfo{volume}{37} (\bibinfo{year}{2023}) \bibinfo{pages}{7494--7502}.
\bibitem[{Shi et~al.(2023)Shi, Yu, and Leung}]{shiFairnessAwareFederatedLearning2023}
\bibinfo{author}{Y.~Shi}, \bibinfo{author}{H.~Yu}, \bibinfo{author}{C.~Leung},
\newblock \bibinfo{title}{Towards {{Fairness-Aware Federated Learning}}},
\newblock \bibinfo{journal}{IEEE Transactions on Neural Networks and Learning Systems}  (\bibinfo{year}{2023}) \bibinfo{pages}{1--17}.
\bibitem[{Carlini et~al.(2021)Carlini, Tram{\`e}r, Wallace, Jagielski, {Herbert-Voss}, Lee, Roberts, Brown, Song, Erlingsson, Oprea, and Raffel}]{carliniExtractingTrainingData2021}
\bibinfo{author}{N.~Carlini}, \bibinfo{author}{F.~Tram{\`e}r}, \bibinfo{author}{E.~Wallace}, \bibinfo{author}{M.~Jagielski}, \bibinfo{author}{A.~{Herbert-Voss}}, \bibinfo{author}{K.~Lee}, \bibinfo{author}{A.~Roberts}, \bibinfo{author}{T.~Brown}, \bibinfo{author}{D.~Song}, \bibinfo{author}{{\'U}.~Erlingsson}, \bibinfo{author}{A.~Oprea}, \bibinfo{author}{C.~Raffel},
\newblock \bibinfo{title}{Extracting {{Training Data}} from {{Large Language Models}}},
\newblock in: \bibinfo{booktitle}{30th {{USENIX Security Symposium}} ({{USENIX Security}} 21)}, \bibinfo{year}{2021}, pp. \bibinfo{pages}{2633--2650}.
\bibitem[{Pasquini et~al.(2022)Pasquini, Francati, and Ateniese}]{pasquiniEludingSecureAggregation2022}
\bibinfo{author}{D.~Pasquini}, \bibinfo{author}{D.~Francati}, \bibinfo{author}{G.~Ateniese},
\newblock \bibinfo{title}{Eluding {{Secure Aggregation}} in {{Federated Learning}} via {{Model Inconsistency}}},
\newblock in: \bibinfo{booktitle}{Proceedings of the {{ACM Conference}} on {{Computer}} and {{ Communications Security}}}, \bibinfo{year}{2022}, pp. \bibinfo{pages}{2429--2443}.
\bibitem[{Li et~al.(2023)Li, Yan, Huang, Pan, Lai, Zhang, Chen, and Li}]{liModelArchitectureLevel2023}
\bibinfo{author}{Y.~Li}, \bibinfo{author}{H.~Yan}, \bibinfo{author}{T.~Huang}, \bibinfo{author}{Z.~Pan}, \bibinfo{author}{J.~Lai}, \bibinfo{author}{X.~Zhang}, \bibinfo{author}{K.~Chen}, \bibinfo{author}{J.~Li},
\newblock \bibinfo{title}{Model architecture level privacy leakage in neural networks},
\newblock \bibinfo{journal}{Science China Information Sciences} \bibinfo{volume}{67} (\bibinfo{year}{2023}) \bibinfo{pages}{132101}.
\bibitem[{Balle et~al.(2022)Balle, Cherubin, and Hayes}]{balleReconstructingTrainingData2022}
\bibinfo{author}{B.~Balle}, \bibinfo{author}{G.~Cherubin}, \bibinfo{author}{J.~Hayes},
\newblock \bibinfo{title}{Reconstructing {{Training Data}} with {{Informed Adversaries}}},
\newblock in: \bibinfo{booktitle}{2022 {{IEEE Symposium}} on {{Security}} and {{Privacy}} ({{SP}})}, \bibinfo{publisher}{IEEE}, \bibinfo{address}{San Francisco, CA, USA}, \bibinfo{year}{2022}, pp. \bibinfo{pages}{1138--1156}.
\bibitem[{Shokri et~al.(2017)Shokri, Stronati, Song, and Shmatikov}]{shokriMembershipInferenceAttacks2017}
\bibinfo{author}{R.~Shokri}, \bibinfo{author}{M.~Stronati}, \bibinfo{author}{C.~Song}, \bibinfo{author}{V.~Shmatikov},
\newblock \bibinfo{title}{Membership {{Inference Attacks Against Machine Learning Models}}},
\newblock in: \bibinfo{booktitle}{2017 {{IEEE Symposium}} on {{Security}} and {{Privacy}} ({{SP}})}, \bibinfo{year}{2017}, pp. \bibinfo{pages}{3--18}.
\bibitem[{Belkin et~al.(2019)Belkin, Hsu, Ma, and Mandal}]{belkinReconcilingModernMachinelearning2019}
\bibinfo{author}{M.~Belkin}, \bibinfo{author}{D.~Hsu}, \bibinfo{author}{S.~Ma}, \bibinfo{author}{S.~Mandal},
\newblock \bibinfo{title}{Reconciling modern machine-learning practice and the classical bias--variance trade-off},
\newblock \bibinfo{journal}{Proceedings of the National Academy of Sciences of the United States of America} \bibinfo{volume}{116} (\bibinfo{year}{2019}) \bibinfo{pages}{15849--15854}.
\bibitem[{Belkin(2021)}]{belkinFitFearRemarkable2021}
\bibinfo{author}{M.~Belkin},
\newblock \bibinfo{title}{Fit without fear: Remarkable mathematical phenomena of deep learning through the prism of interpolation},
\newblock \bibinfo{journal}{Acta Numerica} \bibinfo{volume}{30} (\bibinfo{year}{2021}) \bibinfo{pages}{203--248}.
\bibitem[{Mothukuri et~al.(2021)Mothukuri, Parizi, Pouriyeh, Huang, Dehghantanha, and Srivastava}]{mothukuriSurveySecurityPrivacy2021}
\bibinfo{author}{V.~Mothukuri}, \bibinfo{author}{R.~M. Parizi}, \bibinfo{author}{S.~Pouriyeh}, \bibinfo{author}{Y.~Huang}, \bibinfo{author}{A.~Dehghantanha}, \bibinfo{author}{G.~Srivastava},
\newblock \bibinfo{title}{A survey on security and privacy of federated learning},
\newblock \bibinfo{journal}{Future Generation Computer Systems} \bibinfo{volume}{115} (\bibinfo{year}{2021}) \bibinfo{pages}{619--640}.
\bibitem[{Kulynych et~al.(2022)Kulynych, Yaghini, Cherubin, Veale, and Troncoso}]{kulynychDisparateVulnerabilityMembership2022}
\bibinfo{author}{B.~Kulynych}, \bibinfo{author}{M.~Yaghini}, \bibinfo{author}{G.~Cherubin}, \bibinfo{author}{M.~Veale}, \bibinfo{author}{C.~Troncoso},
\newblock \bibinfo{title}{Disparate {{Vulnerability}} to {{Membership Inference Attacks}}},
\newblock \bibinfo{journal}{Proceedings on Privacy Enhancing Technologies}  (\bibinfo{year}{2022}).
\bibitem[{Hu et~al.(2022)Hu, Salcic, Sun, Dobbie, Yu, and Zhang}]{huMembershipInferenceAttacks2022}
\bibinfo{author}{H.~Hu}, \bibinfo{author}{Z.~Salcic}, \bibinfo{author}{L.~Sun}, \bibinfo{author}{G.~Dobbie}, \bibinfo{author}{P.~S. Yu}, \bibinfo{author}{X.~Zhang},
\newblock \bibinfo{title}{Membership {{Inference Attacks}} on {{Machine Learning}}: {{A Survey} }},
\newblock \bibinfo{journal}{ACM Computing Surveys} \bibinfo{volume}{54} (\bibinfo{year}{2022}) \bibinfo{pages}{1--37}.
\bibitem[{Wang et~al.(2023)Wang, Huang, Song, Wu, Xue, and Ren}]{wangPoisoningAssistedPropertyInference2023}
\bibinfo{author}{Z.~Wang}, \bibinfo{author}{Y.~Huang}, \bibinfo{author}{M.~Song}, \bibinfo{author}{L.~Wu}, \bibinfo{author}{F.~Xue}, \bibinfo{author}{K.~Ren},
\newblock \bibinfo{title}{Poisoning-{{Assisted Property Inference Attack Against Federated Learning}}},
\newblock \bibinfo{journal}{IEEE Transactions on Dependable and Secure Computing} \bibinfo{volume}{20} (\bibinfo{year}{2023}) \bibinfo{pages}{3328--3340}.
\bibitem[{Melis et~al.(2019)Melis, Song, De~Cristofaro, and Shmatikov}]{melisExploitingUnintendedFeature2019}
\bibinfo{author}{L.~Melis}, \bibinfo{author}{C.~Song}, \bibinfo{author}{E.~De~Cristofaro}, \bibinfo{author}{V.~Shmatikov},
\newblock \bibinfo{title}{Exploiting {{Unintended Feature Leakage}} in {{Collaborative Learning }}},
\newblock in: \bibinfo{booktitle}{2019 {{IEEE Symposium}} on {{Security}} and {{Privacy}} ({{SP}})}, \bibinfo{publisher}{IEEE}, \bibinfo{address}{San Francisco, CA, USA}, \bibinfo{year}{2019}, pp. \bibinfo{pages}{691--706}.
\bibitem[{Kim et~al.(2023)Kim, Cho, Lee, Bae, and Paek}]{kimExploringClusteredFederated2023}
\bibinfo{author}{H.~Kim}, \bibinfo{author}{Y.~Cho}, \bibinfo{author}{Y.~Lee}, \bibinfo{author}{H.~Bae}, \bibinfo{author}{Y.~Paek},
\newblock \bibinfo{title}{Exploring {{Clustered Federated Learning}}'s {{Vulnerability}} against {{Property Inference Attack}}},
\newblock in: \bibinfo{booktitle}{Proceedings of the 26th {{International Symposium}} on {{Research }} in {{Attacks}}, {{Intrusions}} and {{Defenses}}}, {{RAID}} '23, \bibinfo{publisher}{Association for Computing Machinery}, \bibinfo{address}{New York, NY, USA}, \bibinfo{year}{2023}, pp. \bibinfo{pages}{236--249}.
\bibitem[{Mahloujifar et~al.(2022)Mahloujifar, Ghosh, and Chase}]{mahloujifarPropertyInferencePoisoning2022}
\bibinfo{author}{S.~Mahloujifar}, \bibinfo{author}{E.~Ghosh}, \bibinfo{author}{M.~Chase},
\newblock \bibinfo{title}{Property {{Inference}} from {{Poisoning}}},
\newblock in: \bibinfo{booktitle}{2022 {{IEEE Symposium}} on {{Security}} and {{Privacy}} ({{SP}})}, \bibinfo{year}{2022}, pp. \bibinfo{pages}{1120--1137}.
\bibitem[{Hitaj et~al.(2017)Hitaj, Ateniese, and {Perez-Cruz}}]{hitajDeepModelsGAN2017}
\bibinfo{author}{B.~Hitaj}, \bibinfo{author}{G.~Ateniese}, \bibinfo{author}{F.~{Perez-Cruz}},
\newblock \bibinfo{title}{Deep {{Models Under}} the {{GAN}}: {{Information Leakage}} from {{ Collaborative Deep Learning}}},
\newblock in: \bibinfo{booktitle}{Proceedings of the 2017 {{ACM SIGSAC Conference}} on {{Computer}} and {{Communications Security}}}, {{CCS}} '17, \bibinfo{publisher}{Association for Computing Machinery}, \bibinfo{address}{New York, NY, USA}, \bibinfo{year}{2017}, pp. \bibinfo{pages}{603--618}.
\bibitem[{Wang et~al.(2019)Wang, Song, Zhang, Song, Wang, and Qi}]{wangInferringClassRepresentatives2019}
\bibinfo{author}{Z.~Wang}, \bibinfo{author}{M.~Song}, \bibinfo{author}{Z.~Zhang}, \bibinfo{author}{Y.~Song}, \bibinfo{author}{Q.~Wang}, \bibinfo{author}{H.~Qi},
\newblock \bibinfo{title}{Beyond {{Inferring Class Representatives}}: {{User-Level Privacy Leakage From Federated Learning}}},
\newblock in: \bibinfo{booktitle}{{{IEEE INFOCOM}} 2019 - {{IEEE Conference}} on {{Computer Communications}}}, \bibinfo{year}{2019}, pp. \bibinfo{pages}{2512--2520}.
\bibitem[{El~Mestari et~al.(2024)El~Mestari, Lenzini, and Demirci}]{elmestariPreservingDataPrivacy2024}
\bibinfo{author}{S.~Z. El~Mestari}, \bibinfo{author}{G.~Lenzini}, \bibinfo{author}{H.~Demirci},
\newblock \bibinfo{title}{Preserving data privacy in machine learning systems},
\newblock \bibinfo{journal}{Computers \& Security} \bibinfo{volume}{137} (\bibinfo{year}{2024}) \bibinfo{pages}{103605}.
\bibitem[{Mo et~al.(2021)Mo, Liu, Huang, and Yan}]{moQueryingLittleEnough2021}
\bibinfo{author}{K.~Mo}, \bibinfo{author}{X.~Liu}, \bibinfo{author}{T.~Huang}, \bibinfo{author}{A.~Yan},
\newblock \bibinfo{title}{Querying little is enough: {{Model}} inversion attack via latent information},
\newblock \bibinfo{journal}{International Journal of Intelligent Systems} \bibinfo{volume}{36} (\bibinfo{year}{2021}) \bibinfo{pages}{681--690}.
\bibitem[{Dinur and Nissim(2003)}]{dinurRevealingInformationPreserving2003}
\bibinfo{author}{I.~Dinur}, \bibinfo{author}{K.~Nissim},
\newblock \bibinfo{title}{Revealing information while preserving privacy},
\newblock in: \bibinfo{booktitle}{Proceedings of the Twenty-Second {{ACM SIGMOD-SIGACT-SIGART}} Symposium on {{Principles}} of Database Systems}, {{PODS}} '03, \bibinfo{publisher}{Association for Computing Machinery}, \bibinfo{address}{New York, NY, USA}, \bibinfo{year}{2003}, pp. \bibinfo{pages}{202--210}.
\bibitem[{Zhu and Han(2020)}]{zhuDeepLeakageGradients2020}
\bibinfo{author}{L.~Zhu}, \bibinfo{author}{S.~Han},
\newblock \bibinfo{title}{Deep {{Leakage}} from {{Gradients}}},
\newblock in: \bibinfo{editor}{Q.~Yang}, \bibinfo{editor}{L.~Fan}, \bibinfo{editor}{H.~Yu} (Eds.), \bibinfo{booktitle}{Federated {{Learning}}: {{Privacy}} and {{Incentive}}}, \bibinfo{publisher}{Springer International Publishing}, \bibinfo{address}{Cham}, \bibinfo{year}{2020}, pp. \bibinfo{pages}{17--31}.
\bibitem[{Zhu et~al.(2019)Zhu, Liu, and Han}]{zhuDeepLeakageGradients2019}
\bibinfo{author}{L.~Zhu}, \bibinfo{author}{Z.~Liu}, \bibinfo{author}{S.~Han},
\newblock \bibinfo{title}{Deep {{Leakage}} from {{Gradients}}},
\newblock in: \bibinfo{booktitle}{Advances in {{Neural Information Processing Systems}}}, volume~\bibinfo{volume}{32}, \bibinfo{publisher}{Curran Associates, Inc.}, \bibinfo{year}{2019}.
\bibitem[{Haim et~al.(2022)Haim, Vardi, Yehudai, Shamir, and Irani}]{haimReconstructingTrainingData2022}
\bibinfo{author}{N.~Haim}, \bibinfo{author}{G.~Vardi}, \bibinfo{author}{G.~Yehudai}, \bibinfo{author}{O.~Shamir}, \bibinfo{author}{M.~Irani},
\newblock \bibinfo{title}{Reconstructing {{Training Data From Trained Neural Networks}}},
\newblock \bibinfo{journal}{Advances in Neural Information Processing Systems} \bibinfo{volume}{35} (\bibinfo{year}{2022}) \bibinfo{pages}{22911--22924}.
\bibitem[{Buzaglo et~al.(2023)Buzaglo, Haim, Yehudai, Vardi, Oz, Nikankin, and Irani}]{buzagloDeconstructingDataReconstruction2023}
\bibinfo{author}{G.~Buzaglo}, \bibinfo{author}{N.~Haim}, \bibinfo{author}{G.~Yehudai}, \bibinfo{author}{G.~Vardi}, \bibinfo{author}{Y.~Oz}, \bibinfo{author}{Y.~Nikankin}, \bibinfo{author}{M.~Irani},
\newblock \bibinfo{title}{Deconstructing {{Data Reconstruction}}: {{Multiclass}}, {{Weight Decay}} and {{General Losses}}},
\newblock \bibinfo{journal}{Advances in Neural Information Processing Systems} \bibinfo{volume}{36} (\bibinfo{year}{2023}) \bibinfo{pages}{51515--51535}.
\bibitem[{Wang et~al.(2023)Wang, Lee, and Lei}]{wangReconstructingTrainingData2023}
\bibinfo{author}{Z.~Wang}, \bibinfo{author}{J.~Lee}, \bibinfo{author}{Q.~Lei},
\newblock \bibinfo{title}{Reconstructing {{Training Data}} from {{Model Gradient}}, {{Provably} }},
\newblock in: \bibinfo{booktitle}{Proceedings of {{The}} 26th {{International Conference}} on {{ Artificial Intelligence}} and {{Statistics}}}, \bibinfo{publisher}{PMLR}, \bibinfo{year}{2023}, pp. \bibinfo{pages}{6595--6612}.
\bibitem[{Carlini et~al.(2023)Carlini, Hayes, Nasr, Jagielski, Sehwag, Tram{\`e}r, Balle, Ippolito, and Wallace}]{carliniExtractingTrainingData2023}
\bibinfo{author}{N.~Carlini}, \bibinfo{author}{J.~Hayes}, \bibinfo{author}{M.~Nasr}, \bibinfo{author}{M.~Jagielski}, \bibinfo{author}{V.~Sehwag}, \bibinfo{author}{F.~Tram{\`e}r}, \bibinfo{author}{B.~Balle}, \bibinfo{author}{D.~Ippolito}, \bibinfo{author}{E.~Wallace},
\newblock \bibinfo{title}{Extracting {{Training Data}} from {{Diffusion Models}}},
\newblock in: \bibinfo{booktitle}{32nd {{USENIX Security Symposium}} ({{USENIX Security}} 23)}, \bibinfo{year}{2023}, pp. \bibinfo{pages}{5253--5270}.
\bibitem[{Morris et~al.(2023)Morris, Kuleshov, Shmatikov, and Rush}]{morrisTextEmbeddingsReveal2023}
\bibinfo{author}{J.~Morris}, \bibinfo{author}{V.~Kuleshov}, \bibinfo{author}{V.~Shmatikov}, \bibinfo{author}{A.~Rush},
\newblock \bibinfo{title}{Text {{Embeddings Reveal}} ({{Almost}}) {{As Much As Text}}},
\newblock in: \bibinfo{editor}{H.~Bouamor}, \bibinfo{editor}{J.~Pino}, \bibinfo{editor}{K.~Bali} (Eds.), \bibinfo{booktitle}{Proceedings of the 2023 {{Conference}} on {{Empirical Methods}} in {{Natural Language Processing}}}, \bibinfo{publisher}{Association for Computational Linguistics}, \bibinfo{address}{Singapore}, \bibinfo{year}{2023}, pp. \bibinfo{pages}{12448--12460}.
\bibitem[{{Gilad-Bachrach} et~al.(2016){Gilad-Bachrach}, Dowlin, Laine, Lauter, Naehrig, and Wernsing}]{gilad-bachrachCryptoNetsApplyingNeural2016}
\bibinfo{author}{R.~{Gilad-Bachrach}}, \bibinfo{author}{N.~Dowlin}, \bibinfo{author}{K.~Laine}, \bibinfo{author}{K.~Lauter}, \bibinfo{author}{M.~Naehrig}, \bibinfo{author}{J.~Wernsing},
\newblock \bibinfo{title}{{{CryptoNets}}: {{Applying Neural Networks}} to {{Encrypted Data}} with {{High Throughput}} and {{Accuracy}}},
\newblock in: \bibinfo{booktitle}{Proceedings of {{The}} 33rd {{International Conference}} on {{ Machine Learning}}}, \bibinfo{publisher}{PMLR}, \bibinfo{year}{2016}, pp. \bibinfo{pages}{201--210}.
\bibitem[{Yi et~al.(2014)Yi, Paulet, and Bertino}]{yiHomomorphicEncryption2014}
\bibinfo{author}{X.~Yi}, \bibinfo{author}{R.~Paulet}, \bibinfo{author}{E.~Bertino},
\newblock \bibinfo{title}{Homomorphic {{Encryption}}},
\newblock in: \bibinfo{editor}{X.~Yi}, \bibinfo{editor}{R.~Paulet}, \bibinfo{editor}{E.~Bertino} (Eds.), \bibinfo{booktitle}{Homomorphic {{Encryption}} and {{Applications}}}, \bibinfo{publisher}{Springer International Publishing}, \bibinfo{address}{Cham}, \bibinfo{year}{2014}, pp. \bibinfo{pages}{27--46}.
\bibitem[{Phong et~al.(2018)Phong, Aono, Hayashi, Wang, and Moriai}]{phongPrivacyPreservingDeepLearning2018}
\bibinfo{author}{L.~T. Phong}, \bibinfo{author}{Y.~Aono}, \bibinfo{author}{T.~Hayashi}, \bibinfo{author}{L.~Wang}, \bibinfo{author}{S.~Moriai},
\newblock \bibinfo{title}{Privacy-{{Preserving Deep Learning}} via {{Additively Homomorphic Encryption}}},
\newblock \bibinfo{journal}{IEEE Transactions on Information Forensics and Security} \bibinfo{volume}{13} (\bibinfo{year}{2018}) \bibinfo{pages}{1333--1345}.
\bibitem[{Zhang et~al.(2019)Zhang, Chen, Yu, and Deng}]{zhangPEFLPrivacyEnhancedFederated2019}
\bibinfo{author}{J.~Zhang}, \bibinfo{author}{B.~Chen}, \bibinfo{author}{S.~Yu}, \bibinfo{author}{H.~Deng},
\newblock \bibinfo{title}{{{PEFL}}: {{A Privacy-Enhanced Federated Learning Scheme}} for {{Big Data Analytics}}},
\newblock \bibinfo{year}{2019}, pp. \bibinfo{pages}{1--6}.
\bibitem[{Hao et~al.(2020)Hao, Li, Luo, Xu, Yang, and Liu}]{haoEfficientPrivacyEnhancedFederated2020}
\bibinfo{author}{M.~Hao}, \bibinfo{author}{H.~Li}, \bibinfo{author}{X.~Luo}, \bibinfo{author}{G.~Xu}, \bibinfo{author}{H.~Yang}, \bibinfo{author}{S.~Liu},
\newblock \bibinfo{title}{Efficient and {{Privacy-Enhanced Federated Learning}} for {{ Industrial Artificial Intelligence}}},
\newblock \bibinfo{journal}{IEEE Transactions on Industrial Informatics} \bibinfo{volume}{16} (\bibinfo{year}{2020}) \bibinfo{pages}{6532--6542}.
\bibitem[{Zhang et~al.(2020)Zhang, Li, Xia, Wang, Yan, and Liu}]{zhangBatchCryptEfficientHomomorphic2020}
\bibinfo{author}{C.~Zhang}, \bibinfo{author}{S.~Li}, \bibinfo{author}{J.~Xia}, \bibinfo{author}{W.~Wang}, \bibinfo{author}{F.~Yan}, \bibinfo{author}{Y.~Liu},
\newblock \bibinfo{title}{\{\vphantom\}{{BatchCrypt}}\vphantom\{\}: {{Efficient Homomorphic Encryption}} for \{\vphantom\}{{Cross-Silo}}\vphantom\{\} {{Federated Learning}}},
\newblock in: \bibinfo{booktitle}{2020 {{USENIX Annual Technical Conference}} ({{USENIX ATC}} 20)}, \bibinfo{year}{2020}, pp. \bibinfo{pages}{493--506}.
\bibitem[{Lindell(2021)}]{lindellSecureMultipartyComputation2021}
\bibinfo{author}{Y.~Lindell},
\newblock \bibinfo{title}{Secure multiparty computation},
\newblock \bibinfo{journal}{Communications of the ACM} \bibinfo{volume}{64} (\bibinfo{year}{2021}) \bibinfo{pages}{86--96}.
\bibitem[{Bonawitz et~al.(2017)Bonawitz, Ivanov, Kreuter, Marcedone, McMahan, Patel, Ramage, Segal, and Seth}]{bonawitzPracticalSecureAggregation2017}
\bibinfo{author}{K.~Bonawitz}, \bibinfo{author}{V.~Ivanov}, \bibinfo{author}{B.~Kreuter}, \bibinfo{author}{A.~Marcedone}, \bibinfo{author}{H.~B. McMahan}, \bibinfo{author}{S.~Patel}, \bibinfo{author}{D.~Ramage}, \bibinfo{author}{A.~Segal}, \bibinfo{author}{K.~Seth},
\newblock \bibinfo{title}{Practical {{Secure Aggregation}} for {{Privacy-Preserving Machine Learning}}},
\newblock in: \bibinfo{booktitle}{Proceedings of the 2017 {{ACM SIGSAC Conference}} on {{Computer}} and {{Communications Security}}}, \bibinfo{publisher}{ACM}, \bibinfo{address}{Dallas Texas USA}, \bibinfo{year}{2017}, pp. \bibinfo{pages}{1175--1191}.
\bibitem[{Dwork(2008)}]{dworkDifferentialPrivacySurvey2008}
\bibinfo{author}{C.~Dwork},
\newblock \bibinfo{title}{Differential {{Privacy}}: {{A Survey}} of {{Results}}},
\newblock in: \bibinfo{editor}{M.~Agrawal}, \bibinfo{editor}{D.~Du}, \bibinfo{editor}{Z.~Duan}, \bibinfo{editor}{A.~Li} (Eds.), \bibinfo{booktitle}{Theory and {{Applications}} of {{Models}} of {{Computation}}}, \bibinfo{publisher}{Springer}, \bibinfo{address}{Berlin, Heidelberg}, \bibinfo{year}{2008}, pp. \bibinfo{pages}{1--19}.
\bibitem[{Ouadrhiri and Abdelhadi(2022)}]{ouadrhiriDifferentialPrivacyDeep2022}
\bibinfo{author}{A.~E. Ouadrhiri}, \bibinfo{author}{A.~Abdelhadi},
\newblock \bibinfo{title}{Differential {{Privacy}} for {{Deep}} and {{Federated Learning}}: {{A Survey}}},
\newblock \bibinfo{journal}{IEEE Access} \bibinfo{volume}{10} (\bibinfo{year}{2022}) \bibinfo{pages}{22359--22380}.
\bibitem[{Agarwal et~al.(2021)Agarwal, Kairouz, and Liu}]{agarwalSkellamMechanismDifferentially2021}
\bibinfo{author}{N.~Agarwal}, \bibinfo{author}{P.~Kairouz}, \bibinfo{author}{Z.~Liu},
\newblock \bibinfo{title}{The {{Skellam Mechanism}} for {{Differentially Private Federated Learning}}},
\newblock in: \bibinfo{booktitle}{Advances in {{Neural Information Processing Systems}}}, volume~\bibinfo{volume}{34}, \bibinfo{publisher}{Curran Associates, Inc.}, \bibinfo{year}{2021}, pp. \bibinfo{pages}{5052--5064}.
\bibitem[{Kim et~al.(2021)Kim, G{\"u}nl{\"u}, and Schaefer}]{kimFederatedLearningLocal2021}
\bibinfo{author}{M.~Kim}, \bibinfo{author}{O.~G{\"u}nl{\"u}}, \bibinfo{author}{R.~F. Schaefer},
\newblock \bibinfo{title}{Federated {{Learning}} with {{Local Differential Privacy}}: {{ Trade-Offs Between Privacy}}, {{Utility}}, and {{Communication}}},
\newblock in: \bibinfo{booktitle}{{{ICASSP}} 2021 - 2021 {{IEEE International Conference}} on {{ Acoustics}}, {{Speech}} and {{Signal Processing}} ({{ICASSP}})}, \bibinfo{year}{2021}, pp. \bibinfo{pages}{2650--2654}.
\bibitem[{{El-Mhamdi} et~al.(2023){El-Mhamdi}, Farhadkhani, Guerraoui, Gupta, Hoang, Pinot, Rouault, and Stephan}]{el-mhamdiImpossibleSafetyLarge2023}
\bibinfo{author}{E.-M. {El-Mhamdi}}, \bibinfo{author}{S.~Farhadkhani}, \bibinfo{author}{R.~Guerraoui}, \bibinfo{author}{N.~Gupta}, \bibinfo{author}{L.-N. Hoang}, \bibinfo{author}{R.~Pinot}, \bibinfo{author}{S.~Rouault}, \bibinfo{author}{J.~Stephan}, \bibinfo{title}{On the {{Impossible Safety}} of {{Large AI Models}}}, \bibinfo{year}{2023}. \href{http://arxiv.org/abs/2209.15259}{{\tt arXiv:2209.15259}}.
\bibitem[{Restuccia et~al.(2017)Restuccia, Ghosh, Bhattacharjee, Das, and Melodia}]{restucciaQualityInformationMobile2017}
\bibinfo{author}{F.~Restuccia}, \bibinfo{author}{N.~Ghosh}, \bibinfo{author}{S.~Bhattacharjee}, \bibinfo{author}{S.~K. Das}, \bibinfo{author}{T.~Melodia},
\newblock \bibinfo{title}{Quality of {{Information}} in {{Mobile Crowdsensing}}: {{Survey}} and {{Research Challenges}}},
\newblock \bibinfo{journal}{ACM Transactions on Sensor Networks} \bibinfo{volume}{13} (\bibinfo{year}{2017}) \bibinfo{pages}{34:1--34:43}.
\bibitem[{Wang and Strong(1996)}]{wangAccuracyWhatData1996}
\bibinfo{author}{R.~Y. Wang}, \bibinfo{author}{D.~M. Strong},
\newblock \bibinfo{title}{Beyond {{Accuracy}}: {{What Data Quality Means}} to {{Data Consumers} }},
\newblock \bibinfo{journal}{Journal of Management Information Systems} \bibinfo{volume}{12} (\bibinfo{year}{1996}) \bibinfo{pages}{5--33}. \href{http://arxiv.org/abs/40398176}{{\tt arXiv:40398176}}.
\bibitem[{Lamport et~al.(1982)Lamport, Shostak, and Pease}]{lamportByzantineGeneralsProblem1982}
\bibinfo{author}{L.~Lamport}, \bibinfo{author}{R.~Shostak}, \bibinfo{author}{M.~Pease},
\newblock \bibinfo{title}{The {{Byzantine Generals Problem}}},
\newblock \bibinfo{journal}{ACM Transactions on Programming Languages and Systems} \bibinfo{volume}{4} (\bibinfo{year}{1982}) \bibinfo{pages}{382--401}.
\bibitem[{Su and Vaidya(2016)}]{suRobustMultiagentOptimization2016}
\bibinfo{author}{L.~Su}, \bibinfo{author}{N.~H. Vaidya},
\newblock \bibinfo{title}{Robust {{Multi-agent Optimization}}: {{Coping}} with {{Byzantine Agents}} with {{Input Redundancy}}},
\newblock in: \bibinfo{editor}{B.~Bonakdarpour}, \bibinfo{editor}{F.~Petit} (Eds.), \bibinfo{booktitle}{Stabilization, {{Safety}}, and {{Security}} of {{Distributed Systems}}}, \bibinfo{publisher}{Springer International Publishing}, \bibinfo{address}{Cham}, \bibinfo{year}{2016}, pp. \bibinfo{pages}{368--382}.
\bibitem[{Blanchard et~al.(2017)Blanchard, El~Mhamdi, Guerraoui, and Stainer}]{blanchardMachineLearningAdversaries2017}
\bibinfo{author}{P.~Blanchard}, \bibinfo{author}{E.~M. El~Mhamdi}, \bibinfo{author}{R.~Guerraoui}, \bibinfo{author}{J.~Stainer},
\newblock \bibinfo{title}{Machine {{Learning}} with {{Adversaries}}: {{Byzantine Tolerant Gradient Descent}}},
\newblock in: \bibinfo{booktitle}{Advances in {{Neural Information Processing Systems}}}, volume~\bibinfo{volume}{30}, \bibinfo{publisher}{Curran Associates, Inc.}, \bibinfo{year}{2017}.
\bibitem[{Yan et~al.(2023)Yan, Wang, Yuan, and Li}]{yanDeFLDefendingModel2023}
\bibinfo{author}{G.~Yan}, \bibinfo{author}{H.~Wang}, \bibinfo{author}{X.~Yuan}, \bibinfo{author}{J.~Li},
\newblock \bibinfo{title}{{{DeFL}}: {{Defending}} against {{Model Poisoning Attacks}} in {{ Federated Learning}} via {{Critical Learning Periods Awareness}}},
\newblock \bibinfo{journal}{Proceedings of the AAAI Conference on Artificial Intelligence} \bibinfo{volume}{37} (\bibinfo{year}{2023}) \bibinfo{pages}{10711--10719}.
\bibitem[{{El-Mhamdi} et~al.(2020){El-Mhamdi}, Guerraoui, Guirguis, Hoang, and Rouault}]{el-mhamdiGenuinelyDistributedByzantine2020}
\bibinfo{author}{E.-M. {El-Mhamdi}}, \bibinfo{author}{R.~Guerraoui}, \bibinfo{author}{A.~Guirguis}, \bibinfo{author}{L.~N. Hoang}, \bibinfo{author}{S.~Rouault},
\newblock \bibinfo{title}{Genuinely {{Distributed Byzantine Machine Learning}}},
\newblock in: \bibinfo{booktitle}{Proceedings of the 39th {{Symposium}} on {{Principles}} of {{ Distributed Computing}}}, \bibinfo{publisher}{ACM}, \bibinfo{address}{Virtual Event Italy}, \bibinfo{year}{2020}, pp. \bibinfo{pages}{355--364}.
\bibitem[{Shejwalkar and Houmansadr(2021)}]{shejwalkarManipulatingByzantineOptimizing2021}
\bibinfo{author}{V.~Shejwalkar}, \bibinfo{author}{A.~Houmansadr},
\newblock \bibinfo{title}{Manipulating the {{Byzantine}}: {{Optimizing Model Poisoning Attacks} } and {{Defenses}} for {{Federated Learning}}},
\newblock in: \bibinfo{booktitle}{Proceedings 2021 {{Network}} and {{Distributed System Security Symposium}}}, \bibinfo{publisher}{Internet Society}, \bibinfo{address}{Virtual}, \bibinfo{year}{2021}.
\bibitem[{Fang et~al.(2020)Fang, Cao, Jia, and Gong}]{fangLocalModelPoisoning2020}
\bibinfo{author}{M.~Fang}, \bibinfo{author}{X.~Cao}, \bibinfo{author}{J.~Jia}, \bibinfo{author}{N.~Z. Gong},
\newblock \bibinfo{title}{Local model poisoning attacks to byzantine-robust federated learning},
\newblock in: \bibinfo{booktitle}{Proceedings of the 29th {{USENIX Conference}} on {{Security Symposium}}}, {{SEC}}'20, \bibinfo{publisher}{USENIX Association}, \bibinfo{address}{USA}, \bibinfo{year}{2020}, pp. \bibinfo{pages}{1623--1640}.
\bibitem[{Farhadkhani et~al.(2022)Farhadkhani, Guerraoui, Hoang, and Villemaud}]{farhadkhaniEquivalenceDataPoisoning2022}
\bibinfo{author}{S.~Farhadkhani}, \bibinfo{author}{R.~Guerraoui}, \bibinfo{author}{L.~N. Hoang}, \bibinfo{author}{O.~Villemaud},
\newblock \bibinfo{title}{An {{Equivalence Between Data Poisoning}} and {{Byzantine Gradient Attacks}}},
\newblock in: \bibinfo{booktitle}{Proceedings of the 39th {{International Conference}} on {{Machine Learning}}}, \bibinfo{publisher}{PMLR}, \bibinfo{year}{2022}, pp. \bibinfo{pages}{6284--6323}.
\bibitem[{{El-Mhamdi} et~al.(2021){El-Mhamdi}, Farhadkhani, Guerraoui, Guirguis, Hoang, and Rouault}]{el-mhamdiCollaborativeLearningJungle2021}
\bibinfo{author}{E.~M. {El-Mhamdi}}, \bibinfo{author}{S.~Farhadkhani}, \bibinfo{author}{R.~Guerraoui}, \bibinfo{author}{A.~Guirguis}, \bibinfo{author}{L.-N. Hoang}, \bibinfo{author}{S.~b. Rouault},
\newblock \bibinfo{title}{Collaborative {{Learning}} in the {{Jungle}} ({{Decentralized}}, {{ Byzantine}}, {{Heterogeneous}}, {{Asynchronous}} and {{Nonconvex Learning}})},
\newblock in: \bibinfo{booktitle}{Advances in {{Neural Information Processing Systems}}}, volume~\bibinfo{volume}{34}, \bibinfo{publisher}{Curran Associates, Inc.}, \bibinfo{year}{2021}, pp. \bibinfo{pages}{25044--25057}.
\bibitem[{Xie et~al.(2018)Xie, Koyejo, and Gupta}]{xieGeneralizedByzantinetolerantSGD2018}
\bibinfo{author}{C.~Xie}, \bibinfo{author}{O.~Koyejo}, \bibinfo{author}{I.~Gupta}, \bibinfo{title}{Generalized {{Byzantine-tolerant SGD}}}, \bibinfo{year}{2018}. \href{http://arxiv.org/abs/1802.10116}{{\tt arXiv:1802.10116}}.
\bibitem[{Li et~al.(2019)Li, Xu, Chen, Giannakis, and Ling}]{liRSAByzantineRobustStochastic2019}
\bibinfo{author}{L.~Li}, \bibinfo{author}{W.~Xu}, \bibinfo{author}{T.~Chen}, \bibinfo{author}{G.~B. Giannakis}, \bibinfo{author}{Q.~Ling},
\newblock \bibinfo{title}{{{RSA}}: {{Byzantine-Robust Stochastic Aggregation Methods}} for {{ Distributed Learning}} from {{Heterogeneous Datasets}}},
\newblock \bibinfo{journal}{Proceedings of the AAAI Conference on Artificial Intelligence} \bibinfo{volume}{33} (\bibinfo{year}{2019}) \bibinfo{pages}{1544--1551}.
\bibitem[{Bagdasaryan et~al.(2020)Bagdasaryan, Veit, Hua, Estrin, and Shmatikov}]{bagdasaryanHowBackdoorFederated2020}
\bibinfo{author}{E.~Bagdasaryan}, \bibinfo{author}{A.~Veit}, \bibinfo{author}{Y.~Hua}, \bibinfo{author}{D.~Estrin}, \bibinfo{author}{V.~Shmatikov},
\newblock \bibinfo{title}{How {{To Backdoor Federated Learning}}},
\newblock in: \bibinfo{booktitle}{Proceedings of the {{Twenty Third International Conference}} on { {Artificial Intelligence}} and {{Statistics}}}, \bibinfo{publisher}{PMLR}, \bibinfo{year}{2020}, pp. \bibinfo{pages}{2938--2948}.
\bibitem[{Chen et~al.(2017)Chen, Liu, Li, Lu, and Song}]{chenTargetedBackdoorAttacks2017}
\bibinfo{author}{X.~Chen}, \bibinfo{author}{C.~Liu}, \bibinfo{author}{B.~Li}, \bibinfo{author}{K.~Lu}, \bibinfo{author}{D.~Song}, \bibinfo{title}{Targeted {{Backdoor Attacks}} on {{Deep Learning Systems Using Data Poisoning}}}, \bibinfo{year}{2017}. \href{http://arxiv.org/abs/1712.05526}{{\tt arXiv:1712.05526}}.
\bibitem[{Tolpegin et~al.(2020)Tolpegin, Truex, Gursoy, and Liu}]{tolpeginDataPoisoningAttacks2020}
\bibinfo{author}{V.~Tolpegin}, \bibinfo{author}{S.~Truex}, \bibinfo{author}{M.~E. Gursoy}, \bibinfo{author}{L.~Liu},
\newblock \bibinfo{title}{Data {{Poisoning Attacks Against Federated Learning Systems}}},
\newblock in: \bibinfo{editor}{L.~Chen}, \bibinfo{editor}{N.~Li}, \bibinfo{editor}{K.~Liang}, \bibinfo{editor}{S.~Schneider} (Eds.), \bibinfo{booktitle}{Computer {{Security}} -- {{ESORICS}} 2020}, \bibinfo{publisher}{Springer International Publishing}, \bibinfo{address}{Cham}, \bibinfo{year}{2020}, pp. \bibinfo{pages}{480--501}.
\bibitem[{Shafahi et~al.(2018)Shafahi, Huang, Najibi, Suciu, Studer, Dumitras, and Goldstein}]{shafahiPoisonFrogsTargeted2018}
\bibinfo{author}{A.~Shafahi}, \bibinfo{author}{W.~R. Huang}, \bibinfo{author}{M.~Najibi}, \bibinfo{author}{O.~Suciu}, \bibinfo{author}{C.~Studer}, \bibinfo{author}{T.~Dumitras}, \bibinfo{author}{T.~Goldstein},
\newblock \bibinfo{title}{Poison {{Frogs}}! {{Targeted Clean-Label Poisoning Attacks}} on {{ Neural Networks}}},
\newblock in: \bibinfo{booktitle}{Advances in {{Neural Information Processing Systems}}}, volume~\bibinfo{volume}{31}, \bibinfo{publisher}{Curran Associates, Inc.}, \bibinfo{year}{2018}.
\bibitem[{Zhang et~al.(2021)Zhang, Chen, Cheng, Binh, and Yu}]{zhangPoisonGANGenerativePoisoning2021}
\bibinfo{author}{J.~Zhang}, \bibinfo{author}{B.~Chen}, \bibinfo{author}{X.~Cheng}, \bibinfo{author}{H.~T.~T. Binh}, \bibinfo{author}{S.~Yu},
\newblock \bibinfo{title}{{{PoisonGAN}}: {{Generative Poisoning Attacks Against Federated Learning}} in {{Edge Computing Systems}}},
\newblock \bibinfo{journal}{IEEE Internet of Things Journal} \bibinfo{volume}{8} (\bibinfo{year}{2021}) \bibinfo{pages}{3310--3322}.
\bibitem[{Zhang et~al.(2019)Zhang, Chen, Wu, Chen, and Yu}]{zhangPoisoningAttackFederated2019}
\bibinfo{author}{J.~Zhang}, \bibinfo{author}{J.~Chen}, \bibinfo{author}{D.~Wu}, \bibinfo{author}{B.~Chen}, \bibinfo{author}{S.~Yu},
\newblock \bibinfo{title}{Poisoning {{Attack}} in {{Federated Learning}} using {{Generative Adversarial Nets}}},
\newblock in: \bibinfo{booktitle}{2019 18th {{IEEE International Conference On Trust}}, {{Security And Privacy In Computing And Communications}}/13th {{IEEE International Conference On Big Data Science And Engineering}} ({{ TrustCom}}/{{BigDataSE}})}, \bibinfo{year}{2019}, pp. \bibinfo{pages}{374--380}.
\bibitem[{Shejwalkar et~al.(2022)Shejwalkar, Houmansadr, Kairouz, and Ramage}]{shejwalkarBackDrawingBoard2022}
\bibinfo{author}{V.~Shejwalkar}, \bibinfo{author}{A.~Houmansadr}, \bibinfo{author}{P.~Kairouz}, \bibinfo{author}{D.~Ramage},
\newblock \bibinfo{title}{Back to the {{Drawing Board}}: {{A Critical Evaluation}} of {{ Poisoning Attacks}} on {{Production Federated Learning}}},
\newblock in: \bibinfo{booktitle}{2022 {{IEEE Symposium}} on {{Security}} and {{Privacy}} ({{SP}})}, \bibinfo{year}{2022}, pp. \bibinfo{pages}{1354--1371}.
\bibitem[{Wu et~al.(2021)Wu, Chen, Wang, Liu, Nguyen, and Yesha}]{wuToleratingAdversarialAttacks2021}
\bibinfo{author}{Y.~Wu}, \bibinfo{author}{H.~Chen}, \bibinfo{author}{X.~Wang}, \bibinfo{author}{C.~Liu}, \bibinfo{author}{P.~Nguyen}, \bibinfo{author}{Y.~Yesha},
\newblock \bibinfo{title}{Tolerating {{Adversarial Attacks}} and {{Byzantine Faults}} in {{ Distributed Machine Learning}}},
\newblock in: \bibinfo{booktitle}{2021 {{IEEE International Conference}} on {{Big Data}} ({{Big Data}})}, \bibinfo{year}{2021}, pp. \bibinfo{pages}{3380--3389}.
\bibitem[{Damaskinos et~al.(2018)Damaskinos, Mhamdi, Guerraoui, Patra, and Taziki}]{damaskinosAsynchronousByzantineMachine2018}
\bibinfo{author}{G.~Damaskinos}, \bibinfo{author}{E.~M.~E. Mhamdi}, \bibinfo{author}{R.~Guerraoui}, \bibinfo{author}{R.~Patra}, \bibinfo{author}{M.~Taziki},
\newblock \bibinfo{title}{Asynchronous {{Byzantine Machine Learning}} (the case of {{SGD}})},
\newblock in: \bibinfo{booktitle}{Proceedings of the 35th {{International Conference}} on {{Machine Learning}}}, \bibinfo{publisher}{PMLR}, \bibinfo{year}{2018}, pp. \bibinfo{pages}{1145--1154}.
\bibitem[{Mhamdi et~al.(2018)Mhamdi, Guerraoui, and Rouault}]{mhamdiHiddenVulnerabilityDistributed2018}
\bibinfo{author}{E.~M.~E. Mhamdi}, \bibinfo{author}{R.~Guerraoui}, \bibinfo{author}{S.~Rouault},
\newblock \bibinfo{title}{The {{Hidden Vulnerability}} of {{Distributed Learning}} in {{ Byzantium}}},
\newblock in: \bibinfo{booktitle}{Proceedings of the 35th {{International Conference}} on {{Machine Learning}}}, \bibinfo{publisher}{PMLR}, \bibinfo{year}{2018}, pp. \bibinfo{pages}{3521--3530}.
\bibitem[{Minsker(2015)}]{minskerGeometricMedianRobust2015}
\bibinfo{author}{S.~Minsker},
\newblock \bibinfo{title}{Geometric median and robust estimation in {{Banach}} spaces},
\newblock \bibinfo{journal}{Bernoulli} \bibinfo{volume}{21} (\bibinfo{year}{2015}) \bibinfo{pages}{2308--2335}.
\bibitem[{Yin et~al.(2018)Yin, Chen, Kannan, and Bartlett}]{yinByzantineRobustDistributedLearning2018}
\bibinfo{author}{D.~Yin}, \bibinfo{author}{Y.~Chen}, \bibinfo{author}{R.~Kannan}, \bibinfo{author}{P.~Bartlett},
\newblock \bibinfo{title}{Byzantine-{{Robust Distributed Learning}}: {{Towards Optimal Statistical Rates}}},
\newblock in: \bibinfo{booktitle}{Proceedings of the 35th {{International Conference}} on {{Machine Learning}}}, \bibinfo{publisher}{PMLR}, \bibinfo{year}{2018}, pp. \bibinfo{pages}{5650--5659}.
\bibitem[{Xia et~al.(2019)Xia, Tao, Hao, and Li}]{xiaFABAAlgorithmFast2019}
\bibinfo{author}{Q.~Xia}, \bibinfo{author}{Z.~Tao}, \bibinfo{author}{Z.~Hao}, \bibinfo{author}{Q.~Li},
\newblock \bibinfo{title}{{{FABA}}: {{An Algorithm}} for {{Fast Aggregation}} against {{ Byzantine Attacks}} in {{Distributed Neural Networks}}},
\newblock \bibinfo{journal}{IJCAI}  (\bibinfo{year}{2019}).
\bibitem[{Wu et~al.(2023)Wu, Chen, and Ling}]{wuByzantineResilientDecentralizedStochastic2023}
\bibinfo{author}{Z.~Wu}, \bibinfo{author}{T.~Chen}, \bibinfo{author}{Q.~Ling},
\newblock \bibinfo{title}{Byzantine-{{Resilient Decentralized Stochastic Optimization With Robust Aggregation Rules}}},
\newblock \bibinfo{journal}{IEEE Transactions on Signal Processing} \bibinfo{volume}{71} (\bibinfo{year}{2023}) \bibinfo{pages}{3179--3195}.
\bibitem[{Fung et~al.(2020)Fung, Yoon, and Beschastnikh}]{fungMitigatingSybilsFederated2020}
\bibinfo{author}{C.~Fung}, \bibinfo{author}{C.~J.~M. Yoon}, \bibinfo{author}{I.~Beschastnikh}, \bibinfo{title}{Mitigating {{Sybils}} in {{Federated Learning Poisoning}}}, \bibinfo{year}{2020}. \href{http://arxiv.org/abs/1808.04866}{{\tt arXiv:1808.04866}}.
\bibitem[{Cao et~al.(2019)Cao, Chang, Lin, Liu, and Sun}]{caoUnderstandingDistributedPoisoning2019}
\bibinfo{author}{D.~Cao}, \bibinfo{author}{S.~Chang}, \bibinfo{author}{Z.~Lin}, \bibinfo{author}{G.~Liu}, \bibinfo{author}{D.~Sun},
\newblock \bibinfo{title}{Understanding {{Distributed Poisoning Attack}} in {{Federated Learning}}},
\newblock in: \bibinfo{booktitle}{2019 {{IEEE}} 25th {{International Conference}} on {{Parallel}} and {{Distributed Systems}} ({{ICPADS}})}, \bibinfo{publisher}{IEEE}, \bibinfo{address}{Tianjin, China}, \bibinfo{year}{2019}, pp. \bibinfo{pages}{233--239}.
\bibitem[{Byabazaire et~al.(2020)Byabazaire, O'Hare, and Delaney}]{byabazaireUsingTrustMeasure2020}
\bibinfo{author}{J.~Byabazaire}, \bibinfo{author}{G.~O'Hare}, \bibinfo{author}{D.~Delaney},
\newblock \bibinfo{title}{Using {{Trust}} as a {{Measure}} to {{Derive Data Quality}} in {{Data Shared IoT Deployments}}},
\newblock in: \bibinfo{booktitle}{2020 29th {{International Conference}} on {{Computer Communications}} and {{Networks}} ({{ICCCN}})}, \bibinfo{year}{2020}, pp. \bibinfo{pages}{1--9}.
\bibitem[{Chuprov et~al.(2020)Chuprov, Viksnin, Kim, Reznikand, and Khokhlov}]{chuprovReputationTrustModels2020}
\bibinfo{author}{S.~Chuprov}, \bibinfo{author}{I.~Viksnin}, \bibinfo{author}{I.~Kim}, \bibinfo{author}{L.~Reznikand}, \bibinfo{author}{I.~Khokhlov},
\newblock \bibinfo{title}{Reputation and {{Trust Models}} with {{Data Quality Metrics}} for {{ Improving Autonomous Vehicles Traffic Security}} and {{Safety}}},
\newblock in: \bibinfo{booktitle}{2020 {{IEEE Systems Security Symposium}} ({{SSS}})}, \bibinfo{year}{2020}, pp. \bibinfo{pages}{1--8}.
\bibitem[{Jia et~al.(2021)Jia, Yaghini, {Choquette-Choo}, Dullerud, Thudi, Chandrasekaran, and Papernot}]{jiaProofofLearningDefinitionsPractice2021}
\bibinfo{author}{H.~Jia}, \bibinfo{author}{M.~Yaghini}, \bibinfo{author}{C.~A. {Choquette-Choo}}, \bibinfo{author}{N.~Dullerud}, \bibinfo{author}{A.~Thudi}, \bibinfo{author}{V.~Chandrasekaran}, \bibinfo{author}{N.~Papernot},
\newblock \bibinfo{title}{Proof-of-{{Learning}}: {{Definitions}} and {{Practice}}},
\newblock in: \bibinfo{booktitle}{2021 {{IEEE Symposium}} on {{Security}} and {{Privacy}} ({{SP}})}, \bibinfo{year}{2021}, pp. \bibinfo{pages}{1039--1056}.
\bibitem[{Garg et~al.(2023)Garg, Goel, Jha, Mahloujifar, Mahmoody, Policharla, and Wang}]{gargExperimentingZeroKnowledgeProofs2023}
\bibinfo{author}{S.~Garg}, \bibinfo{author}{A.~Goel}, \bibinfo{author}{S.~Jha}, \bibinfo{author}{S.~Mahloujifar}, \bibinfo{author}{M.~Mahmoody}, \bibinfo{author}{G.-V. Policharla}, \bibinfo{author}{M.~Wang},
\newblock \bibinfo{title}{Experimenting with {{Zero-Knowledge Proofs}} of {{Training}}},
\newblock in: \bibinfo{booktitle}{Proceedings of the 2023 {{ACM SIGSAC Conference}} on {{Computer}} and {{Communications Security}}}, \bibinfo{publisher}{ACM}, \bibinfo{address}{Copenhagen Denmark}, \bibinfo{year}{2023}, pp. \bibinfo{pages}{1880--1894}.
\bibitem[{Karimireddy et~al.(2021)Karimireddy, He, and Jaggi}]{karimireddyLearningHistoryByzantine2021}
\bibinfo{author}{S.~P. Karimireddy}, \bibinfo{author}{L.~He}, \bibinfo{author}{M.~Jaggi},
\newblock \bibinfo{title}{Learning from {{History}} for {{Byzantine Robust Optimization}}},
\newblock in: \bibinfo{booktitle}{Proceedings of the 38th {{International Conference}} on {{Machine Learning}}}, \bibinfo{publisher}{PMLR}, \bibinfo{year}{2021}, pp. \bibinfo{pages}{5311--5319}.
\bibitem[{Wu et~al.(2023)Wu, Chen, Xie, and Huang}]{wuOnePixelShortcutLearning2023}
\bibinfo{author}{S.~Wu}, \bibinfo{author}{S.~Chen}, \bibinfo{author}{C.~Xie}, \bibinfo{author}{X.~Huang}, \bibinfo{title}{One-{{Pixel Shortcut}}: On the {{Learning Preference}} of {{Deep Neural Networks}}}, \bibinfo{year}{2023}. \href{http://arxiv.org/abs/2205.12141}{{\tt arXiv:2205.12141}}.
\bibitem[{Geirhos et~al.(2020)Geirhos, link will open in a new~tab {Link to external site}, {J{\"o}rn-Henrik}, Michaelis, link will open in a new~tab {Link to external site}, Zemel, Wieland, Matthias, Wichmann, and link will open in a new~tab {Link to external site}}]{geirhosShortcutLearningDeep2020}
\bibinfo{author}{R.~Geirhos}, \bibinfo{author}{t.~link will open in a new~tab {Link to external site}}, \bibinfo{author}{J.~{J{\"o}rn-Henrik}}, \bibinfo{author}{C.~Michaelis}, \bibinfo{author}{t.~link will open in a new~tab {Link to external site}}, \bibinfo{author}{R.~Zemel}, \bibinfo{author}{B.~Wieland}, \bibinfo{author}{B.~Matthias}, \bibinfo{author}{F.~A. Wichmann}, \bibinfo{author}{t.~link will open in a new~tab {Link to external site}},
\newblock \bibinfo{title}{Shortcut learning in deep neural networks},
\newblock \bibinfo{journal}{Nature Machine Intelligence} \bibinfo{volume}{2} (\bibinfo{year}{2020}) \bibinfo{pages}{665--673}.
\bibitem[{Wang et~al.(2023)Wang, Liu, and He}]{wangConceptDriftBasedCheckpointRestart2023}
\bibinfo{author}{L.~Wang}, \bibinfo{author}{J.~Liu}, \bibinfo{author}{Q.~He},
\newblock \bibinfo{title}{Concept {{Drift-Based Checkpoint-Restart}} for {{Edge Services Rejuvenation}}},
\newblock \bibinfo{journal}{IEEE Transactions on Services Computing} \bibinfo{volume}{16} (\bibinfo{year}{2023}) \bibinfo{pages}{1713--1725}.
\bibitem[{Guastella et~al.(2021)Guastella, Marcillaud, and Valenti}]{guastellaEdgeBasedMissingData2021}
\bibinfo{author}{D.~A. Guastella}, \bibinfo{author}{G.~Marcillaud}, \bibinfo{author}{C.~Valenti},
\newblock \bibinfo{title}{Edge-{{Based Missing Data Imputation}} in {{Large-Scale Environments} }},
\newblock \bibinfo{journal}{Information} \bibinfo{volume}{12} (\bibinfo{year}{2021}) \bibinfo{pages}{195}.
\bibitem[{Izzo et~al.(2023)Izzo, Ying, Zou, and Chatterjee}]{izzoTheoryAlgorithmsDatacentric2023}
\bibinfo{author}{Z.~Izzo}, \bibinfo{author}{L.~Ying}, \bibinfo{author}{J.~Zou}, \bibinfo{author}{S.~Chatterjee}, \bibinfo{title}{Theory and Algorithms for Data-Centric Machine Learning}, Ph.D. thesis, Stanford University, \bibinfo{address}{Stanford, California}, \bibinfo{year}{2023}.
\bibitem[{Qi et~al.(2021)Qi, Wang, Li, and Gao}]{qiImpactsDirtyData2021}
\bibinfo{author}{Z.~Qi}, \bibinfo{author}{H.~Wang}, \bibinfo{author}{J.~Li}, \bibinfo{author}{H.~Gao}, \bibinfo{title}{Impacts of {{Dirty Data}}: And {{Experimental Evaluation}}}, \bibinfo{year}{2021}. \href{http://arxiv.org/abs/1803.06071}{{\tt arXiv:1803.06071}}.
\bibitem[{Zhou et~al.(2022)Zhou, Lin, Pi, Zhang, Xu, Cui, and Zhang}]{zhouModelAgnosticSample2022}
\bibinfo{author}{X.~Zhou}, \bibinfo{author}{Y.~Lin}, \bibinfo{author}{R.~Pi}, \bibinfo{author}{W.~Zhang}, \bibinfo{author}{R.~Xu}, \bibinfo{author}{P.~Cui}, \bibinfo{author}{T.~Zhang},
\newblock \bibinfo{title}{Model {{Agnostic Sample Reweighting}} for {{Out-of-Distribution Learning}}},
\newblock in: \bibinfo{booktitle}{Proceedings of the 39th {{International Conference}} on {{Machine Learning}}}, \bibinfo{publisher}{PMLR}, \bibinfo{year}{2022}, pp. \bibinfo{pages}{27203--27221}.

\end{thebibliography}
\addcontentsline{toc}{section}{References}

\end{document}